\documentclass[10pt,journal,compsoc]{IEEEtran}
\ifCLASSOPTIONcompsoc
  \usepackage[nocompress]{cite}
\else
  \usepackage{cite}
\fi
\ifCLASSINFOpdf
\else
\fi

\usepackage{times}
\usepackage{epsfig}
\usepackage{graphicx}
\usepackage{amsmath}
\usepackage{amssymb}
\usepackage{bbm}
\usepackage{subfig}
\usepackage{enumitem}
\usepackage{slashbox}
\usepackage{color,soul}

\begin{document}
\bstctlcite{IEEEexample:BSTcontrol}

\newcommand\allX{\mathbb X}
\newcommand\oneX{\mathbf X}
\newcommand\shape{\mathbf S}

\newcommand\allT{\mathbb W}
\newcommand\oneT{\mathbf W}

\newcommand\allC{\mathbb C}
\newcommand{\oneC}{\mathbf C}

\newcommand{\allr}{\mathbbm r}
\newcommand{\oner}{\mathbf r}

\newcommand{\alld}{\mathbbm d}
\newcommand{\oned}{\mathbf d}

\newcommand\todo[1]{\textbf{\textcolor{red}{TODO: #1}}}
\newcommand\done[1]{\textcolor{green}{#1}}

\newcommand{\csubfloat}[2][]{%
  \makebox[0pt]{\subfloat[#1]{#2}}%
}
\newcommand{\etal}{\mbox{\emph{et al.\ }}}
\newcommand{\ie}{\mbox{\emph{i. e.\ }}}

\title{Self-expressive Dictionary Learning for Dynamic 3D Reconstruction}

\author{Enliang~Zheng,~\IEEEmembership{Member,~IEEE,}
        Dinghuang~Ji,~Enrique Dunn,~\IEEEmembership{Member,~IEEE,}
        and~Jan-Michael~Frahm,~\IEEEmembership{Member,~IEEE}% <-this % stops a space
\IEEEcompsocitemizethanks{\IEEEcompsocthanksitem The authors are with the Department of Computer Science, the University of North Carolina at Chapel Hill, Chapel Hill,
NC, 27599.\protect\\
E-mail: \{ezheng, jdh, dunn, jmf\}@cs.unc.edu
\IEEEcompsocthanksitem An earlier version of this paper appears in the International Conference on Computer vision (ICCV), Dec. 13-16, 2015 \cite{zheng2015l1}.
}% <-this % stops an unwanted space
%\thanks{Manuscript received April 19, 2005; revised September 17, 2014.}
}

\IEEEtitleabstractindextext{%
%\begin{abstract}
%The abstract goes here.
%\end{abstract}
% !TEX root = video3D_l1.tex

\begin{abstract}

We target the problem of sparse 3D reconstruction of dynamic objects observed by multiple unsynchronized video cameras with unknown temporal overlap.
To this end, we develop a framework to recover the unknown structure without sequencing information across video sequences. Our proposed compressed sensing framework poses the estimation of 3D structure as the problem of dictionary learning, where the dictionary is defined as an aggregation of the temporally varying 3D structures. Given the smooth motion of dynamic objects, we observe any element in the dictionary can be well approximated by a sparse linear combination of other elements in the same dictionary (\ie~self-expression). Moreover, the sparse coefficients describing a locally linear 3D structural interpolation reveal the local sequencing information. Our formulation optimizes a biconvex cost function that leverages a compressed sensing formulation and enforces both structural dependency coherence across video streams, as well as motion smoothness across estimates from common video sources. We further analyze the reconstructability of our approach under different capture scenarios, and its comparison and relation to existing methods. Experimental results on large amounts of synthetic data as well as real imagery demonstrate the effectiveness of our approach.
%We solve this problem in an iterative and alternating manner, where we optimize for 3D structure while fixing sequencing information, and {\em vice versa}.

\end{abstract}

}

\maketitle

\IEEEdisplaynontitleabstractindextext

% \fi
\IEEEpeerreviewmaketitle

% !TEX root = video3D_l1.tex

\section{Introduction}

%Given the rapid development of mobile technology, 

%Given the bursting amount of crowd-sourced images and videos available online, there is an urging needs to organize them 
%As the rapid development of mobile technology, most people have their own 

%Last decade has witnessed a bursting amount of crowd sourced images and videos available online. This is largely due to the widely availability of handy mobile cameras.

\IEEEPARstart{T}{hanks} to the rapid development of mobile technology, it has become common that many people use their own cameras to capture a common event of interest, such as a concert or a wedding. These real-life videos and photos usually have the dynamic objects as the main focus of the scene. With the bursting growth of such crowd-sourced data, it is of interest to develop methods of dynamic scene analysis that enrich understanding and visualization of the captured events.
%process it for further applications such as automatic event analysis or better visualization of the dynamic scene. 

\begin{figure}
\centering
\includegraphics[width=0.99\columnwidth]{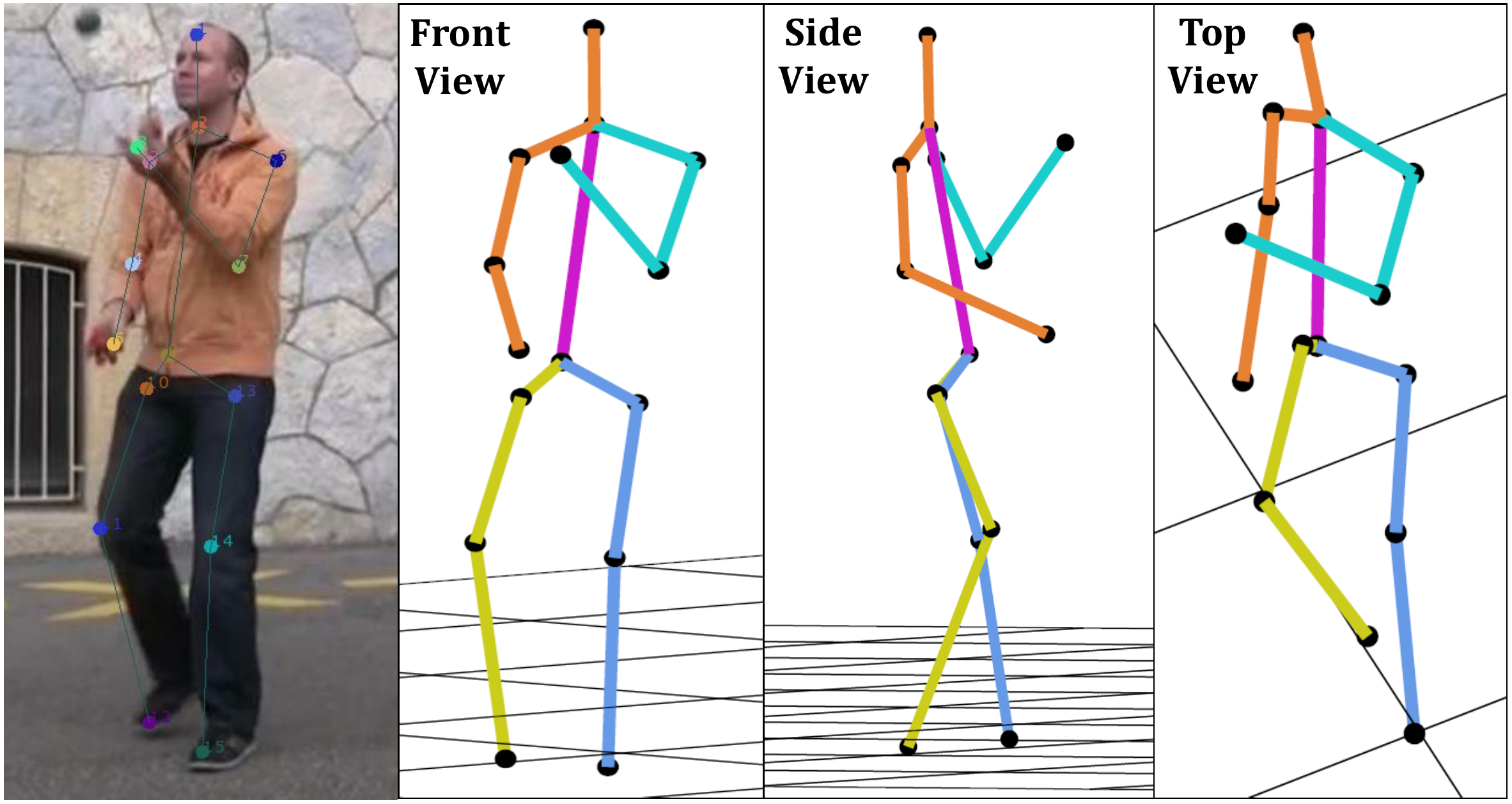}
\caption{\label{fig:first_image}(Left) Example frame from the multiple videos capturing a performance serving as input to our method, with overlaid structure (points), and (right three) different views of the reconstructed 3D points. Note our method only estimates the 3D points but no topology. The skeleton lines are plotted  for visualization purposes.}
\end{figure}

\begin{figure*}
\centering
\includegraphics[width=0.95\textwidth]{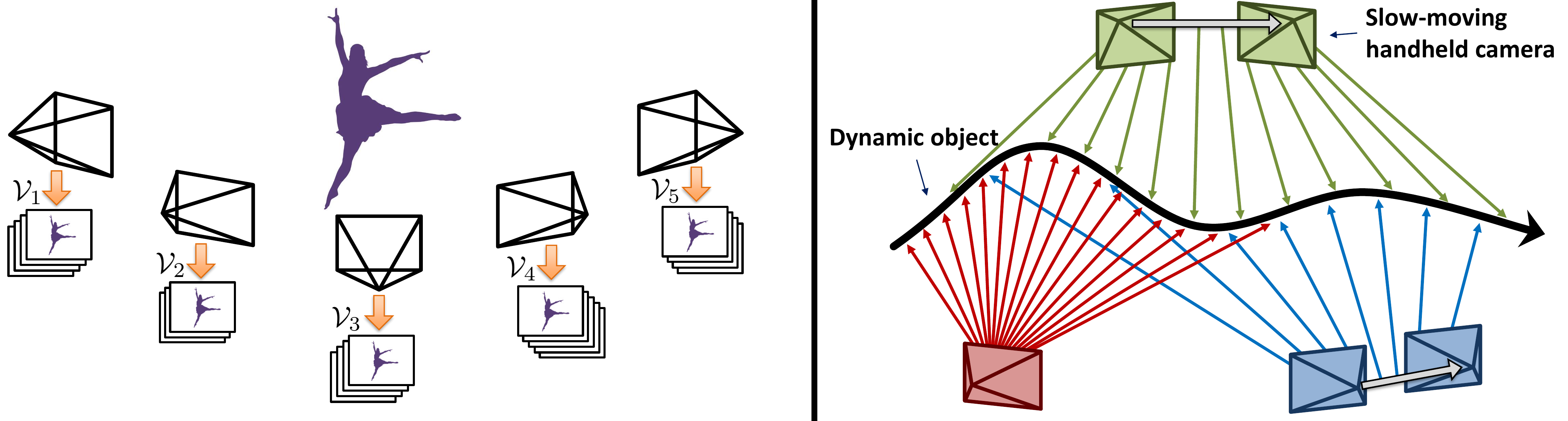}
\caption{\label{fig:overview} Left: Multiple videos capture a performance.
The corresponding set of independent image streams  serves as input to our method.
Right: Each input video has a different sampling of a 3D point's trajectory.}
\end{figure*}

In this work, we target the problem of dynamic 3D object reconstruction from multiple unsynchronized videos. 
More specifically, the method takes as input a collection of video streams without inter-sequence temporal information. The video streams could potentially have different, irregular, and unknown frame rates (see Fig.~\ref{fig:overview}). 
As output, the method reconstructs the 3D positions of sparse feature points at each time instance (\emph{e.g.}, Fig.~\ref{fig:first_image}). 
Dynamic object reconstruction from unsynchronized videos is a challenging problem due to various factors, such as unknown temporal overlap among video streams, possible non-concurrent captures, and dynamic object motion. Any of these factors impedes the valid reconstruction from traditional 3D triangulation, which relies on the assumption of concurrent captures or a static scene.

Despite the ubiquity of uncontrolled video collections, there are currently no methods that can successfully address our problem.
Static scene reconstruction from photo collections has reached a high level of maturity~\cite{Snavely2,zheng2014patchmatch,Heinly} thanks to the development of structure from motion and depth estimation, but the reconstruction of dynamic objects using videos currently falls far behind the maturity of reconstruction of static scene elements. Existing methods of trajectory triangulation \cite{park20103d, valmadre2012general} from monocular image sequences inherently require temporal order information (sequencing information).
However, with independently captured videos, it is challenging to obtain this  information across videos. Zheng \etal~\cite{zheng2014joint} recently propose to jointly estimate the photo sequencing and 3D point estimation by solving a generalized minimum spanning tree (GMST) problem. However, the NP-hard GMST problem itself limits the scalability of the approach. Also in this vein, the non-rigid structure from motion (NRSFM) problems have received extensive study over the two decades \cite{Tomasi_IJCV92,hartley2008perspective,dai2014simple}, but such methods are still under further exploration, especially if a perspective camera model is applied. 

%However, there are significant challenges ahead in order to leverage uncontrolled video capture such as those available in crowd sourced video collections.
% which is typically the main interest of real-life videos or photos. 

%only reconstruct the static parts of the environment captured by the photo collections.However, most real-life videos and photos have dynamic elements, e.g., imagery with people as the main object of interest.
%Take, for example, videos captured at music concerts, sports events, etc.
%It is these dynamic objects that we often aim to capture as they bring the static scenes to life.

%In this paper, we specifically target the reconstruction of the shape of dynamic objects captured by a variety of unsynchronized video cameras (See Fig.~\ref{fig:overview}).
%This setup is encountered, for example, when several people capture an event, such as people dancing, each using their own cameras.
%In that case, all videos capture the same event, but since the cameras are not synchronized, the temporal order of the frames is only known within each video sequence.
%Hence, for a successful reconstruction, it is required to recover the temporal alignment across video sequences, accounting for the potentially different and unknown frame rates of the cameras, and to determine the shape of the dynamic objects.
To solve the problem, we observe that,
given the smooth motion of a dynamic object, any 3D shape at one time instance can be sparsely approximated by other shapes across time. 
Based on this self-expressive representation, our solution leverages the compressive sensing technique ($l_1$ norm), and tackles the problem in a dictionary learning framework \cite{aharon2006img,elad2006image}, where the dictionary is defined by the temporally varying 3D structure. Though the self-expression technique has been previously used in subspace clustering for motion segmentation \cite{elhamifar2009sparse}, and dictionary learning has been used in other applications such as image denoising \cite{elad2006image}, we are the first to explore learning a self-expressive dictionary for the problem of dynamic object reconstruction. 

%Fig.~\ref{fig:first_image} shows one sample output of our approach.

% Talk about the difficulty and the state of the art

%In this paper, we propose a novel approach to jointly determine the sequencing information and the object shape.
%To achieve scalability, our approach relies on compressed sensing~\cite{ramalingam2013lifting} and learned motion trajectory bases.
%In this paper,  we propose a continuous formulation that jointly poses the problems of dynamic structure estimation and cross-stream image sequencing  as a compressive sensing dictionary learning task \cite{elad2006image,aharon2006img}.

The remainder of the paper is organized as follows.
We briefly discuss related works in Sec.~\ref{sec:related}.
After introducing the notations in Sec.~\ref{sec:problem_and_notations}, we begin describing foundations of our proposed approach in Sec.~\ref{sec:principle}. Sec.~\ref{sec:method} presents our model for dynamic object reconstruction without sequencing information, followed by the parameterization of the 3D structure given different kinds of 2D measures in Sec.~\ref{sec:parameterization}.
Sec.~\ref{sec:solver} describes our ADMM-based optimization solver to minimize the model. Then, Sec.~\ref{sec:reconstructability} illustrates the reconstructablity of our algorithm. We provide experimental evaluations in Sec.~\ref{sec:experiment}, and conclude the paper in Sec.~\ref{sec:conclusion}
%In the end, we conclude the paper with an experimental evaluation in Sec.~\ref{sec:experiment}.

% !TEX root = video3D_l1.tex

\section{Related Work}
\label{sec:related}
\subsection{Trajectory triangulation}
Our work is closely related to trajectory triangulation from monocular images \cite{avidan2000trajectory,park20103d,valmadre2012general,zheng2014joint,ZhuCL_CVPR11,park20153d}.
Avidan and Shashua \cite{avidan2000trajectory} first coined the task of trajectory triangulation that reconstructs the 3D coordinates of a moving point from monocular images. That is, each dynamic point is observed by only one camera at a time. Their method assumes the dynamic point moves along a simple parametric trajectory, such as a straight line or a conic section.

Recent works focus on a more general model for trajectory triangulation.
Park \etal~\cite{park20103d} represent the trajectory with a linear combination of low-order discrete cosine transform (DCT) bases, and the trajectory is triangulated by estimating the coefficients of the linear combination.
%The approach requires a known temporal order of all observations, i.e.~the temporal order of the multiple image sequences has to be known during the computation.
There are two fundamental limitations of their method as observed in  \cite{valmadre2012general}.
First, there is no automated scheme to determine the optimal number ($K$) of DCT bases.
Second, the correlation between the object trajectory and the camera motion inherently limits the reconstruction accuracy.
To overcome the first limitation, Park \etal~\cite{park20153d} select $K$ by checking the consistency of the reconstructed trajectory in an N-cross validation scheme.
Alternatively, Valmadre \etal~\cite{valmadre2012general} propose a new method without using DCT bases. They estimate the trajectory by minimizing the trajectory's response to a bank of high-pass filters.
To overcome the second limitation,
Zhu \etal~\cite{ZhuCL_CVPR11} propose to incorporate the 3D structures of a number of key frames to enhance the reconstructability. However, obtaining those key-frame 3D structures requires manual interaction. 
The methods in \cite{park20103d,valmadre2012general,zheng2014joint,ZhuCL_CVPR11} require the sequencing information of the images, 
but in natural capture setups, the availability of sequencing information and high reconstructability typically cannot be fulfilled simultaneously \cite{ZhuCL_CVPR11,park20153d}. 

Zheng \etal~\cite{zheng2014joint} address a slightly different problem. They triangulate the object class trajectory, which is defined by the connection of the objects of the same class moving in a common 3D path, from a collection of unordered images. Their method jointly estimates the trajectory and sequencing, but has low scalability and efficiency due to the NP-hard GMST problem. In contrast, our proposed method reconstructs the dynamic objects without sequencing information across videos.

%Recently, Vicente \etal \cite{vicente2014reconstructing} introduced a dense reconstruction approach from unordered image sets leveraging manual object segmentation and semantic labeling for objects contained in the PASCAL-VOC dataset. Their method relies on the fact that similar objects of the same class in a similar pose are visible within the dataset. In contrast, our proposed method reconstructs the object shape without requiring the presence of similar object shapes in the dataset or leveraging semantic information to enable the reconstruction. Russell \etal \cite{russell2014video} introduce a monocular video based reconstruction method, which relies on hierarchical segmentation of a scene into objects, object parts, and background, in order to successfully reconstruct the scene. This approach assumes a known temporal ordering of the video frames and only considers a single camera's information. Our proposed method does not require known temporal ordering and leverages all available observations.

\subsection{Non-rigid SfM}
One class of related works solve the non-rigid structure from motion (NRSFM) problem, which targets simultaneous recovery of camera motion and 3D structure using an image sequence. These methods typically start from a set of 2D correspondences across frames. 
%As an important extension of the well-known Tomasi-Kanade factorization\cite{tomasi1992shape}, 
The work by Bregler \etal~\cite{Bregler_CVPR2000} tackles the NRSFM problem through matrix factorization, with the assumption that deforming non-rigid objects can be represented by a linear combination of low-order shape bases. 
%It was later shown that in order to achieve a unique solution, more than just the orthogonality constraints have to be used~\cite{Xiao_ECCV2004}. To solve the ambiguity, prior knowledge is required to obtain a unique solution. 
It is later shown by Xiao \etal~\cite{Xiao_ECCV2004} that utilizing only orthogonality constraints on the camera rotation is not enough, and a basis prior is required to uniquely determine the shape bases. Not until very recently, Dai \etal~\cite{dai2014simple} propose a new prior-free method that minimizes the nuclear norm of the shape matrix.

As a dual method to the above shape-based methods, Akhter \etal~\cite{Akhter_NIPS08} propose the first trajectory-based NRSFM approach, which leverages DCT bases to approximately represent object point trajectories.
%While the trajectory-based NRSFM strictly require the sequencing information, 
%Shape-based NRSFM approaches typically do not require the sequencing information as opposed to trajectory-based NRSFM approaches. 
While shape-based approaches typically do not require sequencing information,  trajectory-based approaches completely fail if image frames are randomly shuffled \cite{dai2014simple}.

At first glance, it seems that the shape-based approaches can be applied to our problem without much modification. 
However, these approaches assume orthographic or weak perspective camera models, and 
%It is well known that linear orthographic transform makes problems easier comparing to nonlinear projective transform. 
%It is well known structure recovery with the nonlinear projective model is more difficult than the linear orthographic model, since the former model has one more unknown variable of depth.
it has been shown empirically that the extension of these methods to projective camera model is not straightforward \cite{park20103d}. 
There are works for projective non-rigid shape and motion recovery based on tensor estimation \cite{hartley2008perspective,vidal2006nonrigid}, but this challenging problem is still under on-going research.
Moreover, the NRSFM methods only recover the shape of the object without absolute translation. 

%In contrast, our approach aims for solving the non-rigid structure from motion problem by leveraging unordered trajectories observed by perspective cameras.
%He \etal \cite{YuchaoDai_CVPR2012} proposed a prior-free solution to the non-rigid structure from motion problem for orthographic cameras which does not require any known temporal relations. However, in order to succeed, they require known ordered trajectories. In contrast, our proposed method does not require ordered trajectories.

%Another related class of methods are the approaches for articulated object reconstruction from monocular videos and images sequences.
%Urtasun \etal \cite{Urtasun_CVPR2006} use Gaussian Process Dynamical Models (GPDMs) to learn human pose and motion priors for 3D people tracking.
%With Bayesian model averaging, a GPDM can be learned from relatively small amounts of data, and it generalizes gracefully to motions outside the training set.
%A popular class of approaches are the particle filter based 3D tracking methods \cite{SminchisescuT_CVPR2003, Sidenbladh_ECCV2000}.

\subsection{Sequencing and synchronization}
Sequencing information is important in trajectory triangulation. Recently,
Basha \etal~\cite{Basha_ECCV2012, Basha_ICCV2013} target the problem of determining the temporal order of a collection of photos without recovering the 3D structure of the dynamic scene. The method in \cite{Basha_ECCV2012} relies on two images taken from roughly the same location to eliminate the uncertainty in the sequencing. Basha \etal~\cite{Basha_ICCV2013} later introduce a solution that leverages the known temporal order of the images within each camera. Both of these methods assume dynamic objects move closely to a straight line within a short time period, but in practice, points can deviate considerably from the linear motion model, especially when the temporal discrepancy between images is large. 

Video synchronization has attracted much attention in the computer vision community \cite{Tuytelaars_CVPR,shrestha2010synchronization,rao2003view}. Those methods have various constraints such as camera motion, availability of sound, and number of videos. While our approach aims at dynamic 3D reconstruction without sequencing, the local temporal order can be recovered as a byproduct of our approach. 

%Tuytelaars \etal~\cite{Tuytelaars_CVPR} propose a method to automatically synchronize two video sequences of the same event. They do not use any constraints from the scene or cameras, but rather rely on point correspondences among the video sequences.

%Unsupervised dictionary learning has also been widely used for reconstructing noisy signals and is closely related to our proposed framework. 
%Along these lines, Mairal \etal~\cite{Mairal_PAMI} show that better results can be obtained when the dictionary is tuned to the specific task.
%In~\cite{Mairal_NIPS} the authors further observe that using sparse coding to the problem of dictionary learning can reduce storage and computational requirements, as well as obviate the need of an explicit learning rate tuning.

%Tuytelaars \etal~\cite{Tuytelaars_CVPR} propose a method to automatically synchronize two video sequences of the same event. They do not use any constraints from the scene or cameras, but rather rely on point correspondences among the video sequences. Wedge \etal \cite{wedge2005_cvpr} propose an efficient coarse-to-fine approach for synchronizing two video sequences recorded at the same frame rate by stationary cameras with fixed internal parameters. In contrast to these methods, we propose a framework for the temporal alignment of multiple video sequences with arbitrary capture characteristics. After briefly discussing related work, we now proceed to provide the notation and intuitions behind our proposed method.

% !TEX root = video3D_l1.tex

\section{Problem and notations} \label{sec:problem_and_notations}

%The goal of the initial spatial registration in our method is to establish camera registration in a common coordinate system. Given the static background structure, we use the publicly available structure from motion tool VisualSFM by Wu for camera registration and calibration.
%Assuming the 2D measures of 3D points are known, we can construct .
%In this paper, we reconstruct the 3D points of the dynamic object using unsynchronized videos.
We now describe the notations of our problem.
Let $\mathcal{I}$ denote an aggregated set of images obtained from $N$ video sequences $\mathcal{V}_n$.
Assuming a total of $F$ available images, we can denote each individual image as $I_f \in \mathcal{I}$, where  $f=1,\dots,F$.
Alternatively, we can refer to  the $m$-th frame in the $n$-th video as $I_{(n,m)} \in \mathcal{V}_n$, where $n=1,\dots,N$ and $m=1,\dots, \left\vert{\mathcal{V}_n}\right\vert$.

%We now discuss the parametrization of  3D structure within our framework.
We assume an {\em a priori} camera registration  through structure-from-motion analysis of  static background structures within the environment~\cite{wu2013towards}.
Accordingly, for each available image  $I_f$ we know the capturing camera's pose matrix
$\mathbf M_f=\left[  \mathbf{R}_f  \; |-\mathbf{R}_f \mathbf{C}_f\right]$,
along with its intrinsic camera matrix $\mathbf K_f$.

Without loss of generality, we first assume each image $I_f$ captures a common set of $P$ 3D points $\{\oneX_{(p,f)}~|~p=1,\dots,P\}$, and the 2D measure of each point is denoted as $\mathbf{x}_{(p,f)}$.
%\hl{(the case with missing measures will be discussed in Sec)}.
We also assume the correspondences of image measures $\mathbf{x}_{(p,f)}$ across images are available. 
Then for each measure $\mathbf x_{(p,f)}$, % with $p \in \{1, \dots, P \}$, 
we can compute a viewing ray with direction by 
$ \mathbf{r}_{(p,f)}=\mathbf{R}_f^{\text{T}} \mathbf{K}^{-1}_f 
 [\mathbf{x}_{(p,f)}^{\text{T}}  \enspace 1]^\text{T}$, and followed with a normalization to a unit vector.

Hence, the position of the dynamic 3D point $\oneX_{(p,f)}$ corresponding to $\mathbf x_{(p,f)}$ can be described by the distance along the viewing ray $\mathbf{r}_{(p,f)}$ given by
\begin{equation}
\oneX_{(p,f)} = \mathbf{C}_f + d_{(p,f)} \mathbf{r}_{(p,f)},
\label{eq:X_representedBy_d}
\end{equation}
where % $\mathbf{C}_f$ is the camera center, $\mathbf{r}_{(f,p)}$ is the viewing directions, and
$d_{(p,f)}$
is the unknown distance of the 3D point from the camera center.

Given $F$ frames with each frame observing $P$ dynamic 3D points, we denote our aggregated observed 3D datum as
\begin{equation}
\label{eq:observationstructure}
\allX =
\begin{bmatrix}
\oneX_{(1,1)} & \cdots & \oneX_{(1,F)} \\
            \vdots  & \ddots & \vdots              \\
\oneX_{(P,1)} & \cdots & \oneX_{(P,F)} \\
\end{bmatrix} =
\left[
\shape_1\; \;  \cdots \; \;   \shape_F
\right],
\end{equation}
where the $f$-th column of the matrix $\allX$, denoted as $\shape_f$, is obtained by stacking all the $P$ 3D points observed in the $f$-th frame.

Then by defining $\allC$, $\allr$, and $\mathbbm{d}$ as follows,
\begin{equation}
\allC =
\begin{bmatrix}
\oneC_{1} & \cdots & \oneC_{F}
\end{bmatrix},
\end{equation}
\begin{equation}
\allr = 
\begin{bmatrix}
\oner_{(1,1)} & \cdots & \oner_{(1,F)} \\
            \vdots  & \ddots & \vdots              \\
\oner_{(P,1)} & \cdots & \oner_{(P,F)} \\
\end{bmatrix},
\end{equation}
\begin{equation}
%\mathbf{d} =
\mathbbm{d} =
\begin{bmatrix}
d_{(1,1)} & \cdots & d_{(1,F)} \\
            \vdots  & \ddots & \vdots              \\
d_{(P,1)} & \cdots & d_{(P,F)} \\
\end{bmatrix},
\end{equation}
Eq.~(\ref{eq:X_representedBy_d}) for all the points can be rewritten in matrix form as
\begin{equation}
\allX = \mathbf{1}_{P\text{x}1} \otimes \allC + 
(\mathbbm{d} \otimes \mathbf{1}_{3\text{x}1}) \odot \allr,
\label{eq:X_representedBy_d_all}
\end{equation}
where $\mathbf{1}_{P\text{x}1}$ is a $P$-by-$1$ matrix with values equal to 1, $\otimes$ is the Kronecker product, and $\odot$ is the component-wise matrix product. %(Hadamard product)

%Since the point $\oneX_{(f,p)}$ has one unknown variable $d_{(f,p)}$, we denote the matrix describing each of the dependent variables associated with $\allX$ as\hl{The use of $\mathbf d$ within our framework will be explained in section} \ref{sec:solver}. 
Our task is to recover  $\allX$ from the 2D measures without image sequencing information across the videos.

% !TEX root = video3D_l1.tex

\section{Principle} \label{sec:principle}
%We first describe the observations motivating our solution before presenting a detailed description of our proposed method for determining 3D structure from a set of unsynchronized videos.
The key observation driving our approach is that dynamic shape exhibits temporal coherence. In this section, we demonstrate how this principle can be leveraged to recover local temporal ordering with known shapes. Our proposed method will extend these ideas to situations with unknown structures.

For our method, we assume a smooth 3D motion under the sampling provided by the videos.
Hence, we can approximate the 3D structure  $\shape_f$ observed in image $f$ in terms of a linear combination of the structures corresponding to the set of immediately preceding ($\shape_{prev})$ and succeeding ($\shape_{next}$) frames in time.
That is, we have
\begin{equation}
\shape_f \approx w \cdot \shape_{prev} + (1-w) \cdot \shape_{next},
\label{eq:linear_comb_2}
\end{equation}
with $0 \leq w \leq 1$.
If our structure matrix $\allX$ from Equation (\ref{eq:observationstructure}) was temporally ordered, which it is not in general, the two neighboring frames would be $\shape_{f-1}$ and $\shape_{f+1}$.
Clearly, such perfect temporal order can be extracted from a single video sequence. However, the reconstructability constraints 
%\cite{valmadre2012general,zheng2014joint}
make single-camera structure estimation ill-posed (see Sec.~\ref{sec:system_condition} for details). Hence, \textit{we rely on inter-sequence temporal ordering information to solve the dynamic structure estimation problem}. The absence of a global temporal ordering requires us to search for temporal adjacency relations across the different video streams having potentially different frame rates. 

% with the constraints $t_{(f,prev)} + t_{(f,next)} = 1$, and $t_{(f,prev)}\geq 0$, $t_{(f,next)} \geq 0$.
%Given the smooth motion assumption $\shape_f$ can approximated by
% $\shape_f = t_{(f,prev)} \shape_{prev} + t_{(f,next)} \shape_{next}$ with the constraints $t_{(f,prev)} + t_{(f,next)} = 1$, and $t_{(f,prev)}\geq 0$, $t_{(f,next)} \geq 0$.

In the most simple scenario, the pool of candidate neighboring frames is comprised by all other frames except $f$.
%Since the objects move continuously and smoothly in the 3D space,
%all 3D points $\mathbf{X}_f$ of $f$th frame can be approximated by a linear combination of the 3D points from the two neighboring frames $f-1$ and $f+1$.
%Considering that the image sequencing information is unknown, we need to search for the neighboring frames in a set of candidate frames.
%If no prior information is presented about the  pool of the candidate frames, we can take all the frames except the current frame $f$ as the pool.
Writing the 3D points of the current frame $\shape_f$ as a linear combination of other frames, we have
\begin{equation}
\shape_f = \allX  \oneT_f,
\end{equation}
where $\oneT_f=\left(w_{(1,f)}, \dotsc, w_{(f-1,f)}, 0, w_{(f+1,f)}, \dotsc, w_{(F,f)}\right)^\text{T}$
is a vector of length $F$ representing the coefficients for the linear combination.
Note that the $f$-th element in $\oneT_f$ equals 0, since the $f$-th column of $\allX$ (corresponding to $\shape_f$) is not used as an element of the linear combination. 

Moreover,
since only a few shapes in the close temporal neighborhood of $\shape_f$ are likely to provide a good approximation, we expect the vector $\oneT_f$ to be sparse.
%Accordingly, we propose to find the most related 3D points through compressive sensing by introducing the $l_1$ norm as follows,
Accordingly, we propose to find the local temporal neighborhood of a shape $\shape_f$ through a compressive sensing formulation leveraging the $l_1$ norm:
\begin{equation}
\underset{\oneT_f}{\text{minimize}} ~ ||\shape_f - \allX \oneT_f||_2^2 + \lambda||\oneT_f||_1,
\label{eq:l1_orig}
\end{equation}
where $\lambda$ is a positive weight. Here, the $l_1$ norm serves as an approximation of the $l_0$ norm and favors the attainment of sparse coefficient vectors $\oneT_f$ \cite{bach2012optimization}.
Moreover,  we incorporate the desired properties of our linear combination framework (Eq.~(\ref{eq:linear_comb_2})) and reformulate Eq.~(\ref{eq:l1_orig}) as
\begin{equation}
\begin{aligned}
& \underset{\oneT_f}{\text{minimize}} && ||\shape_f - \allX\oneT_f||_2^2 \\
& \text{subject to} && \oneT_f \cdot \mathbf 1_{F\times 1}= 1 \\
&                   && \oneT_f \geq 0.
\end{aligned}
\label{eq:l1_new_equal}
\end{equation}
The affine constraints of Eq.~(\ref{eq:l1_new_equal}) constrain the variable $\oneT_f$ to reside in the simplex $\Delta_f$ defined as
\begin{equation}
\Delta_f \triangleq \{ \oneT_f \! \in \! \mathbb{R}^F \text{ s.t. } \oneT_f \! \geq \! 0, w_{(f,f)}=0 \text{ and } \sum_{j=1}^{F} w_{(j,f)} \! = \! 1 \}
\end{equation}

Despite the lack of an explicit $l_1$ norm regularization term in
Eq.~(\ref{eq:l1_new_equal}), as a variant of compressive sensing, the formulation still keeps the sparsity-inducing effect \cite{bach2012optimization,chen:hal-00995911,opac-b1127878}. This is true for the present problem, since we know a shape can be well represented by temporally close shapes. A similar formulation has been used in modeling archetypal analysis for representation learning \cite{chen:hal-00995911}. There, the authors also provide a new efficient solver for this kind of problem.

Finally, we generalize our formulation from Eq.~(\ref{eq:l1_new_equal}) to include all available structure estimates $\shape_f$, with  $f=1,\dotsc,F$, into the following equation
\begin{equation}
\begin{aligned}
& \underset{\allT}{\text{minimize}}&& ~ || \allX - \allX \allT||_\text{F}^2\\
& \text{subject to} && \oneT_f \in \Delta_f, f=1,\cdots,F,
\end{aligned}
\label{eq:l1_given3D}
\end{equation}
where $||\cdot||_\text{F}$ denotes the Frobenius norm and $\allT=[\oneT_1 \dots \oneT_F]$ is an $F\times F$ matrix
with the $f$-th column equal to $\oneT_f$.
By construction, $\allT$ has all its diagonal elements equal to zero.

\begin{figure}
  \centering
  \includegraphics[width=0.48\textwidth]{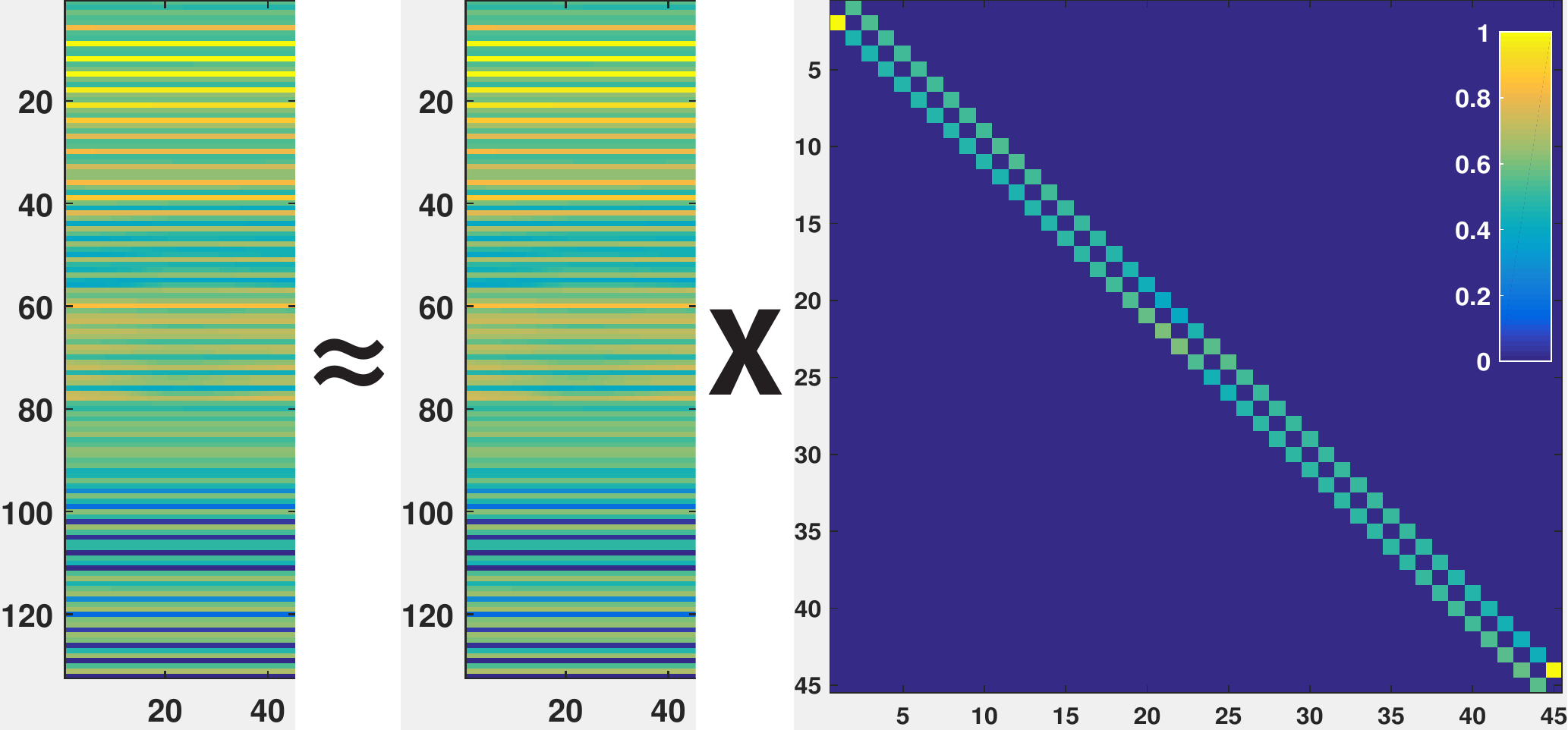}\\
  \caption{We illustrate the output of Eq.~(\ref{eq:l1_given3D})  on a real motion capture dataset. % ``Clap1Rep" presented in \cite{cg-2007-2}. 
For easy visualization, the shortest motion capture dataset (45 frames) presented in \cite{cg-2007-2} is used. Each element/column in $\allX$  corresponds to  ground truth 3D structure. The estimation of $\allT$ through Eq.~(\ref{eq:l1_given3D}) approximates the correct ordering after enforcing all elements in the diagonal to be $0$.}
  \label{fig:principle}
\end{figure}

As an illustration of the validity of our compressed sensing formulation, Fig.~\ref{fig:principle} shows the output of Eq.~(\ref{eq:l1_given3D}) on a real motion capture dataset given known 3D points $\allX$. 
Although image sequencing is assumed unknown, we show results in temporal order for visualization purposes.
The coefficients in $\allT$ approximate a matrix having non-vanishing values only on the locations directly above and below the main diagonal.
This indicates that the 3D points $\shape_f$ are a linear combination of $\shape_{f-1}$ and $\shape_{f+1}$. 

Minimizing Eq.~(\ref{eq:l1_new_equal}) is equivalent to finding the most related shapes to linearly represent $\shape_f$. It is usually true that the temporally close shapes $\shape_{f-1}$ and $\shape_{f}$ are most related, and therefore local temporal information is recoverable from the non-vanishing values in $\allX$. However, if object motion is repetitive or if the object is static for a period of time, there is no guarantee that the most related shapes are the temporally closest ones. Even though this is true, the analysis in Sec.~\ref{sec:shape_approximation} shows that this does not cause any problem for our method in regard to 3D reconstruction. 

To validate our prior of sparse representation for real motion, we quantitatively evaluate the estimated coefficients $\allT$ by minimizing Eq.~(\ref{eq:l1_given3D}) on all 130 real motion capture datasets presented in \cite{cg-2007-2}. 
For a shape at a given time sample, we measure the sum of the two largest estimated coefficient values for this sample, and the frequency with which these top two coefficients correspond to the ground truth temporally neighboring shape samples. Given our prior, values of 1 for both measures are expected. The average values we obtain are 0.9972 and 0.9994, supporting the validity of our prior.

%We note that the self-representation in Eq.~(\ref{eq:l1_given3D}) is previously used in sparse subspace clustering \cite{elhamifar2009sparse}, where the element in each subspace can be sparsely represented by other elements in the same subspace, and the coefficients of sparse coding is used to build a graph for clustering.

% !TEX root = video3D_l1.tex
\section{Method}
\label{sec:method}

We address the problem of estimating sparse dynamic 3D structure from a set of spatially registered video sequences with unknown temporal overlap.
Sec.~\ref{sec:principle} presented a compressive sensing formulation leveraging the self-expressiveness of all the shapes in the context of known 3D geometry.
However, our goal is to estimate the unknown structure without sequencing information.
To this end, we define our dictionary as the temporally varying 3D structure and propose a compressive sensing framework which poses the estimation of 3D structure as a dictionary learning problem.
%Moreover, we define our dictionary as the temporally varying 3D structure, while we define sequencing information in terms of the sparse coefficients describing a locally linear 3D structural interpolation.
We solve this problem in an iterative and alternating manner, where we optimize for 3D structure while fixing the sparse coefficients, and {\em vice versa}.
This is achieved through the optimization of a biconvex cost function that leverages the compressed sensing formulation described in Sec.~\ref{sec:principle}  and, additionally, enforces both structural dependence coherence across video streams and motion smoothness among estimates from common video sources.

\subsection{Cost function}
To achieve the stable estimation of both the structure $\allX$ and the sequencing information $\allT$, we extend our formulation from Equation (\ref{eq:l1_given3D}) to the following cost function:
\begin{equation}
\begin{aligned}
& \underset{\allX, \allT}{\text{minimize}}&& ~ \frac{1}{FP} || \allX - \allX\allT||_\text{F}^2
 + \lambda_1\Psi_1(\allT) + \lambda_2 \Psi_2(\allX) \\
& \text{subject to} && \oneT_f \in \Delta_f, f=1,\cdots,F,
\end{aligned}
\label{eq:ourproblem_final}
\end{equation}
where $\Psi_1(\allT)$ and  $\Psi_2(\allX)$ are two convex cost terms regulating the spatial relationships between 3D observations within and across video streams. We also add the normalization term $FP$ to cancel the influence of number of frames and number of points per shape. Next, we describe each of the cost terms in detail.

\subsection{Dictionary space reduction in self-representation} \label{sec:dictionary_space_reduction}
The first cost term in Eq.~(\ref{eq:ourproblem_final}) serves to find shapes in the dictionary to sparsely represent each shape. The search space can be reduced if some elements of $\allT$ are forced to be 0. As mentioned, the diagonal elements of $\allT$ are forced to be 0, since a shape is not used to represent itself. Moreover, it is possible that if {\em a priori} knowledge of rough temporal information across video steams is available, we can also leverage that knowledge to reduce the search space.

In our solution, we explicitly enforce that the shape observed by one video is not used to represent the shape observed in the same video, because 
the reconstructibility analysis in Sec.~\ref{sec:system_condition} shows such estimation is ill-posed. In our implementation, enforcing this constraint is achieved by not defining the corresponding variables in $\allT$ during the optimization.
 
%represent each shape as a sparse linear combination of other shapes.
% Then describe how to get camera center and viewing ray directions.

\subsection{Coefficient relationships: $\Psi_1(\allT)$}
As described in Sec.~\ref{sec:principle}, a given structure $\shape_f$ in frame $f$ can be obtained from the linear combination of the 3D shapes captured in other frames. The coefficients or weights of the linear combination are given by the elements of the matrix $\allT$. In particular, the element in the $j$-th row and $f$-th column  of $\allT$ is denoted as $w_{(j,f)}$, and it describes the relative contribution (weight) from $\shape_{j}$ in estimating $\shape_{f}$.
Similarly, $w_{(f,j)}$ represents the contribution of $\shape_{f}$ towards the 3D points in $\shape_{j}$.
Accordingly, a value of $w_{(f,j)}=0$ indicates the absence of any contribution from $\shape_{f}$ to $\shape_{j}$, which is desired for tempo-spatially non-proximal 3D shapes.

We note that, if $\shape_{f}$ contributes to $\shape_{j}$, it means the two sets of points are highly correlated, which
 %further means if $\shape_{f}$ contributes to the reconstruction of $\shape_{j}$, then $\shape_{j}$ should reciprocally contribute to estimating $\shape_{f}$.
further implies that  $\shape_{j}$ should reciprocally contribute to estimating $\shape_{f}$.
We deem this reciprocal influence within our estimation process as {\em structural dependence coherence} and develop a cost term that
contributes toward enforcing this property within the estimation of $\allT$.
We encode this relationship into our cost function as an additional term of the form % $\Psi_1$ in Eq.~
\begin{equation}
\Psi_1(\allT) = \frac{1}{F} ||\allT-\allT^\top||_\text{F}^2.
\end{equation}

A strict interpretation of the above formulation aims to identify symmetric matrices.
In general, the reciprocal influence between $\shape_{f}$ and $\shape_{j}$ does not imply symmetric contribution,
as the values of $w_{(f,j)}$ and  $w_{(j,f)}$ depend on the actual 3D motion being observed.
\begin{figure}
  \centering
  \includegraphics[width=0.95\columnwidth]{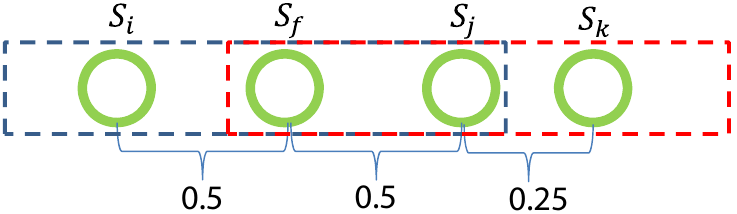}\\
  \caption{Illustration of the triplets influencing the weights for $\shape_f$ and $\shape_j$ leading to an asymmetric $\allT$. The values in the figure represent the distance between adjacent points.}
  \label{fig:Triplet}
\end{figure}
More specifically, these values describe the linear structural  dependencies between two different, but overlapping, 3-tuples of 3D points, \emph{e.g.}, ($\shape_{i}$,$\shape_{f}$,$\shape_{j}$) and ($\shape_{f}$,$\shape_{j}$,$\shape_{k}$) as illustrated in Fig. \ref{fig:Triplet}. In the toy example of Fig. \ref{fig:Triplet}, it can be seen that $\shape_{i}$ and $\shape_{j}$ are at equal distance to $\shape_{f}$ and hence equally contribute to it, \emph{i.e.}, $w_{(i,f)} = w_{(j,f)}=\frac{1}{2}$. However, in order to determine the linear combination weights for specifying $\shape_j$, we need to consider $\shape_f$ and $\shape_k$. Here, $\shape_f$ is twice as far from $\shape_j$ as $\shape_k$, and thus $w_{(f,j)}=\frac{1}{3}$, which is lower than $w_{(j,f)}$.
Accordingly, we do not expect a fully symmetric weight matrix   $\allT$. However,  given our expectation of a sparse coefficient matrix $\allT$, we can focus on finding congruence between the zero-value elements of the $\allT$ and $\allT^\top$, which $\Psi_1(\allT)$ effectively encodes.
Moreover, $\Psi_1(\allT)$  is convex, which enables its use within our biconvex optimization framework.
%But notice that  depending on the object motion, the amount of contributions can be different.

\subsection{Sequencing information:$\Psi_2(\allX)$}
%As mentioned in Sec.~\ref{sec:principle}, while the availability of video sequences enables enforcing constraints among frames attained from the same video, these constraints are insufficient to robustly estimate 3D geometry. 
Under the assumption of sufficiently smooth 3D motion w.r.t.~the frame-rate of each video capture, we define a 3D spatial smoothness term that penalizes large displacements among successive frames from the same video.
Therefore, we define a pairwise term over the values of $\allX$
\begin{equation}
\Psi_2(\allX) = \frac{1}{M} \sum_{n=1}^{N}\sum_{m=1}^{\vert\mathcal V_n\vert-1}\left|\left|\shape_{(n,m)} - \shape_{(n,m+1)} \right| \right|_2^2,
\end{equation}
where $n$ is the video index, $m$ is the image index within a video, $\vert \mathcal V_n\vert$ denotes the number of video frames within each sequence, and $M=\sum_{n=1}^{N}(\vert \mathcal V_n\vert - 1)$ is a normalization factor.
Note that $\Psi_2(\allX)$ does not explicitly enforce ordering information across video sequences, but instead fosters a compact 3D motion path within a sequence.
%However, using this term in conjunction with the explicit inter-sequence dependencies for 3D structure estimation of our model, provides an effective regularization mechanism.
Moreover,  $\Psi_2(\allX)$ is a convex term.

However, this regularization term $\Psi_2(\allX)$ is a double-edged sword.
Since this term minimizes the sum-of-squared distances, if a video camera is static or has small motion, the estimated 3D points are likely to be pulled towards the camera center. 
This typically biases the estimated 3D points slightly away from their real positions. 
Therefore, we propose to first minimize Eq.~(\ref{eq:ourproblem_final}) until convergence to obtain values for $\allX$ and $\allT$, and then taking those values as initialization, we further optimize Eq.~(\ref{eq:ourproblem_final}) setting $\lambda_2=0$, effectively discarding 
$\Psi_2(\allX)$.

%Next we describe how to minimize the cost function (\ref{eq:ourproblem_final}) efficiently.

\section{Parameterization of $\allX$} \label{sec:parameterization}

Eq.~(\ref{eq:X_representedBy_d_all}) explicitly defines the 3D structures $\allX$ to be constrained  to lie on the viewing rays defined by the 2D measurements and camera poses. This corresponds to an implicit assumption of noise-free measurements.
However, 2D feature measurements can be subject to localization inaccuracies or, in extreme cases, detection failure due to image capture aberrations (\emph{e.g.},  motion blur or non-linear camera gain). %The discussion in section \ref{sec:miss_data} addresses the robustness of our framework to missing data. 
Next, we discuss the parameterization of $\allX$ given noisy and missing 2D observations.

%Given accurate 2D measurements, 
%the 3D structures $\allX$ are constrained to lie on the viewing rays defined by the 2D measures and camera poses. Therefore, we can use Eq.~(\ref{eq:X_representedBy_d_all}) to represent $\allX$. This is deemed as a hard constraint, as the points have to lie on the viewing ray. However, in practice, the measures are typically noisy or unavailable due to, for example, inaccurate feature detection or motion blur. 
%Next, we discuss the parameterization of $\allX$ given noisy and missing 2D observations.

%Given camera parameters, viewing ray directions solely depend on 2D measures 
%When using real images, the 2D measures in practice are not perfect. 
%Missing and noisy measures occur due to various reasons such as occlusion, motion blur and inaccurate feature detection.
%Next, we discuss our solutions to these two problems.

%\subsection{Accurate measures}
%When the 2D measures are noise-free, the 3D points $\allX$ are constrained to lie on the viewing rays determined by the camera centers and the 2D measures. Therefore, we can use Eq.~(\ref{eq:X_representedBy_d_all}) to represent $\allX$. This is deemed as a hard constraint as the points have to lie on the viewing ray.

\subsection{Noisy observations} \label{sec:noisy_measure}
The parameterization using Eq.~(\ref{eq:X_representedBy_d}) enforces the hard constraint that 3D points lie on the viewing rays. Given that this may not be appropriate under the circumstance of noisy measurements, we can change this hard constraint to a soft constraint by adding a regularization term into the original Eq.~(\ref{eq:ourproblem_final}). Defining the objective function in Eq.~(\ref{eq:ourproblem_final}) as $\Phi(\allX,\allT)$, we propose a revised version as
\begin{equation}
\begin{aligned}
& \underset{\allX, \allT, \alld}{\text{minimize}} &&\Phi(\allX,\allT) + 
%\lambda_3 ||\mathbf{C} + \mathbf{r} \odot \mathbf{d} - \allX ||_F^2 \\
\lambda_3 ||\mathbf{1}_{P\text{x}1} \otimes \allC + (\mathbbm{d} \otimes \mathbf{1}_{3\text{x}1}) \odot \allr - \allX||_{\text{F}}^2 \\
 & \text{subject to} && \oneT_f \in \Delta_f, f=1,\cdots,F.
\end{aligned}
\label{eq:soft_constraint}
\end{equation}
The formulation converts the hard constraint of Eq.~(\ref{eq:X_representedBy_d_all}) as a soft constraint by adding a penalization if the 3D points deviate from the viewing ray. The value of $\lambda_3$ controls how much a point can deviate away from the viewing ray, and it depends on the noise level of the 2D observations. A larger value of $\lambda_3$ should be used when the level of noise is lower. Note the new formulation is equivalent to the hard constraint if the weight $\lambda_3$ is set to $\infty$. Moreover, in Eq.~(\ref{eq:soft_constraint}), $\alld$ is an auxiliary variable solely depending on $\allX$. More details about the optimization of Eq.~(\ref{eq:soft_constraint}) are presented in Sec.~\ref{sec:optimize_over_x}.
%The main drawback of this formulation is that it adds one more parameter $\lambda_3$.

\subsection{Missing data} \label{sec:miss_data}
Each 3D point, given its accurate 2D measurement, lies on the corresponding viewing ray. Hence, the 3D point has one degree of freedom -- depth along the ray. However, in the absence of 2D observations, which can happen in the case of occlusion, the 3D points are no longer constrained by the viewing ray and thus have three degrees of freedom. 

%To solve this problem in our framework is straightforward. We still minimize the same cost function (Eq.~(\ref{eq:ourproblem_final})), except that the 3D points corresponding to missing measures have three degree of freedom. More specifically, the step of optimization over $\allT$ given 3D points remains the same, but the step of optimization over $\allX$ is changed based on the available measures. optimization over $\allX$ still remains as a quadratic programming and can be easily solved by setting the derivative relative to unknown variables to 0. 

In our method, the 3D points with missing 2D observations are interpolated by the estimated linear coefficients $\allT$. Therefore, this scheme is likely to produce larger errors if a dynamic 3D point is not observed by multiple consecutive frames across time. In our experiments, we test the accuracy of our algorithm under different missing-data rates.

% !TEX root = video3D_l1.tex

\section{Optimization}\label{sec:solver}

The biconvex function in Eq.~(\ref{eq:ourproblem_final}) is non-convex, but it is convex if one set of the variables $\allX$ or $\allT$ is fixed.
%Though a  more complicated dictionary update scheme such as K-SVD \cite{aharon2006img} is possible,in this paper we use the simplest optimization scheme for Eq.~(\ref{eq:ourproblem_final}) that alternates the optimizations over $\allX$ and $\allT$.
The optimization scheme employed for Eq.~(\ref{eq:ourproblem_final}) alternates the optimizations over $\allX$ and $\allT$. We preferred this approach due to its relative simplicity over elaborate dictionary update schemes such as K-SVD \cite{aharon2006img}. Nevertheless, since the alternating optimization steps need to be performed until convergence, each step must be reasonably fast.
Although optimizing over $\allX$ is easy, optimizing over $\allT$ is relatively more difficult due to the simplicial constraint.
We find that optimizing over $\allT$ with a general solver, such as CVX \cite{cvx}, is too slow even for a moderate number of frames $F$.
Moreover, during our iterative optimization, the output of the previous step can be fed into the current step for better initializaiton (hot start), but typical general solvers, such as those based on the interior point algorithm, do not allow for a hot start.
To solve the problem with speed and scalability, we propose a new solver based on alternating direction method of multipliers (ADMM) \cite{boyd2011distributed}.

\subsection{Optimize over $\allX$} \label{sec:optimize_over_x}
If $\allT$ in Eq.~(\ref{eq:ourproblem_final}) is fixed, the optimization over $\allX$ is straightforward, as the problem is quadratic programming without any constraint, regardless of the difficulties discussed in Sec.~\ref{sec:parameterization}.
\begin{enumerate}%[topsep=-1ex,itemsep=-1ex,partopsep=1ex,parsep=1ex]
\item {If the data are noise-free,
we can substitute Eq.~(\ref{eq:X_representedBy_d_all}) into Eq.~(\ref{eq:ourproblem_final}),
and obtain a quadratic programming problem without any constraint on the unknown variable $\alld$. }
\item{
In the case of noisy measurements, $\alld$ are dependent on $\allX$. More specifically, $d_{(p,f)}$ is given by
\begin{equation}
d_{(p,f)}=(\oneX_{(p,f)}-\mathbf{C}_{f})^\text{T}\mathbf{r}_{(p,f)}, 
\end{equation}
\ie~the projection of $\oneX_{(p,f)}-\mathbf{C}_{f}$ onto the viewing ray. Then, after replacing $\alld$ with $\allX$, we obtain a quadratic programming problem over unknown $\allX$.
}
\item{ For the case of missing observations, the corresponding 3D points are unknown variables. Therefore, for a given miss rate, the problem is quadratic over some unknown variables both in $\alld$ and in $\allX$.
}
\end{enumerate}
For the quadratic programming without constraints, the solution can be found at the zero value of the derivative of the cost function over the unknown variables.
%\todo{derivative of what w.r.t. to what?}.

\subsection{Optimize over $\allT$} \label{sec:initlization}
The optimization over $\allT$ is more complex mainly due to the simplex constraints.
By fixing the variable $\allX$ in Eq.~(\ref{eq:ourproblem_final}), the cost function becomes,
\begin{equation}
\begin{aligned}
& \underset{ \allT }{\text{minimize}}&& ~ \frac{1}{FP} || \allX - \allX \allT||_\text{F}^2
 + \frac{\lambda_1}{F} ||\allT - \allT^\top||_2^2 \\
& \text{subject to} && \oneT_f \in \Delta_f, f=1,\cdots,F.
\end{aligned}
\label{eq:ourproblem_TT}
\end{equation}
Notice that if the term $||\allT - \allT^\top||_\text{F}^2$ vanishes, the cost function is the same to Eq.~(\ref{eq:l1_given3D}), which can be decomposed into Eq.~(\ref{eq:l1_new_equal}), and optimized over $\oneT_f$ for each $f=1,\dots,F$ independently.
Advantageously, the number of variables for each subproblem is much smaller compared to the total number of variables in $\allT$, and it can be parallelized on the level of subproblems.
Moreover, Chen \etal \cite{chen:hal-00995911} propose a fast solver to the optimization problem in Eq.~(\ref{eq:l1_new_equal}) based on an active-set algorithm that can benefit from the solution sparsity.
However, the cost term $||\allT - \allT^\top||_\text{F}^2$ prevents the decomposition.

In this paper, we propose an ADMM algorithm that enables the decomposition.
By introducing a new auxiliary variable $\mathbb{Z}$, Eq.~(\ref{eq:ourproblem_TT}) can be rewritten as
\begin{equation}
\begin{aligned}
& \underset{ \allT }{\text{minimize}}&& ~ \frac{1}{FP} || \allX - \allX \allT||_\text{F}^2
 + \frac{\lambda_1}{F} ||\mathbb{Z} - \mathbb{Z}^\top||_\text{F}^2 \\
& \text{subject to} && \oneT_f \in \Delta_f, f=1,\cdots,F \\
&                   && \allT = \mathbb{Z}.
\end{aligned}
\end{equation}
Though this change may seem trivial, the objective function is now separated in $\mathbb{W}$ and $\mathbb{Z}$. 
The ADMM technique allows this problem to be solved approximately by first solving for $\mathbb{W}$ with $\mathbb{Z}$ fixed, then solving for $\mathbb{Z}$ with $\mathbb{W}$ fixed, and next proceeding to update a dual variable $\mathbb{Y}$ (introduced below). This three-step process is repeated until convergence. Next, we describe each step of our ADMM-based algorithm. 

In step 1, $\mathbb{W}$ is updated by
\begin{equation}
\begin{aligned}
\allT^{k+1}=&\underset{\mathbf{T}_f \in{\Delta_f} \text{, for} 1\leq f\leq F}{\text{argmin}} \frac{1}{FP} ||\allX - \allX\allT||_\text{F}^2 \\+ &\text{vec}(\mathbb{Y}^k)^\top \text{vec} (\allT) + \frac{\rho}{2} ||\allT-\mathbb{Z}^k||_\text{F}^2, \\
\end{aligned}
\label{eq:step1}
\end{equation}
where the superscript $k$ is the iteration index. $\mathbb{Y}^k$ is the matrix of  dual variables and is initialized with 0. Note that the values of $\mathbb{Y}^{k}$ and $\mathbb{Z}^{k}$ are known during this step -- we only optimize over the variable $\allT$. The optimization can be decomposed into optimizing over $\oneT_f$ independently and in parallel, and we employ the fast solver proposed in \cite{chen:hal-00995911}. 

In step 2, we update the auxiliary variable $\mathbb{Z}$ according to
\begin{equation}
\begin{aligned}
\mathbb{Z}^{k+1} &= \underset{\mathbb{Z}}{\text{argmin }} \frac{\lambda_1}{F}||\mathbb{Z} - \mathbb{Z}^\top ||_\text{F}^2 - \text{vec}(\mathbb{Y}^k)^\top \text{vec}(\mathbb{Z}) \\
& + \frac{\rho}{2}||\allT^{k+1}-\mathbb{Z}||_\text{F}^2.
\end{aligned}
\label{eq:step2}
\end{equation}
This is a quadratic programming problem in the unknown variable $\mathbb{Z}$ without constraint and can be easily solved by setting the derivative of Eq.~(\ref{eq:step2}) with respect to $\mathbb{Z}$ equal to 0. 

In step 3, the dual variables $\mathbb{Y}$ are updated directly by
\begin{equation}
\mathbb{Y}^{k+1} = \mathbb{Y}^k + \rho(\allT^{k+1} - \mathbb{Z}^{k+1}).
\label{eq:step3}
\end{equation}
The three Eqs.~(\ref{eq:step1}), (\ref{eq:step2}), and (\ref{eq:step3}) iterate until the stop criterion is met. We use the stop criterion described in \cite{boyd2011distributed}.

% !TEX root = video3D_l1.tex

\subsection{Initialization of the Optimization}

\begin{figure}[t]
\centering
\includegraphics[width=0.31\textwidth]{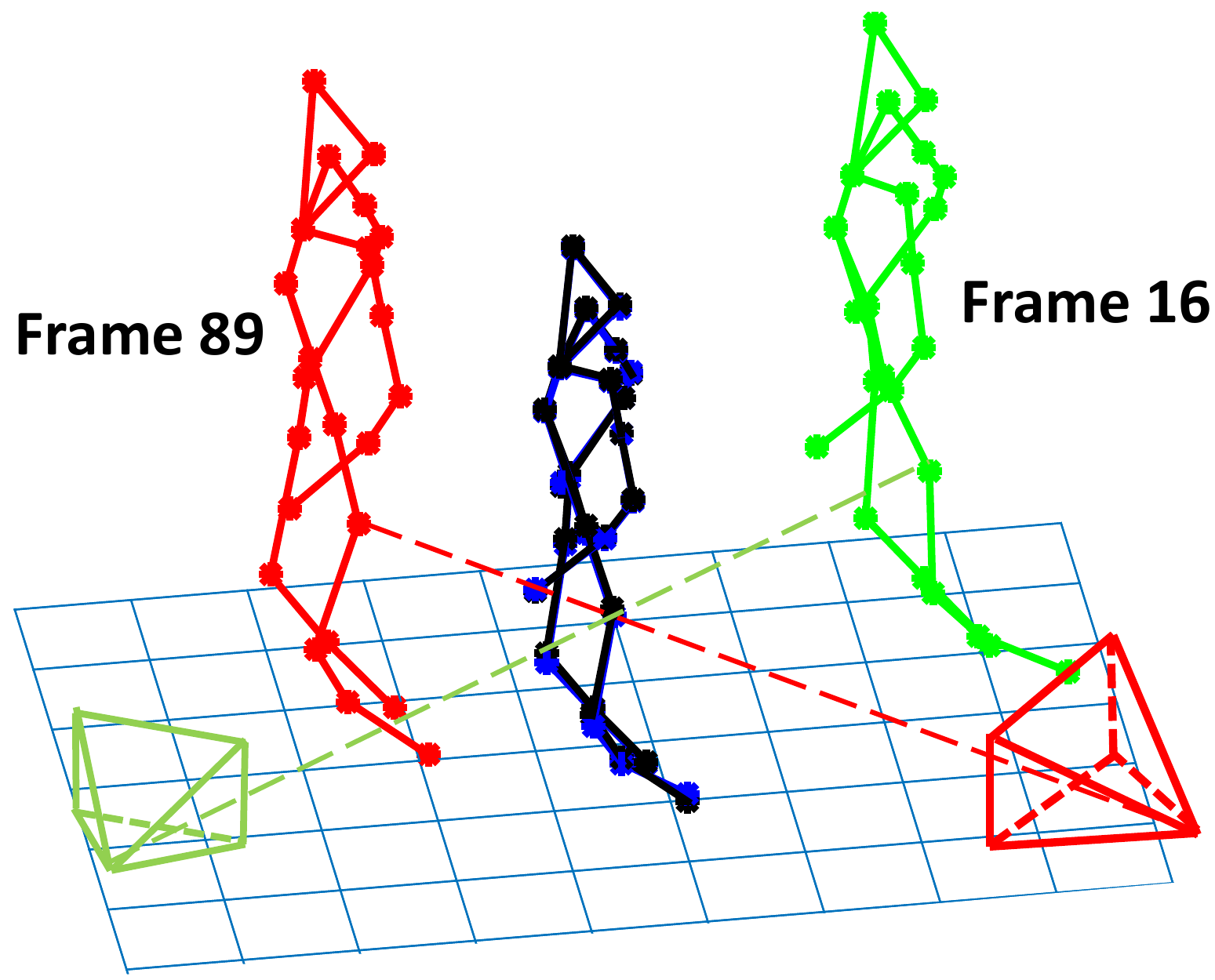}
\caption{Example of incorrect initialization. The dataset `hopBothLegs3hops' in \cite{cg-2007-2} has the motion of hopping forward three times. The black and blue shapes (almost overlapped) are the incorrect initialization of the real shapes (shown in green and red) of frames 16 and 89 due to the accidental ray intersections. This typically happens in the case of  periodic motion such as walking or jogging. In the figure, only one set of nearly intersecting rays is plotted.}
\label{fig:init}
\end{figure}

Given the non-convexity of our original cost function (Eq.~(\ref{eq:ourproblem_final})), the accuracy of our estimates is  sensitive to the initialization values used by our iterative optimization. 
Hence, we design a 3D structure (\ie$\allX$) initialization mechanism aimed at
enhancing the robustness and accelerating the convergence of our biconvex framework.
While our approach explicitly encodes the absence of concurrent 2D observations, we aim to leverage the existence of nearly-incident corresponding viewing rays as a cue for the depth initialization of a given 3D point $\oneX_{(p,f)}$.
To this end, we identify for each bundle of viewing rays captured in $I_f$ (\emph{i.e.}, associated with a given shape structure $\shape_f$) an alternative structure instance captured at $I_j$ that minimizes the Euclidean 3D triangulation error across all corresponding viewing rays. In order to avoid a trivial solution arising from the small-baseline typically associated with consecutive frames of a single video, we restrict our search to ray bundles captured from distinct video sequences.

The position of each point $\oneX_{(p,f)}$ in $\shape_f$ is determined by $d_{(p,f)}$ as in Eq.~(\ref{eq:X_representedBy_d}).
Denoting $\mathbf{d}_f = [d_{(1,f)}, \dots, d_{(P,f)} ]$, we can find the  distance between
shapes of $\shape_f$ and $\shape_j$ by minimizing the following cost function over the unknown variables $\mathbf{d}_f$ and $\mathbf{d}_j$
\begin{equation}
\{\mathbf{d}_f^*,\mathbf{d}_j^*\} =  \underset{\mathbf{d}_f,\mathbf{d}_j } {\text{argmin}} ||\shape_f - \shape_j ||_2^2.
\label{eq:init}
\end{equation}
This is a quadratic cost function with a closed-form solution. % corresponding to the intersection points of each corresponding pair of viewing rays and their common normal.

We then build a symmetric distance matrix $\mathbf{D}$ with element $D_{(f,j)}$ equal to the minimum cost of Eq.~(\ref{eq:init}).
If the frames $f$ and $j$ are from the same video, $D_{(f,j)}$ is set to infinity.
%Next, we enforce cheirality constraints on the depth values, as we find many pseudo-intersection points with negative ray depth for spatio-temporally distant pairs of viewing rays.
Next, we identify many pseudo-intersection points with negative depth (\ie divergent pairs of viewing rays), and set the corresponding element in $\mathbf{D}$ to infinity.
Finally, we determine the minimum element of each $f$-th row in our distance matrix $\mathbf{D}$ and assign the corresponding depth values $\mathbf d^*_f$ as our initialization for the definition of our 3D structure $\mathbf{S}_f$.

The above initialization is done regardless of available measurements, since we only look for an approximate initialization for the solver. In the case of missing data, the corresponding 3D points in the shape are simply ignored when minimizing Eq.~(\ref{eq:init}).

The output of the initialization is typically close to the ground truth, but may fail occasionally, as is shown in Fig.~\ref{fig:init}. This kind of wrong initialization may lead to wrong estimation of the two shapes if the smoothness term $\Phi_2(\allX)$ in Eq.~(\ref{eq:ourproblem_final}) is not present, because these two shapes can well represent each other. Our cost term $\Phi_2(\allX)$ helps to pull the occasional incorrect shapes out of local minima.
%If the wrong initialization in this  example is used and the cost term $\Phi_2(\allX)$ in Eq.~(\ref{eq:ourproblem_final}) is not present, it is very likely the shapes of frame 16 and 89 will be wrongly estimated in the end, because these two shapes can well represent each other. Our cost term $\Phi_2(\allX)$ helps pulling the occasional wrong shapes out of local minima.

\section{Analysis and discussion}	\label{sec:reconstructability}

%The reconstructability tells in which circumstances the algorithm \ref{eq:ourproblem_final} is able to recover the dynamic structure close to the ground truth. 
%From the analysis, it can be shown the importance of choosing neighboring shape associated with another video stream.
%The reconstructability of the algorithm describes the reconstruction accuracy under different capture scenario. 
%This sections starts from reconstructability analysis, 
%from where it is revealed why in our method the shape observed by one video is not used to represent the shape observed in the same video. 
%Moreover, we describes 
%Why sequencing information is not required in our method, while it is vital for all the other trajectory triangulation methods.
This section provides key insights to our algorithm for dynamic object reconstruction without sequencing. %The following important statements will be explained at length.
The following statements will be illustrated in detail.
\begin{enumerate}
\item %To obtain good reconstructability, a shape observed by one video should not be used to represent the shape observed in the same video. 
Interleaved 2D measures across video streams yields favorable viewing ray geometry for 3D shape estimation.
\item %To achieve accurate results, a 3D shape should be well approximated by other shapes.
High-frequency 2D observations and smooth object motion jointly validate our self-expressive structure prior for accurate shape estimation.
\item %While sequencing information is vital in other trajectory triangulation methods for structure estimation \cite{park20103d,valmadre2012general}, our method does not require it.
No dependence on the availablity of sequencing information as opposed to existing approaches \cite{park20103d,valmadre2012general}.
\end{enumerate} 
Next, we first describe the formulation of reconstruction errors by our method, based on which the above statements are illustrated at length in the subsequent three subsections.

\begin{figure*}[t]
\centering
\subfloat[System condition: 1.48e+04]{
\includegraphics[width=0.315\textwidth]{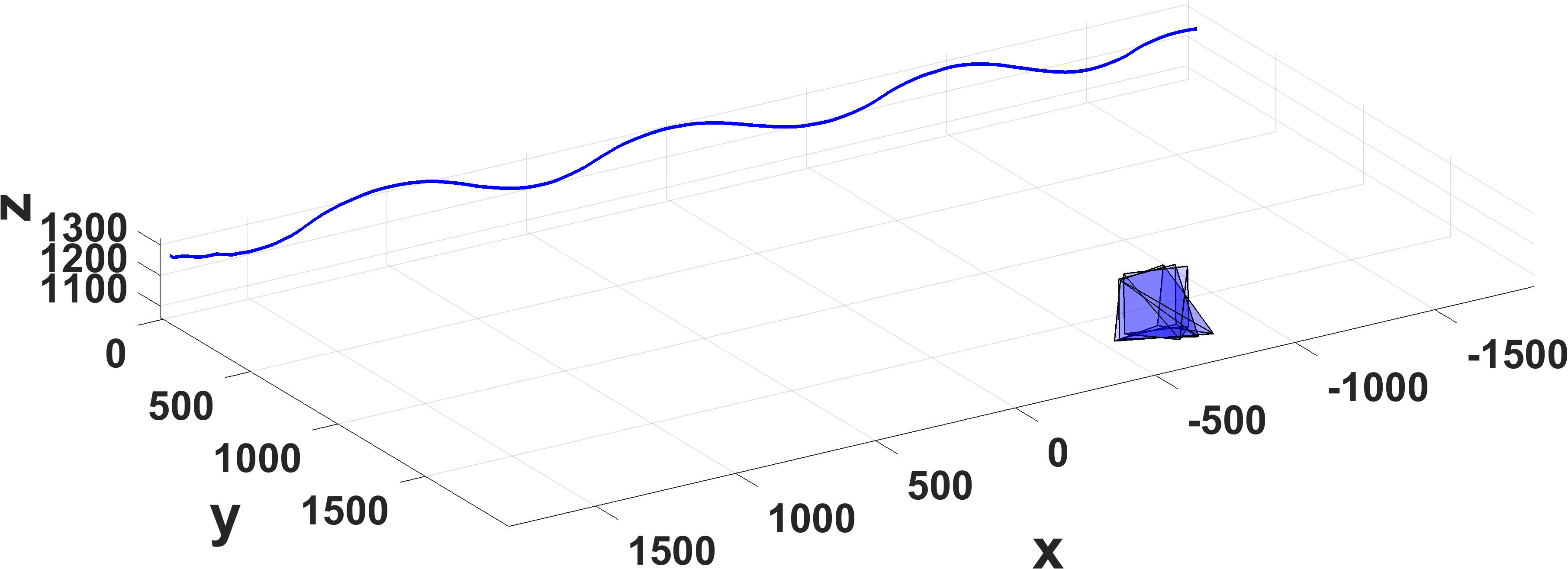}
\label{fig:oneCam}
}
\subfloat[System condition: 22.3]{
\includegraphics[width=0.315\textwidth]{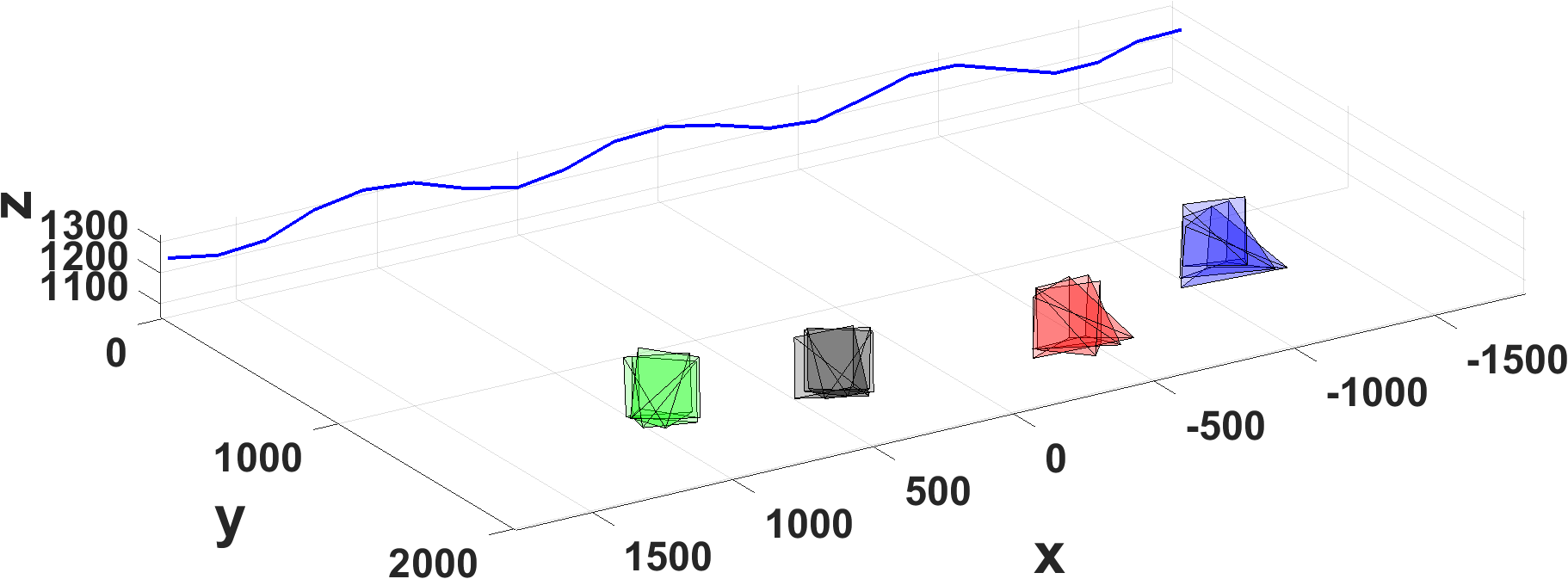}
\label{fig:multiCam}
}
\subfloat[System condition: 29.0]{
\includegraphics[width=0.315\textwidth]{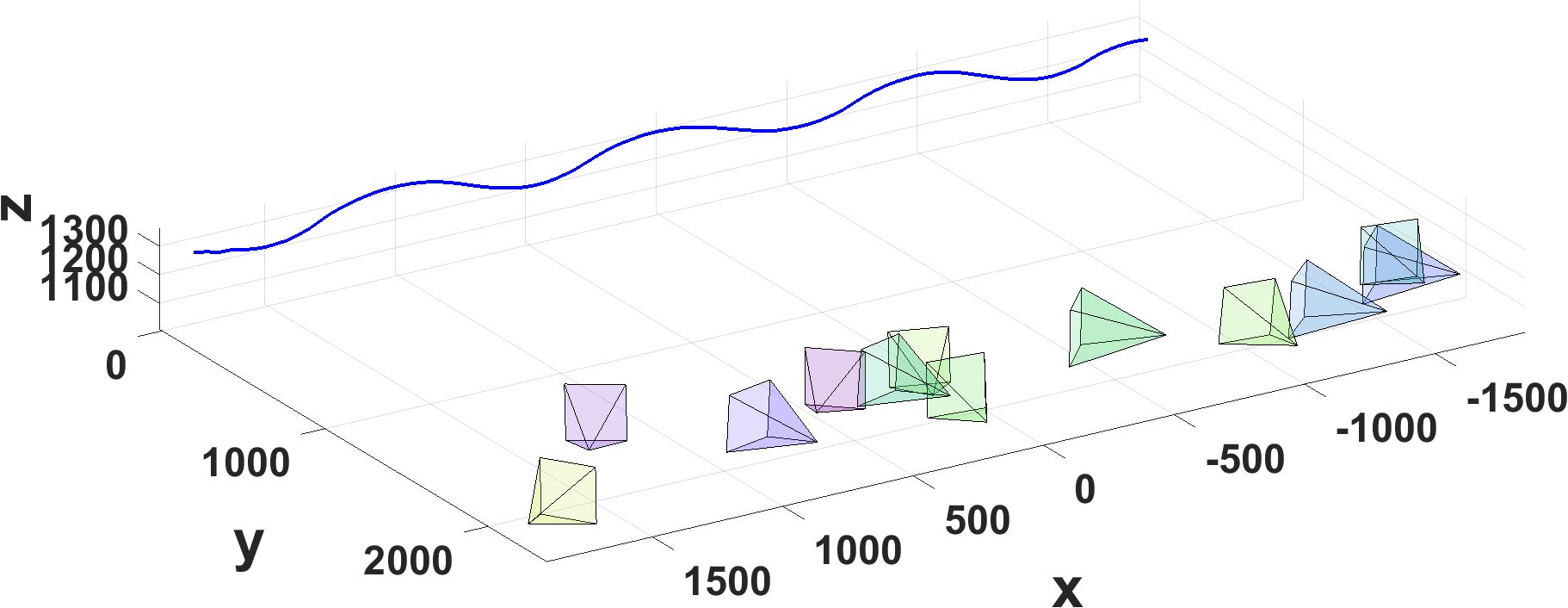}
\label{fig:randCam}
}
\caption{Simulated camera setups. The blue curve is a trajectory of a 3D point obtained from  motion capture data. Figs.~\ref{fig:oneCam} and \ref{fig:multiCam} depict the camera setups of  one and four slow-moving handheld cameras. Fig.~\ref{fig:randCam} depicts a scenario where each random camera only captures one image. Fig.~\ref{fig:multiCam} and Fig.~\ref{fig:randCam} show the camera setups used in our method and \cite{zheng2014joint}, respectively. Coordinates are in millimeters (mm).
}
\label{fig:condition}
\end{figure*}

\subsection{Representation of Reconstruction Errors}
Our solution computes 3D structure by minimizing the non-convex function Eq.~(\ref{eq:ourproblem_final}).
Since direct analysis of the non-convex function is difficult, we only analyze the problem with the assumption that the ground truth of $\allT$, which is defined as the output of Eq.~(\ref{eq:ourproblem_final}) given ground truth structure, is already known. 
%Our analysis reveals the effectiveness of our approach if a good $\allT$ is given. 
Without loss of generality, we also assume the 2D observations are noise-free.

Given that in our method $\lambda_2$ is set to 0 in the end, and $\allT$ is known and fixed, Eq.~(\ref{eq:ourproblem_final}) is equivalent to
\begin{equation}
\begin{aligned}
& \underset{\allX}{\text{minimize}}&& ~ || \allX - \allX\allT||_\text{F}^2 . 
\end{aligned}
\label{eq:recon_original}
\end{equation}
From Eq.~(\ref{eq:recon_original}), it can be seen when $\allT$ is fixed, all points in a shape are computed independently, and computing one 3D point per shape versus multiple points per shape basically follows the same routine. 
Therefore, for the sake of more concise presentation, the analysis in this section assumes only one point per shape, and the point index $p$ for the shape is omitted.

To analyze the reconstruction error, we assume that the ground truth of the 3D points is already known, and then analyze how much the computed structure deviates away from the ground truth, which is deemed as reconstruction error.
We denote the ground truth 3D point as
$\allX^*=[\oneX_1^*, \cdots, \oneX_f^*,  \cdots, \oneX_F^*]$. 
Then, any point $\oneX_f$ on the viewing ray that passes through $\oneX^*_f$ can be parameterized as 
\begin{equation}
\oneX_f = \oneX^*_f + l_f \mathbf{r}_f,
\label{eq:new_line_represent}
\end{equation}
where the unknown $l_f$ is the signed distance from the ground truth along the viewing ray. 

When minimizing Eq.~(\ref{eq:recon_original}), using either Eq.~(\ref{eq:new_line_represent}) or Eq.~(\ref{eq:X_representedBy_d}) to represent $\oneX_f$ in practice generates different values of $d_f$ and $l_f$, but the estimated 3D points are actually identical.
Therefore, $|l_f|$ represents the Euclidean error of our  method. 
%It means if $l_f$ equals 0, the dynamic 3D point is ideally reconstructed.

Eq.~(\ref{eq:recon_original}) is a quadratic objective function without any constraint and has a closed-form solution. We use Eq.~(\ref{eq:new_line_represent}) to represent the 3D point, and by setting the derivative of Eq.~(\ref{eq:recon_original}) over variables $\mathbf{l} = [l_1,\dots,l_f,\cdots,l_F ]$ to 0, we obtain a linear equation system denoted as
\begin{equation}
\mathbf{A}\mathbf{l} = \mathbf{b},	\label{eq:alb}
\end{equation}
where $\mathbf{A}$ is an $F \times F$ matrix with the $f$-th row given by
\begin{equation}
\mathbf{A}_{:f} = (\mathbf{I}-\allT)_{:f}(\mathbf{I}-\allT)^\text{T} \text{diag}([\mathbf{r}^T_1 \mathbf{r}_f, \cdots, \mathbf{r}_F^T \mathbf{r}_f] ),
\label{eq:A}
\end{equation}
and $\mathbf{b}$ is an $F \times 1$ vector with the $f$-th element given by
\begin{equation}
%\mathbf{b}_{f} = \mathbf{r}_f^T[X^*_1,\dots,X^*_F] (\mathbf{I}-\allT) (I-\allT)_{:f}.
\mathbf{b}_{f} = \mathbf{r}_f^T \allX^* (\mathbf{I}-\allT) (I-\allT)_{:f}^{\text{T}}.
\label{eq:b}
\end{equation}
In Eqs.~(\ref{eq:A}) and (\ref{eq:b}), the subscript $_{:f}$ denotes the $f$-th row of a matrix, and $\mathbf{I}$ is an identity matrix. Then the solution for $\mathbf{l}$ is
\begin{equation}
\mathbf{l} = \mathbf{A}^{-1}\mathbf{b}. 
\label{eq:a_minus_1_b}
\end{equation}
As mentioned, $\mathbf{l}$ is the reconstruction error, 
which is bounded by 
\begin{equation}
||\mathbf{l}||_2= ||\mathbf{A}^{-1} \mathbf{b}||_2 \leq ||\mathbf{A}^{-1}||_2 ||\mathbf{b}||_2.\label{eq:upbound}
\end{equation}
A large upbound in Eq.~(\ref{eq:upbound}) means unstable reconstruction results and typically larger errors.
In this paper, we use the term reconstructability (first defined in \cite{park20103d}) as a criterion to characterize the reconstruction accuracy of our algorithm. 
In order to achieve high reconstructability, $||\mathbf{A}^{-1}||_2$ and $||\mathbf{b}||_2$ should be small. Next, we discuss $||\mathbf{A}^{-1}||_2$ and $||\mathbf{b}||_2$ in detail.

%%%%%%%%%%%%%%%%%%%%%%%%%%%%%%%%%%%%%%%%%%%%%%%%%%%%%%%%%%%%%%%%%%%%%%%%%%%%%%%%

%%%%%%%%%%%%%%%%%%%%%%%%%%%%%%%%%%%%%%%%%%%%%%%%%%%%%%%%%%%%%%%%%%%%%%%%%%%%%%%%
\subsection{System condition} \label{sec:system_condition}

\begin{figure}[b]
\centering
\subfloat[Observation without overlap]{
\includegraphics[height=0.16\textwidth]{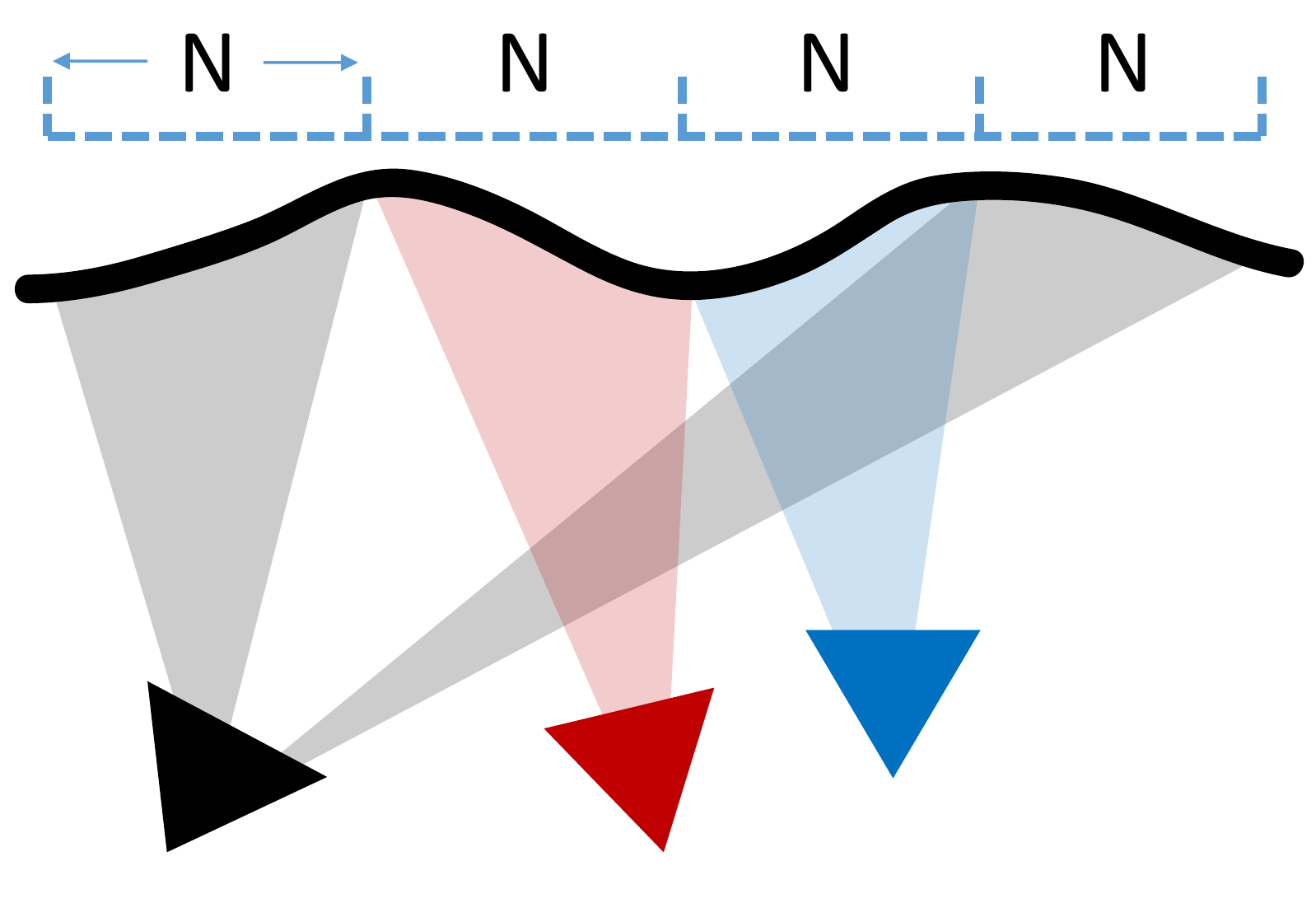}
\label{fig:frameDist}
}
\subfloat[System condition]{
\includegraphics[height=0.16\textwidth]{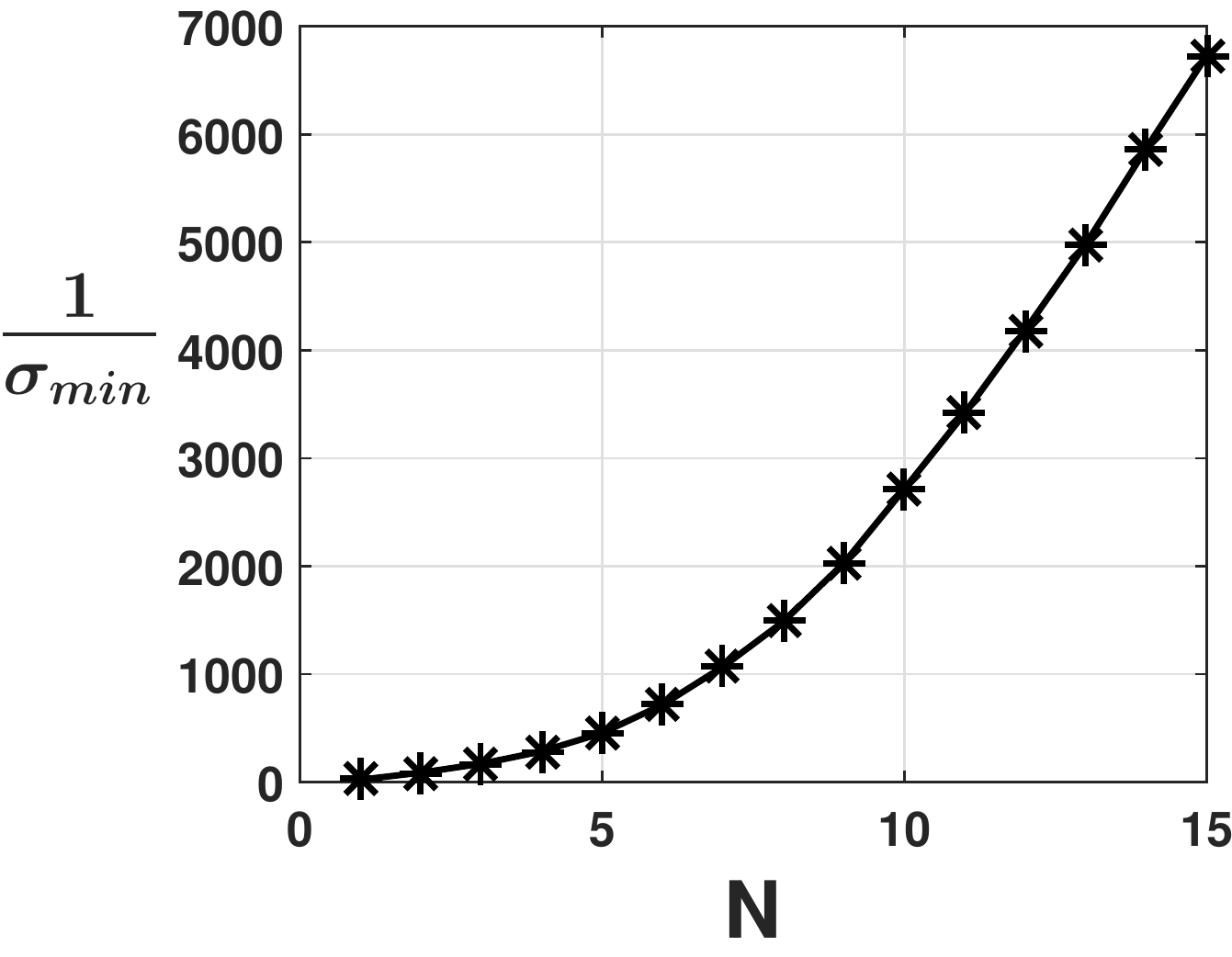}
\label{fig:frameDist_cond}
}
\caption{The reconstructability of the system is lower if the period of single-camera capture is longer.}
\end{figure}

Based on the definition of the matrix Euclidean norm, we have
\begin{equation}
||\mathbf{A}^{-1}||_2 = 1/\sigma_{\text{min}}, 
\label{eq:sysCondition}
\end{equation}
where $\sigma_{\text{min}}$ is the smallest singular value of matrix $\mathbf{A}$. 
With fixed $\allT$, we observe from Eq.~(\ref{eq:A}) that $\mathbf{A}$ solely relies on the viewing ray directions and does not depend on the exact positions of the 3D points $\allX^*$ along the viewing rays.
Since $\sigma_{\text{min}}$ is closely related to reconstruction errors and is determined by the camera system setup, we call it system condition. 
Note the system condition introduced here is in essence very similar to the system condition number described in the works \cite{valmadre2012general,zheng2014joint}.

Since direct analysis of the system condition given viewing ray directions $\{\mathbf{r}_1,\dots,\mathbf{r}_F\}$ based on Eq.~(\ref{eq:A}) is difficult, we next use empirical simulation to demonstrate the system condition under different camera setups. 
In the experiments, we synthesize 2D input features from real
motion capture datasets that sample the 3D structure of real dynamic objects at 40 Hz. Figs.~\ref{fig:oneCam} and \ref{fig:multiCam} simulate setups of one handheld camera and multiple handheld cameras that record videos of a person walking. To mimic small random motion in each handheld camera, 
the camera centers at different time instances are perturbed by Gaussian noise with standard deviation of 10 mm around a fixed center. 
We also test the case of completely random cameras (Fig.~\ref{fig:randCam}), with each taking one photo. 
The 3D structure at each time instance is projected to one of the virtual cameras to generate a set of 2D observations. For the scenario in Fig.~\ref{fig:multiCam}, we ensure no two shapes at consecutive time instances are projected into the same video stream.

We estimate the system condition using Eq.~(\ref{eq:sysCondition}) on 500 trials with random cameras. 
%The average system conditions for the cases of Figs.~\ref{fig:oneCam}, \ref{fig:multiCam}, and \ref{fig:randCam} are 1.48e+04, 22.3, and 29.0 respectively. 
The average system conditions in Fig.~\ref{fig:condition}
show the setup with one handheld camera has very low reconstructability.
Note that even though the system conditions of the camera setups in Figs.~\ref{fig:multiCam} and \ref{fig:randCam} are favorable, in practice the important sequencing information (see Sec.~\ref{sec:importance_of_image_sequencing}) across different cameras for these two cases is not readily available.

To illustrate the importance of cross-sequence 2D observations for our structure estimation process (statement 1), we evaluate system condition as a function of increased temporal gaps between cross-sequence samples. As shown in Fig.~\ref{fig:frameDist}, the dynamic object is observed by one camera for $N$ frames, and then observed by another camera for $N$ frames. We show empirically that as $N$ increases, the system condition increases monotonically, which indicates more unstable reconstruction and typically larger errors (see experiments in Sec.~\ref{sec:comparison_to_other_methods}), even under the assumption that $\allT$ can be correctly estimated. This also illustrates that temporally consecutive shapes observed by the same video stream should not be used to represent each other, as is done in Sec.~\ref{sec:dictionary_space_reduction}.

In fact, we observe that the reconstructability is closely related to the camera motion and the object motion. 
Specifically, if shape $\mathbf{S}_j$ is the most related shape to $\mathbf{S}_f$, as indicated by $\allT$, the relative directions of viewing rays $\mathbf{r}_f$ and $\mathbf{r}_j$ (note we only have one point per shape in this analysis), determine the reconstructability. If the directions of $\mathbf{r}_f$ and $\mathbf{r}_j$ converge, \ie the camera motion is relatively larger than the object motion, the reconstructability is higher. 
In the case of one handheld camera, the camera motion can be much smaller than the dynamic objects, and the viewing rays diverge, yielding low reconstructability. In contrast, if $\mathbf{r}_j$ and $\mathbf{r}_f$ are associated with different video cameras, the distance between the camera centers is much larger than the motion of the object. Hence the reconstructability is high.
This observation is analogous to the classic triangulation of static scenes, where small baselines produce inaccurate reconstruction. Note the same conclusion was also made by Park \etal~\cite{park20153d}, though their reconstruction algorithm is different from ours.

%%%%%%%%%%%%%%%%%%%%%%%%%%%%%%%%%%%%%%%%%%%%%%%%%%%%%%%%%%%%%%%%%%%%%%%%%%%%%%%%%%%%

%%%%%%%%%%%%%%%%%%%%%%%%%%%%%%%%%%%%%%%%%%%%%%%%%%%%%%%%%%%%%%%%%%%%%%%%%%%%%%%%%%%%
\subsection{Shape approximation residual} \label{sec:shape_approximation}
\begin{figure}
\centering
\includegraphics[width=0.28\textwidth]{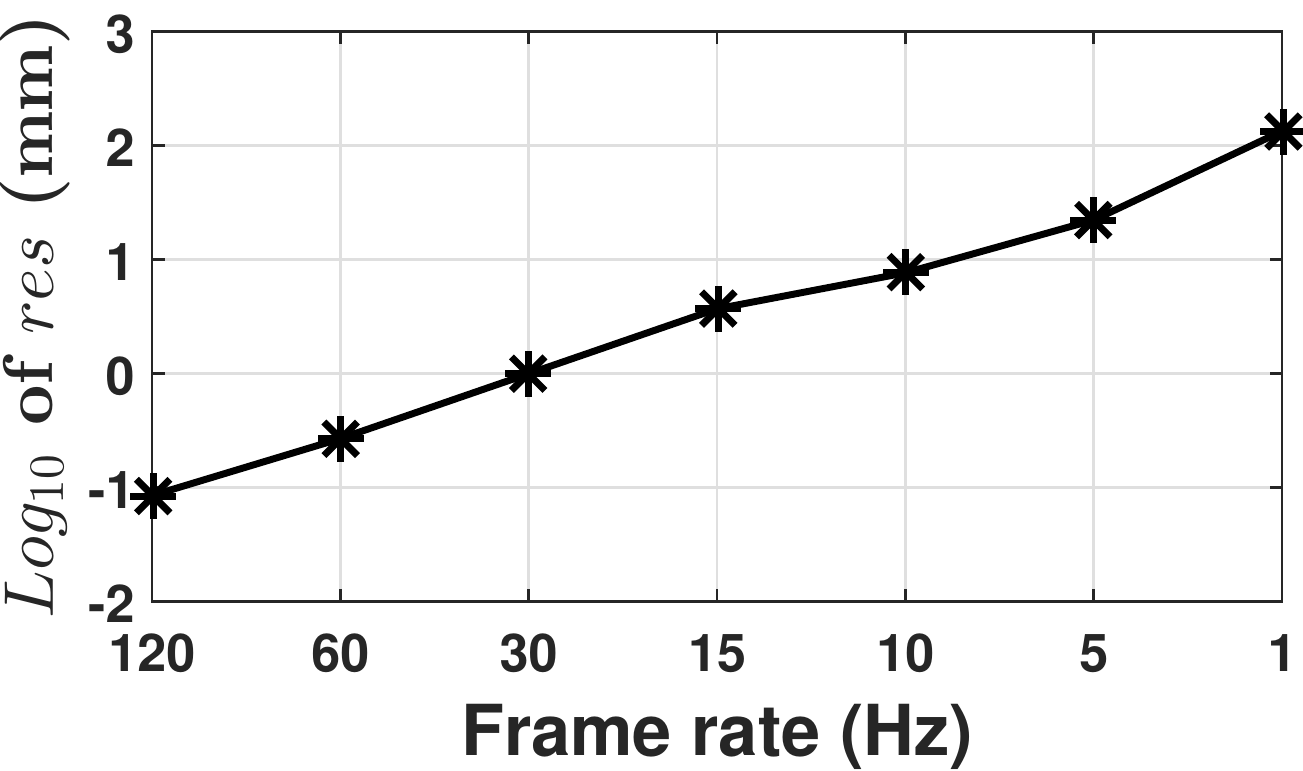}
\caption{Average residuals $res$ at different camera frame rates. Results are attained from 130 motion capture datasets in \cite{cg-2007-2}.}
\label{fig:residual}
\end{figure}

While $\mathbf{A}$ depends on the viewing ray directions, which are available before reconstruction, $\mathbf{b}$ relies on the actual unknown positions of the ground truth structure $\allX^*$ (Eq.~(\ref{eq:b})). To achieve accurate reconstruction, each value in the vector $\mathbf{b}$ should be close to 0. 

Since in Eq.~(\ref{eq:b}), $(I-\allT)_{:f}^\text{T}$ is sparse, $\mathbf{b}_f$ can be considered as a linear combination of a few columns of matrix $\allX^*(I-\allT)$  multiplied using dot product with the unit vector $\mathbf{r}_f$.
Therefore, the value of $\mathbf{b}_f$ mainly relies on $||\allX^*(I-\allT)||_\text{F}$.
Accordingly, we define the residual per point as
\begin{equation}
res = \frac{1}{PF}||\allX^*(I-\allT)||_\text{F}. \label{eq:residual}
\end{equation}
The residual $res$ is small if all the shapes can be well represented by other shapes. It relies on speed of object motion and the capturing frame rate. 
We test the residual $res$ given motion capture data sampled at different frame rates. Fig.~\ref{fig:residual} shows $res$ becomes larger as the frame rate goes down. This fits the intuition that shapes that are tempo-spatially farther away are less correlated. This also implies that our method cannot achieve accurate reconstruction from discrete images with large temporal discrepancy.

%%%%%%%%%%%%%%%%%%%%%%%%%%%%%%%%%%%%%%%%%%%%%%%%%%%%%%%%%%%%%%%%%%%%%%%%%%%%%%%%%%%

%%%%%%%%%%%%%%%%%%%%%%%%%%%%%%%%%%%%%%%%%%%%%%%%%%%%%%%%%%%%%%%%%%%%%%%%%%%%%%%%%%%
\subsection{Importance of image sequencing} \label{sec:importance_of_image_sequencing}

The temporal order of images, \ie image sequencing, plays an important role in dynamic object reconstruction \cite{park20103d,valmadre2012general}. The work by Valmadre \etal~\cite{valmadre2012general} generalizes the method of \cite{park20103d} in a new framework based on high-pass filters. Here, we briefly describe the method in \cite{valmadre2012general} and its relation to our method, from which it can be revealed why their methods \cite{park20103d,valmadre2012general} require sequencing information as opposed to ours. 

Assuming the object moves smoothly in the space, Valmadre \etal~\cite{valmadre2012general} triangulate the 3D trajectory of an 3D point by minimizing its response to a set of high-pass filters.
Given a predefined high pass filter $\mathbf{g}=[g_M,\dots,g_1]$, the trajectory is estimated by 
\begin{equation}
\underset{\allX}{\text{minimize}} ~ || \allX \mathbf{G}||_\text{F}^2,
\label{eq:valmadre_equ}
\end{equation}
where $\mathbf{G}$ is defined as
\begin{equation}
\mathbf{G} = \left[ 
\begin{matrix}
g_M  	&  		 &   	  	\\
\vdots	& \ddots &    	  	\\
g_1		& \ddots & g_M	  	\\
		& \ddots & \ddots 	\\
		&		 & g_1	
\end{matrix}
\right].
\label{eq:valmadre_filter_G}
\end{equation}
Each column of $\mathbf{G}$ is a high-pass filter for the local region of a smooth trajectory. From the method model, it requires all the shapes (columns of $\allX$) and hence the 2D meansurements to be ordered temporally. 

Comparing Eq.~(\ref{eq:valmadre_equ}) with Eq.~(\ref{eq:recon_original}), we can see the two equations are the same if $\mathbf{G}$ equals $\mathbf{I}-\allT$.
In effect, the method in \cite{valmadre2012general} can be regarded as our method with a predefined $\allT$.
%it can be shown $\mathbf{G}$ is equivalent to our method with a predefined value of $\allT$, under the assumption that all the columns of $\allX$ are ordered in temporal order. 
For instance,  if the high pass filter is set to $\mathbf{g}=[1,-1]$, it is equivalent (ignoring the difference at boundary) that $\allT$ is set to
\begin{equation}
\allT = \left[
\begin{matrix}
0 &   & \\
1 & 0 & \\[-7pt]
  &	1 & \ddots \\[-7pt]
  &   & \ddots 
\end{matrix}
\right].
\end{equation}
Therefore, an alternative interpretation of their method \cite{valmadre2012general} using the high-pass filter $\mathbf{g}=[1,-1]$ in terms of our theory is approximating the current shape using only the temporally closest shape. 
Another high-pass filter proposed in \cite{valmadre2012general} is $\mathbf{g}=[-1, 2, -1]$, which in our case is equivalent to fixing the weights of two neighboring shapes to 0.5. %as follows
%\begin{equation}
%\allT = \left[
%\begin{matrix}	  
%	0.5   &      & \\
%	0   & 0.5  & \\[-7pt]
%	0.5 & 0    & \ddots \\[-7pt]
% 	    & 0.5  & \ddots \\[-7pt]
%  		&      & \ddots 
%\end{matrix}
%\right]
%\end{equation}
%In effect, their method can be deemed as our method with predefined $\allT$. 

The importance of sequencing can also be revealed from analysis of residual defined by Eq.~(\ref{eq:residual}). For the method in \cite{valmadre2012general} with predefined $\mathbf{G}$, the residual will be large if columns of $\allX^*$ are randomly shuffled. In contrast, our method leverages compressive sensing to estimates $\allT$ (instead of predefined), which automatically picks the most related shapes to produce small residuals.
%In Eq.~(\ref{eq:b}), $\mathbf{b}_f$ is bounded by residual $r = ||\allX^*(I-\allT)||_{\text{F}}$. 
%If columns of $\allX$ is in temporal order, which means the shape of neighboring columns are spatially close,which is likely to produce small $\mathbf{R}$ with a predefined high pass filter. However, this is not true with shuffled columns of $\allX$. 

% !TEX root = video3D_l1.tex

\begin{figure*}
\centering
  \begin{minipage}[c]{0.35\linewidth}
    \centering
    \includegraphics[width=0.85\linewidth]{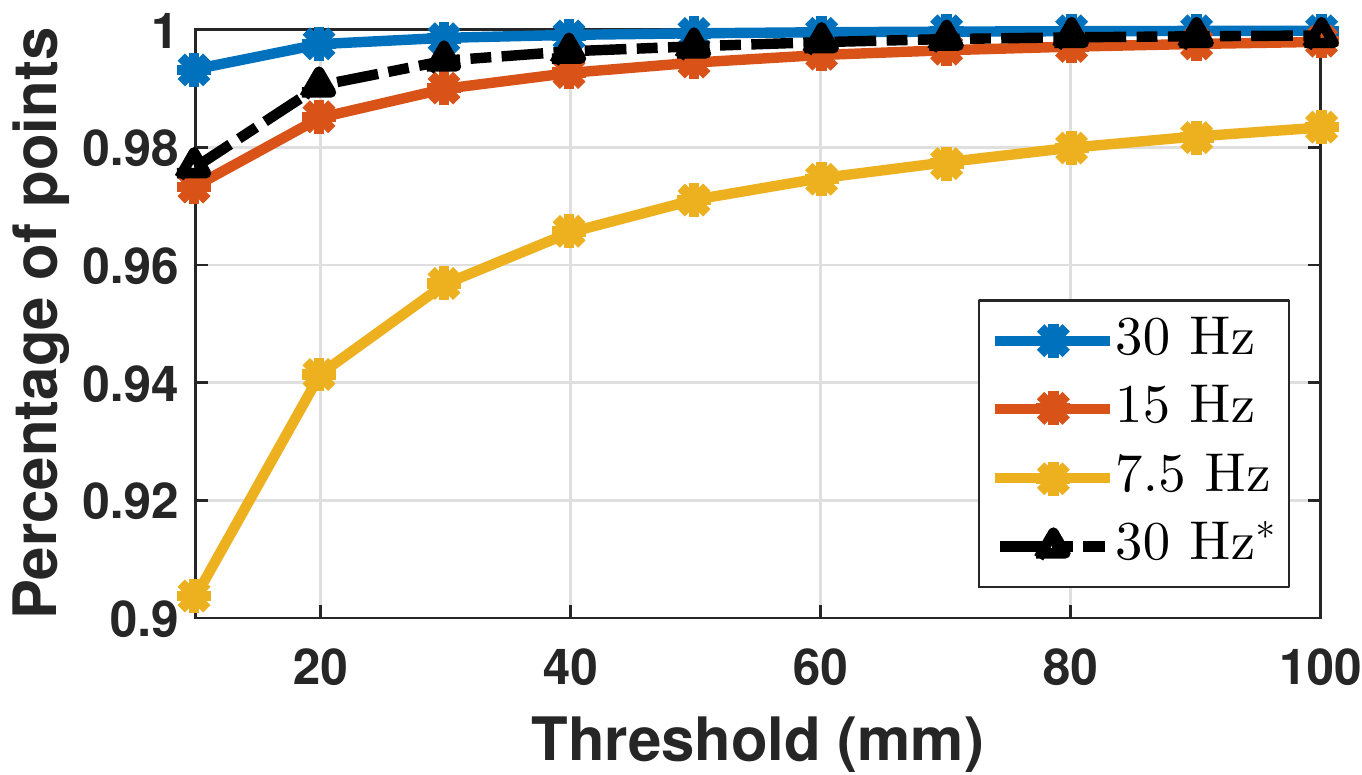}  
  \end{minipage}%
  \begin{minipage}[c]{0.65\linewidth}
    \centering
 \begin{tabular}[b]{|c|*{6}{c|}}
	\hline
  \backslashbox{Frame rate\kern-1em}{\kern-1emThreshold}
	& {10} & {20} & {30} & {40} & {50} & {100}\\\hline
	{30}  & 0.9933 &   0.9975  &  0.9986  &  0.9991  &  0.9994  &  0.9998\\
	\hline
	{15}  &  0.9734  &  0.9850  &  0.9899  &  0.9926 &   0.9944  &  0.9979\\
	\hline
	{7.5}  &  0.9036  &  0.9415  &  0.9568 &   0.9655 &   0.9711 &  0.9833\\
	\hline	
	\hline
	\begin{tabular}{@{}c@{}} $30^*$ (unconstrained \\ assignment) \end{tabular} & 0.9766  &  0.9905 &   0.9947 &   0.9963  &  0.9971 &   0.9990\\
	\hline	
\end{tabular}
\end{minipage}
\caption{The reconstruction accuracy given different camera frame rates. We also test the case that the captures of object motion are randomly assigned to any of the image sequences without any constraint. 30 Hz$^*$ in the figure represents the unconstrained assignment.}
\label{fig:error_framerate}
\end{figure*}

\begin{figure*}
\centering
  \begin{minipage}[c]{0.35\linewidth}
    \centering
    \includegraphics[width=0.85\linewidth]{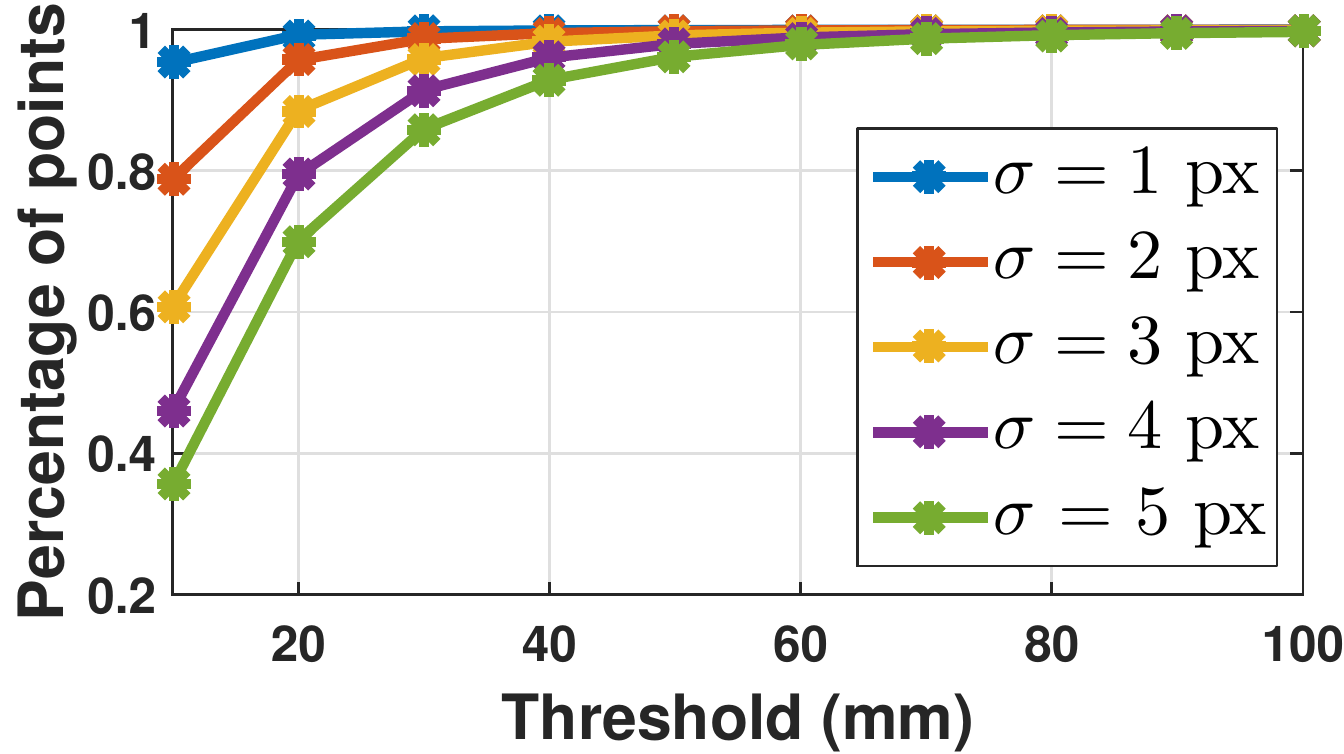}  
  \end{minipage}%
  \begin{minipage}[c]{0.65\linewidth}
    \centering
 \begin{tabular}[b]{|c|*{6}{c|}}
	\hline
  \backslashbox{Noise\kern-3em}{\kern-1emThreshold}
	& {10} & {20} & {30} & {40} & {50} & {100}\\\hline
	{$\mathcal{N}(0,1)$}  & 0.9529  &  0.9925 &   0.9974  &  0.9987  &  0.9992 &   0.9998\\
	\hline
	{$\mathcal{N}(0,2)$}  &   0.7878 &   0.9568  &  0.9869 &   0.9949 &   0.9976 &   0.9997\\
	\hline
	{$\mathcal{N}(0,3)$}  & 0.6074 &   0.8855 &   0.9593  &  0.9828  &  0.9917  &  0.9991\\
	\hline
	{$\mathcal{N}(0,4)$}  & 0.4601  &  0.7941 &   0.9144  &  0.9602  &  0.9797  &  0.9980\\
	\hline
	{$\mathcal{N}(0,5)$}  &  0.3551  &  0.7008  &  0.8590  &  0.9287 &   0.9615  &  0.9966\\
	\hline	
\end{tabular}
\end{minipage}
\caption{The reconstruction accuracy when the 2D observations are corrupted with Gaussian noise of different standard deviation ($\sigma$).}
\label{fig:error_noise}
\end{figure*}

\begin{figure*}
\centering
  \begin{minipage}[c]{0.35\linewidth}
    \centering
    \includegraphics[width=0.9\linewidth]{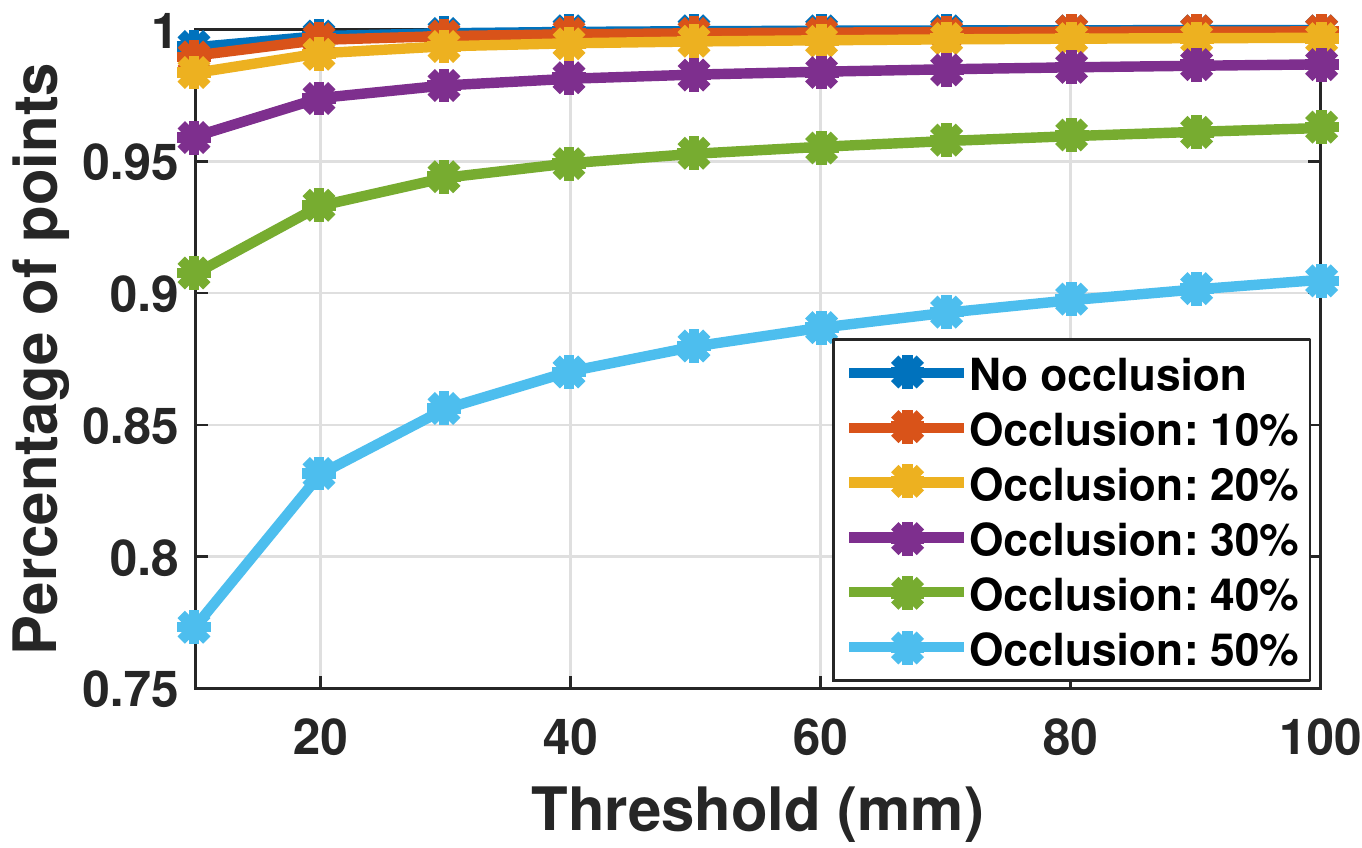}  
  \end{minipage}%
  \begin{minipage}[c]{0.65\linewidth}
    \centering
\begin{tabular}[b]{|c|*{6}{c|}}
	\hline
  \backslashbox{Miss rate\kern-2em}{\kern-1emThreshold}
	& {10} &{20} & {30} & {40} & {50} & {100} \\\hline
	{0\%} & 0.9933 & 0.9975 & 0.9986 & 0.9991 & 0.9994 & 0.9998 \\
	\hline
	{10\%} & 0.9901 & 0.9961 & 0.9975  &  0.9982  &  0.9986  &  0.9993 \\
	\hline
	{20\%} & 0.9835  &  0.9910 &   0.9936 &   0.9948  &  0.9955 &   0.9968 \\
	\hline
	{30\%} & 0.9594  &  0.9740  &  0.9788 &   0.9813 &   0.9829 &   0.9868 \\
	\hline
	{40\%} & 0.9074  &  0.9331  &  0.9438  &  0.9493 &   0.9529 &   0.9626 \\
	\hline
	{50\%} & 0.7734  &  0.8313  &  0.8560  &  0.8703 &   0.8798 &   0.9050 \\
	\hline
\end{tabular}
\end{minipage}
\caption{The reconstruction accuracy under different percentages of occluded points.}
\label{fig:error_occlusion}
\end{figure*}

\section{Experiments} \label{sec:experiment}
In our experiments, we evaluate our algorithm on both synthetic and real datasets. $\lambda_1$ and $\lambda_2$ in Eq.~(\ref{eq:ourproblem_final}) are set empirically to 0.05 and 0.1 for all the experiments. To alleviate the influence of different camera system scales (\ie~differing the scale of $\allX$), the average distance between camera centers is normalized to 1 before applying our method. The soft constraint parameterization is used only in the presence of noisy measurements.

\subsection{Simulation}
We use synthetic datasets to evaluate the accuracy and robustness of our proposal, and also compare against two state-of-the-art methods \cite{valmadre2012general,dai2014simple}.
To generate synthetic data, we use the real motion capture datasets from \cite{cg-2007-2}, and leverage them as ground truth structure for our estimation.
The whole datasets contain 130 different real motions including hopping, jogging, cartwheel, punching, \emph{etc}.
Each motion capture dataset is comprised of the temporal sequences of a common set of 44 3D points in real scale, which corresponds within our framework to ground truth structure $\allX_{GT}$. The frame rate of the motion datasets, \ie the sampling rate of the real continuous motion, is 120 Hz. The length of each dataset ranges from 45 to 701 frames, and with an average of 273. 

These 3D points are projected onto virtual cameras to generate input 2D measures into our methods.
We select 4 virtual cameras with a resolution of 1M and focal length of 1000, and we position the static cameras around the centroid defined by $\allX_{GT}$. 
The distance of the camera to the centroid is approximately twice the scale of $\allX_{GT}$, and on average the distance is 2.7 meters.
Considering the frame rate of the motion capture datasets is 120 Hz and there are 4 virtual cameras, the average frame rate for each camera is 30 Hz.
%The ratio between the distance to the centroid and the maximal distance between any two points in $\allX_{GT}$ is set to one.
Every temporal 3D capture is randomly assigned to each camera to build 4 disjoint image sequences. %To better evaluate how well our method recover the coefficient matrix $\allT$, 
Unless otherwise mentioned, we enforce that no temporally consecutive captures are assigned to the same image sequence.

To evaluate our method, Euclidean errors between the ground truth and the estimated 3D points are computed.
We define the accuracy by counting the percentage of points having errors less than thresholds of $10$, $20$, $30$, $40$, $50$, and $100$ mm.

\subsubsection{Accuracy} \label{sec:experiment_accuracy}
\textbf{Different frame rates.}
We first evaluate how the algorithm behaves under different capture frame rates.
2D measures without noise are used to evaluate the accuracy of our method. In addition to the original motion capture data at 120 Hz, we also downsample the data to 60 and 30 Hz, so that each camera has frame rate of 15 and 7.5 Hz on average. As shown in Fig.~\ref{fig:error_framerate}, the accuracy becomes worse as the frame rate gets slower. 
The main reason is that the self-representation residual is larger at lower frame rate. We notice that at a frame rate of 7.5 Hz, our method does not work well on the quick motions with large and nonlinear shape deformation, such as hopping or arms rotation. However, still  more than 97\% of 3D points have errors less than 5 cm, which is already very small considering the scale of a person and the distance range of the cameras.
%One of the main reasons is that the residual becomes larger and then the reconstructability becomes smaller. Note that the residual depends on both the frame rate and the speed of motion.
%We also notice that \hl{to be continued...}
%which is already very small considering the scale of the person and the distance of cameras from the person.

\textbf{Local temporal information.}
We also quantitatively evaluate the estimated $\allT$. Using the same two measures described in Sec.~\ref{sec:principle}, 
we get values of 0.9902 and 0.9923, compared to 0.9972 and 0.9994 if the 3D points are given. Therefore, our method very accurately recovers the local temporal information.

\begin{figure}[t]
\centering
\includegraphics[width=0.29\textwidth]{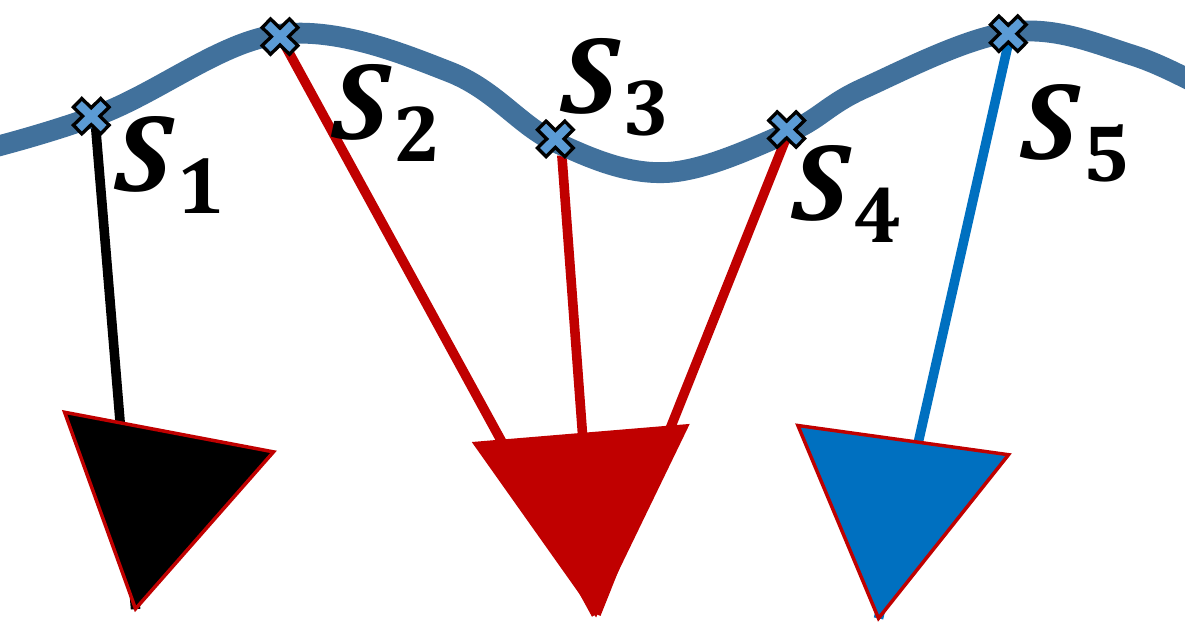}
\caption{Consecutive captures are assigned to the same red camera. For easy visualizations, only one point per shape is drawn.}
\label{fig:unconstrained_assign}
\end{figure}

\textbf{Unconstrained capture assignment.}
We test the case that each capture is randomly assigned to one of the four cameras so that temporally consecutive captures could have a chance to be assigned to the same camera, as is shown in Fig.~\ref{fig:unconstrained_assign}. In this specific case, shapes $\mathbf{S}_1$ and $\mathbf{S}_5$ are used to represent $\mathbf{S}_2$, $\mathbf{S}_3$ and $\mathbf{S}_4$. Based on the theory in Sec.~\ref{sec:shape_approximation}, using spatially further away shapes to represent the current shape has larger residual and hence larger reconstruction errors, as is validated in Fig.~\ref{fig:error_framerate}.
 
\subsubsection{Data robustness}

To evaluate the robustness of our method, we test it in the case of noisy measurements and missing data.

\begin{figure}[t]
\centering
\includegraphics[width=0.29\textwidth]{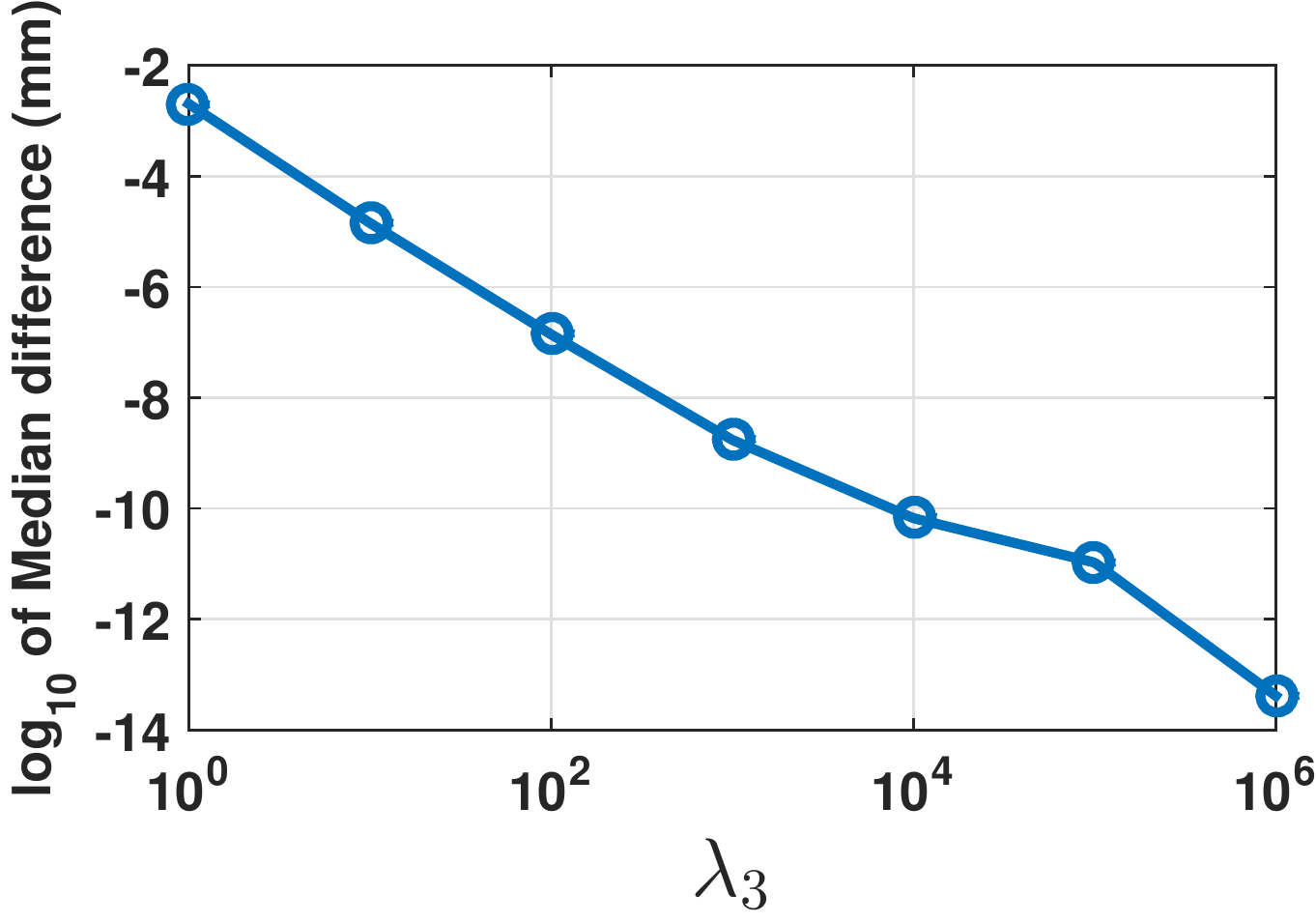}
\caption{The difference of the estimated results by the hard constraint formulation in Eq.~(\ref{eq:X_representedBy_d_all}) and the soft constraint formulation in Eq.~(\ref{eq:soft_constraint}) with different $\lambda_3$} 
\label{fig:soft_hard_diff}
\end{figure}

\textbf{Noisy measurements.} We add zero-mean Gaussian noise with different standard deviations to the 2D measurements. Considering that the focal length of the image is 1000 pixel, one pixel error corresponds to one millimeter if the object is one meter away.
We apply the soft constraint formulation described in Sec.~\ref{sec:noisy_measure} and empirically set the parameter $\lambda_3$ to 100. As depicted in Fig.~\ref{fig:error_noise}, the quality of reconstruction degrades as the noise level increases. As $\lambda_3$ increases, the soft constraint approximates the hard constraint.  
We evaluate the difference of the estimated results by the hard constraint formulation and the soft constraint formulation with different $\lambda_3$, and we show the median difference in Fig.~\ref{fig:soft_hard_diff}. It is apparent that as $\lambda_3$ increases, the difference of the output between the two formulations becomes smaller.

We have tested the hard constraint formulation using noisy measurements, and the overall accuracy of the output is very similar. Though the soft constraint appears more robust in the presence of noise as it allows the points off the viewing ray, there is no guarantee or proof this constraint will achieve more accurate results, as it depends on the exact motion of the objects.

\textbf{Missing data.} %Missing observations occur in practice due to occlusion or mis-detection. 
In our evaluation, we randomly set some 2D measures to be unavailable. Fig.~\ref{fig:error_occlusion} depicts the accuracy under different percentages of missing data. We observe that under 20\% of occlusion, there is not much difference in reconstruction accuracy. Moreover, under a large amount of 40\% occlusion, our method still produces accurate results,  with 94.38\% of points having errors less than 30 mm. 

Our method essentially linearly interpolates the 3D points along the trajectory using estimated  $\allT$. It can still produce 3D estimates in the presence of consecutive missing observations across time, but the accuracy in such scenarios depends on the object motion. Particularly, given large displacement of nonlinear motion, our method is likely to produce less accurate results.

%and use the same evaluation criterion as described in Sec.~\ref{sec:experiment_accuracy}. 
%Figs.~\ref{fig:hard_constraint_noise} and \ref{fig:soft_constraint_noise} show the errors of our method using hard and soft constraint for the parameterization of $\allX$. 
 
%\begin{figure}[t]
%\centering
%\includegraphics[width=0.4\textwidth]{resource/initialization/Init_result_compare.pdf}
%\caption{The average Euclidean error of the initialization and the final output.}
%\label{fig:initialization_optimal_noise}
%\end{figure}

\subsubsection{Comparison to other methods} \label{sec:comparison_to_other_methods}
We compare our method with a NRSFM method \cite{dai2014simple} and A the trajectory triangulation method \cite{valmadre2012general}. Both of these methods are state-of-the-art for dynamic object reconstruction.

\textbf{NRSFM method.} Non-rigid structure from motion (NRSFM) recovers both the camera motion and the dynamic structure. It is tempting to use those methods to solve our problem, since our problem with known camera poses seems to be easier. 
However, most NRSFM methods work on an orthographic or weak perspective camera model, and it is unclear of their applicability under the perspective model. Park \etal~\cite{park20103d} test the NRSFM methods \cite{Akhter_NIPS08,torresani2008nonrigid,paladini2009factorization} under a perspective camera model, but all of them fail to produce reasonably good results. In this paper, we test the state-of-the-art NRSFM method by Dai \etal~\cite{dai2014simple}.

%The first method we compare to is prior-free NRSFM By Dai \etal \cite{dai2014simple}.
The method by Dai \etal \cite{dai2014simple} is based on the assumption that each non-rigid shape $\oneX_f$ is a linear combination of $K$ shape bases, and hence the shape matrix (corresponding to $\allX$ in our problem description) has low rank.
After estimating the camera motion,  they recover the structure by minimizing the rank of the shape matrix, which is achieved through the minimization of the matrix nuclear norm.  Their method applies to an orthographic camera model, but can be easily adapted to a perspective model, as described below. 

Given the camera poses, we use the block matrix method proposed in \cite{dai2014simple} for shape estimation. Denoting
\begin{equation}
\resizebox{0.98\linewidth}{!}{%
$\allX^\# = 
\begin{bmatrix}
X_{(1,1)}& \dots& X_{(P,1)} & Y_{(1,1)}& \dots& Y_{(P,F)} & Z_{(1,1)}& \dots& Z_{(P,F)} \\
\vdots   &  				& \vdots   & \vdots 		  & \vdots   & \vdots			  \\
X_{(1,F)}& \dots& X_{(P,F)} & Y_{(1,F)}& \dots& Y_{(P,F)} & Z_{(1,1)}& \dots& Z_{(P,F)}\nonumber		  
\end{bmatrix}$ },
\end{equation}
where $\mathbf{X}_{(p,f)} = (X_{(p,f)},Y_{(p,f)},Z_{(p,f)})^\text{T}$, the shape of the object can be recovered through 
\begin{equation}
\begin{aligned}
& \underset{\allX^\#, \allT}{\text{minimize}} &&
||\allX^\#||_* + \mu ||\mathbf{1}_{P\text{x}1} \otimes \allC + (\mathbbm{d} \otimes \mathbf{1}_{3\text{x}1}) \odot \allr - \allX||_{\text{F}}\\
&\text{subject to} && \allX^\# = \mathcal{L}(\allX), \nonumber
\end{aligned}
\end{equation}
where $||\cdot||_*$ is the matrix nuclear norm, $\mu$ is a positive weight, and $\mathcal{L}$ is a linear operator that reshapes $\allX$ into $\allX^\#$. 

\begin{figure*}[t]
\centering
\subfloat[Our method accurately reconstructs the 3D points ($1/\sigma_{\text{min}}=7.589$, $err = 0.0825$).]{
\includegraphics[width=0.115\textwidth]{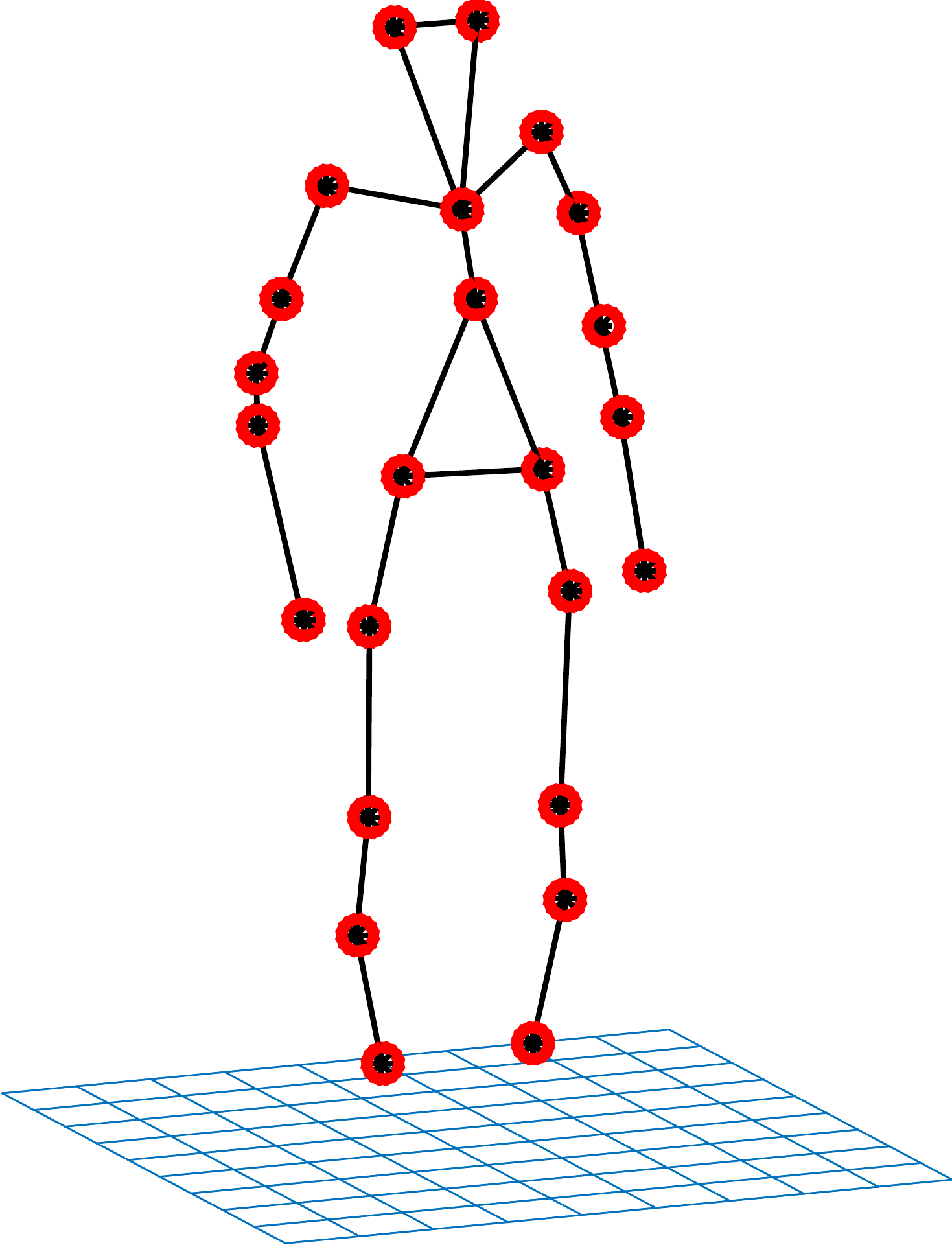}
\includegraphics[width=0.115\textwidth]{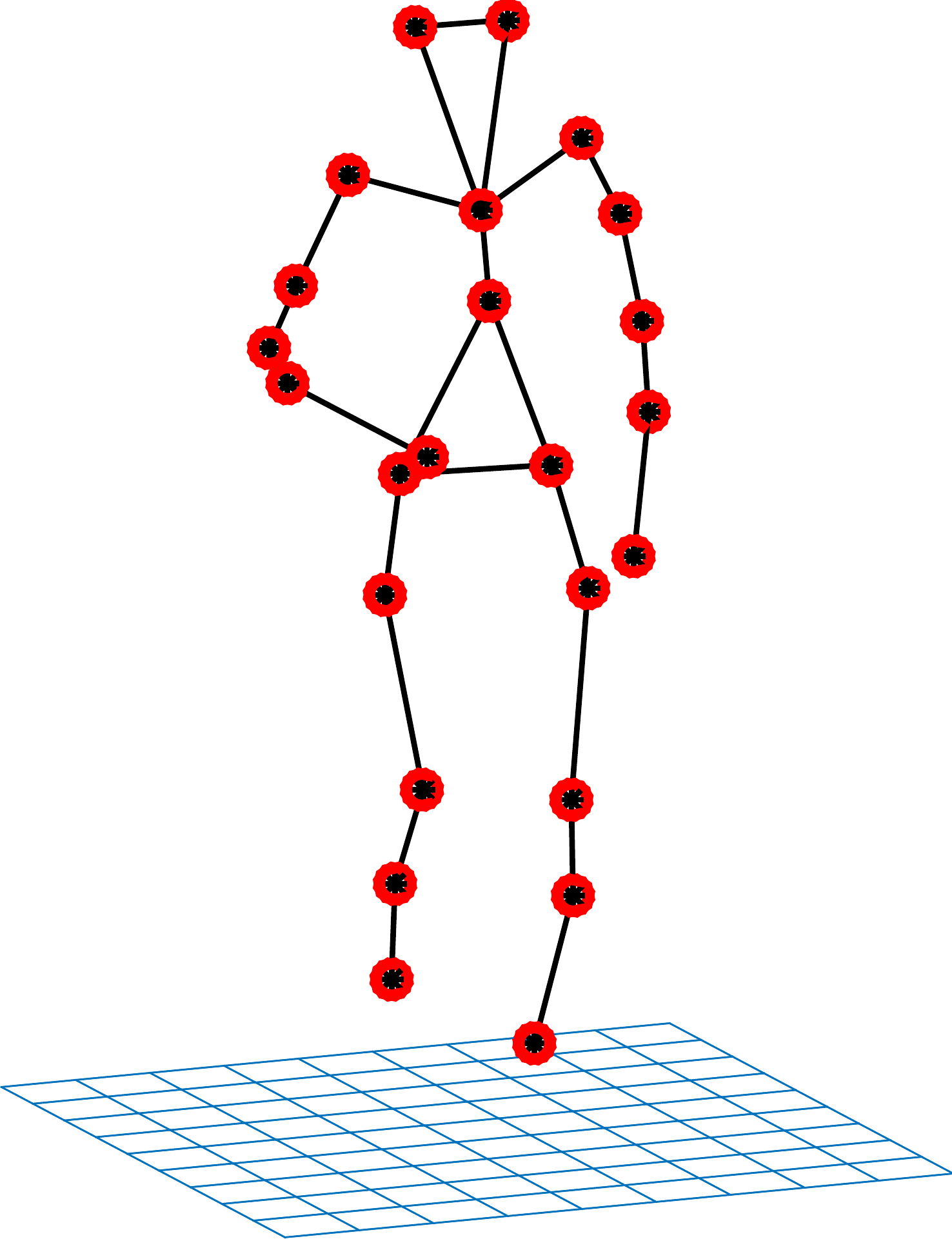}
\includegraphics[width=0.115\textwidth]{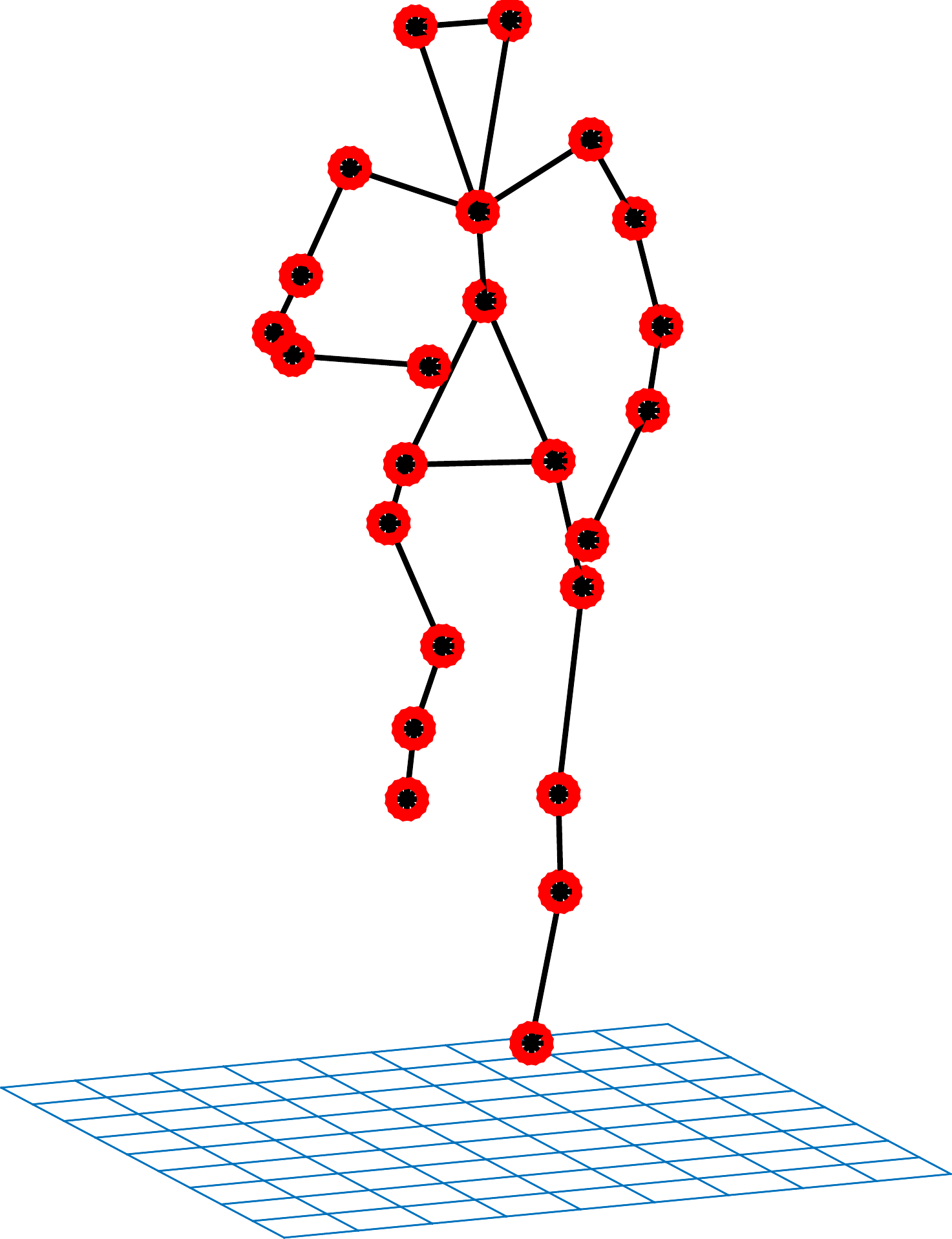}
\includegraphics[width=0.115\textwidth]{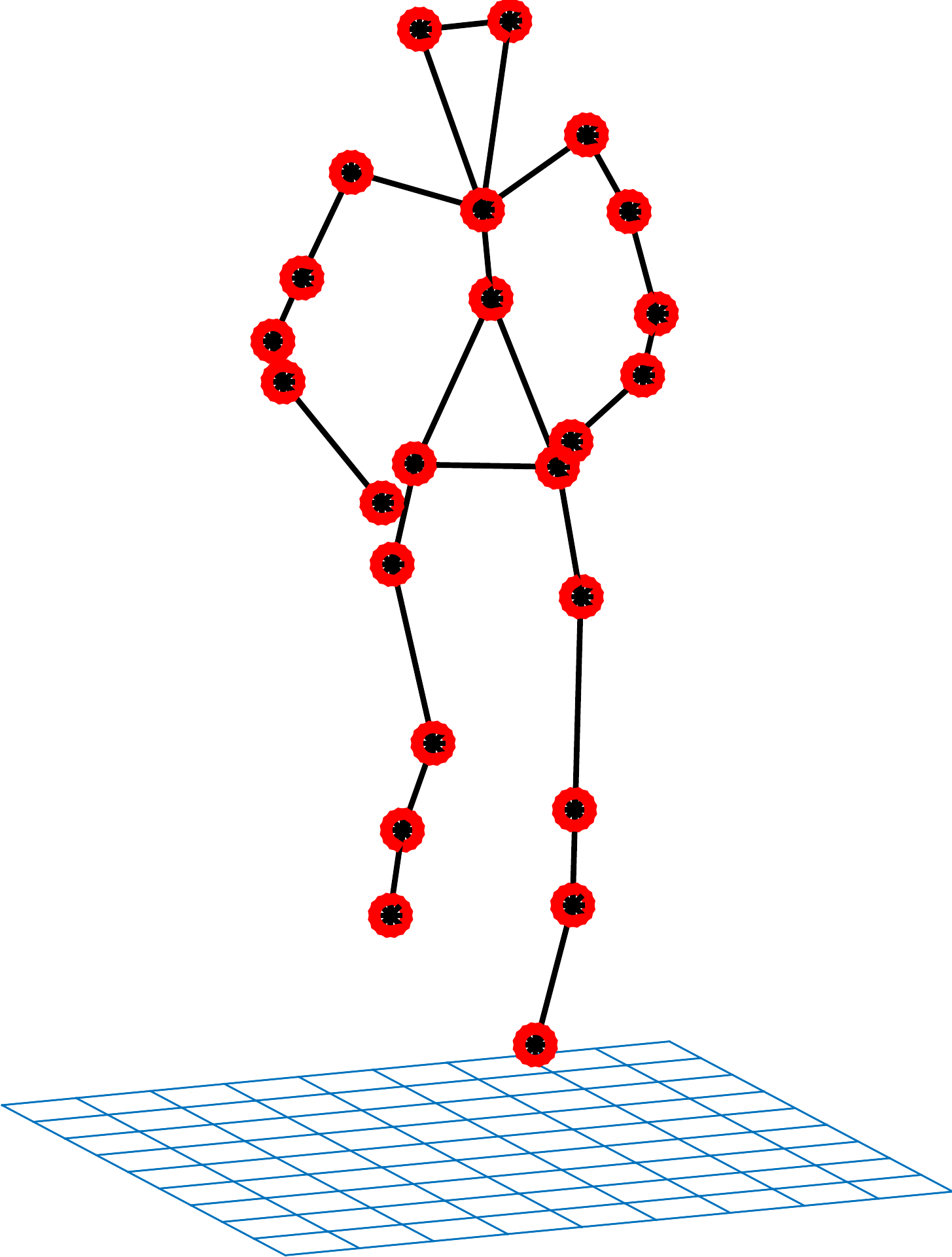}
\includegraphics[width=0.115\textwidth]{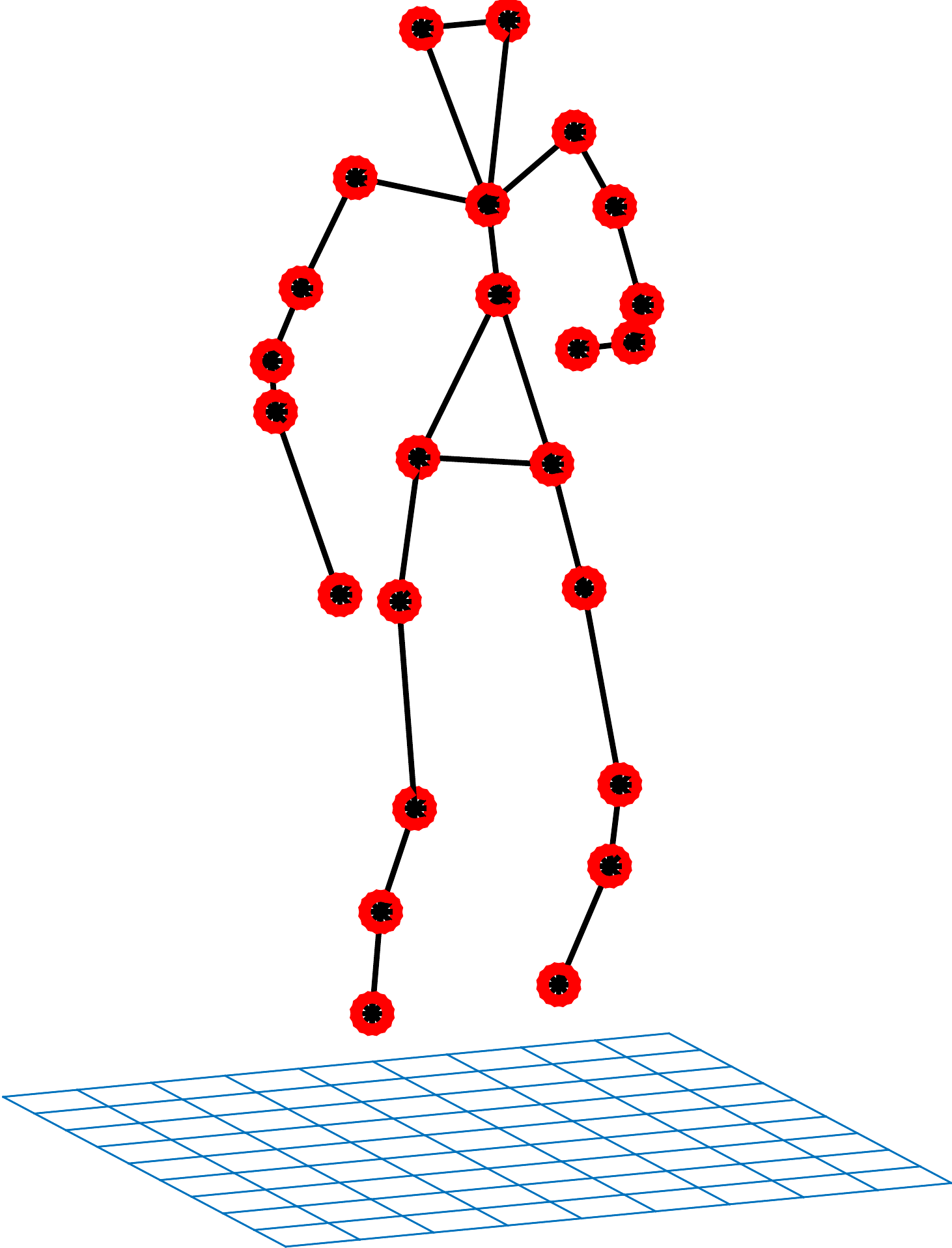}
\includegraphics[width=0.115\textwidth]{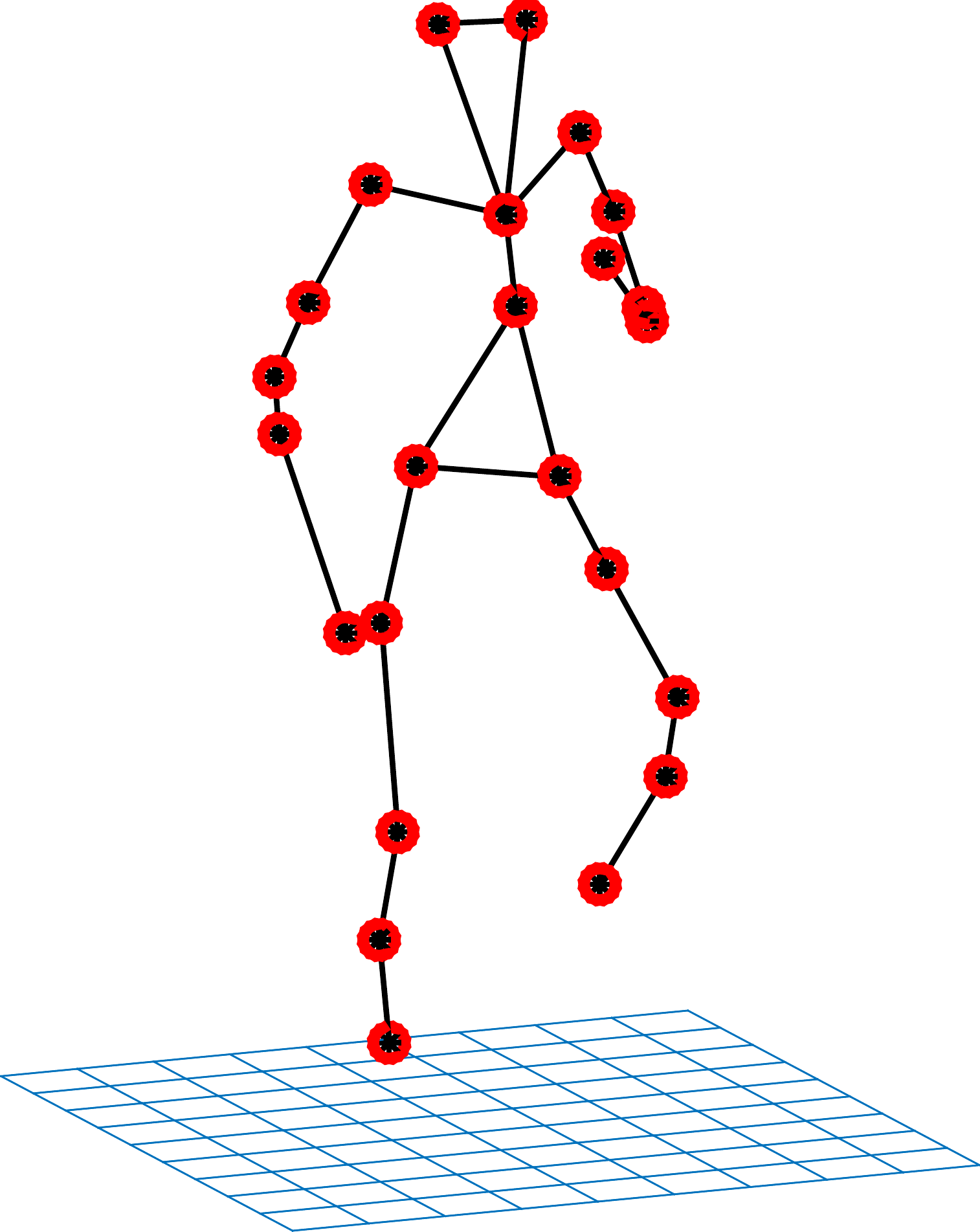}
\includegraphics[width=0.115\textwidth]{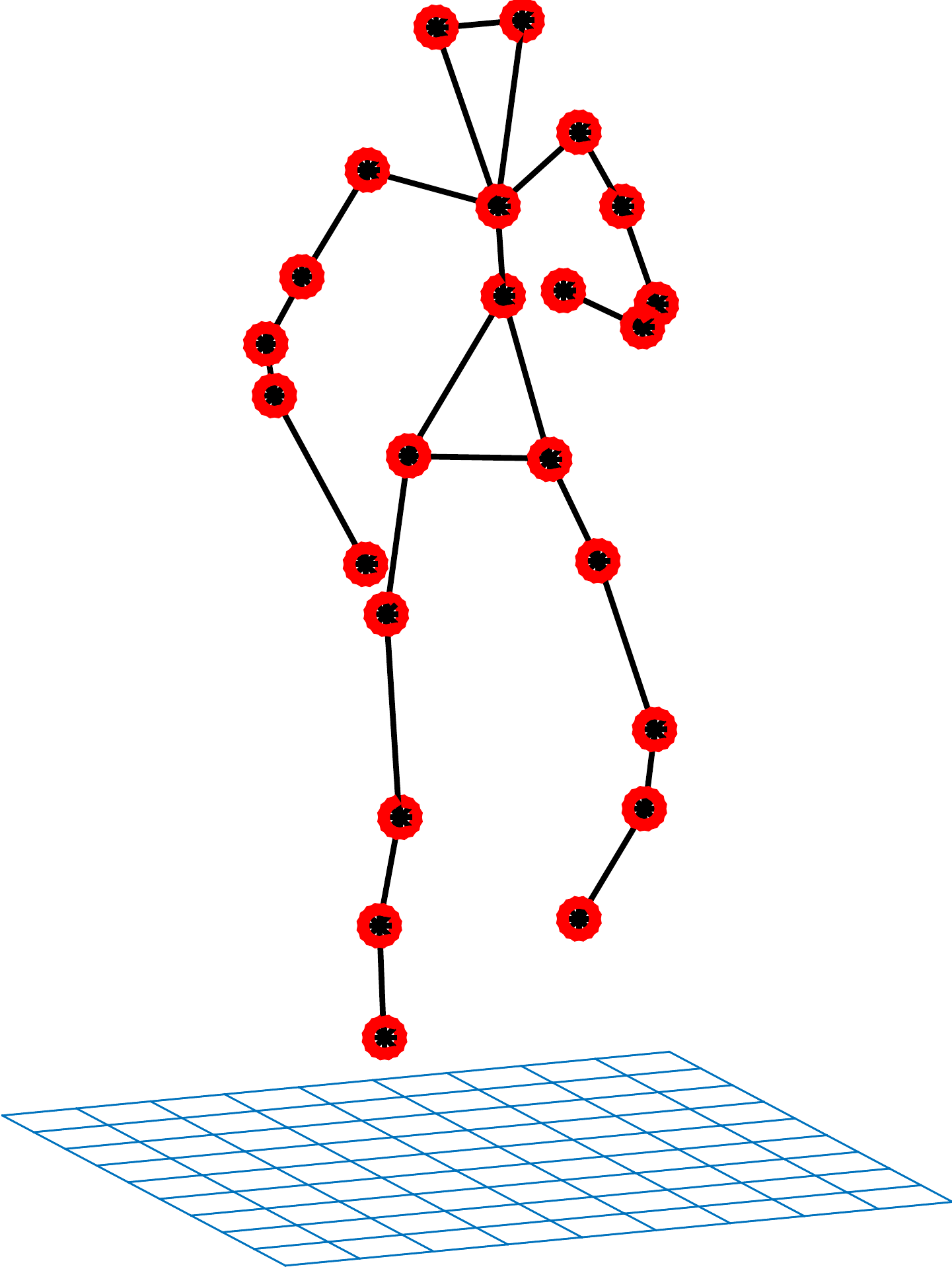}
\label{fig:ourmethod_Qualitative_result}
}\\
\subfloat[The modified prior-free method \cite{dai2014simple} fails to produce reasonable results. ($err = 472.9033$)]{
\includegraphics[width=0.115\textwidth]{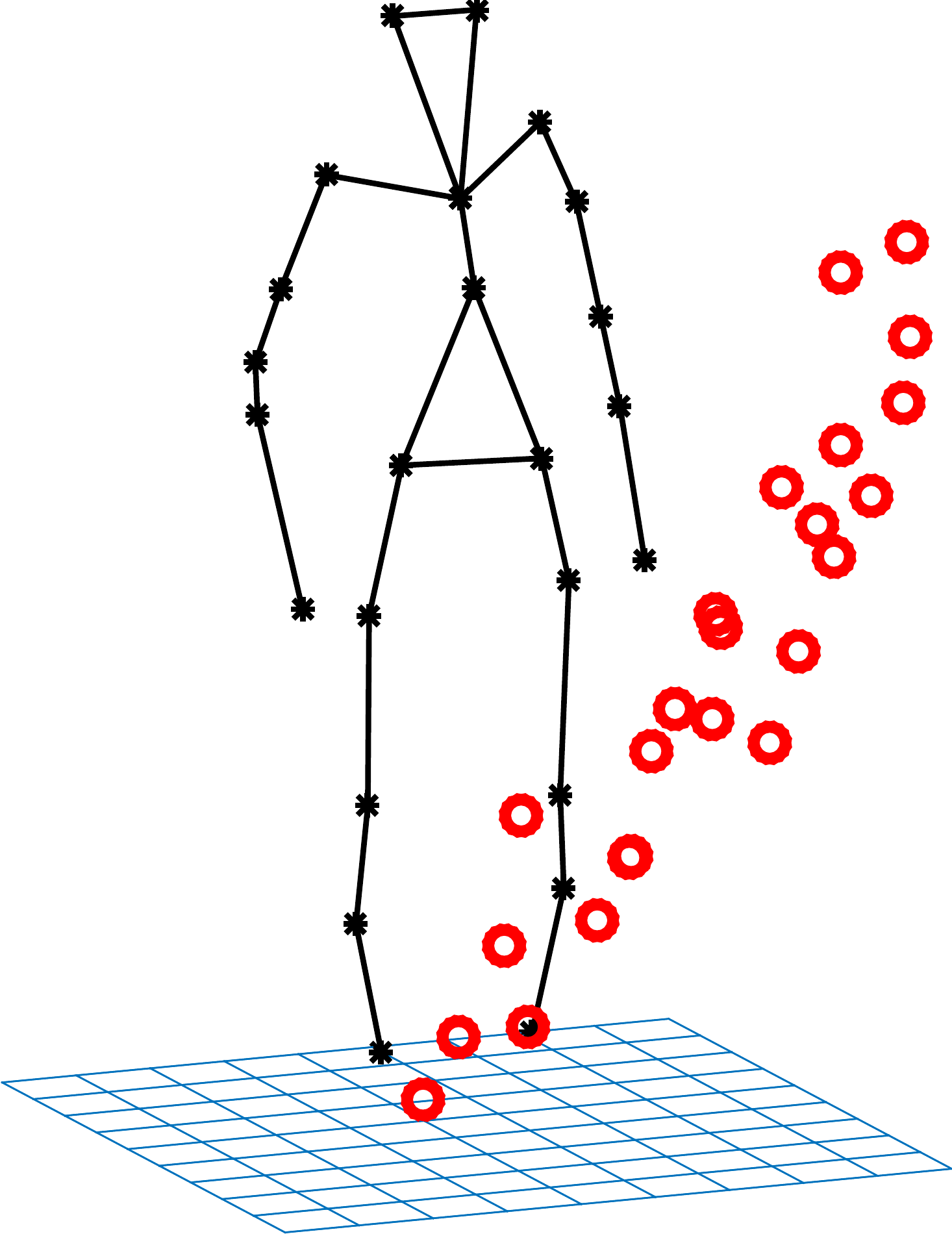}
\includegraphics[width=0.115\textwidth]{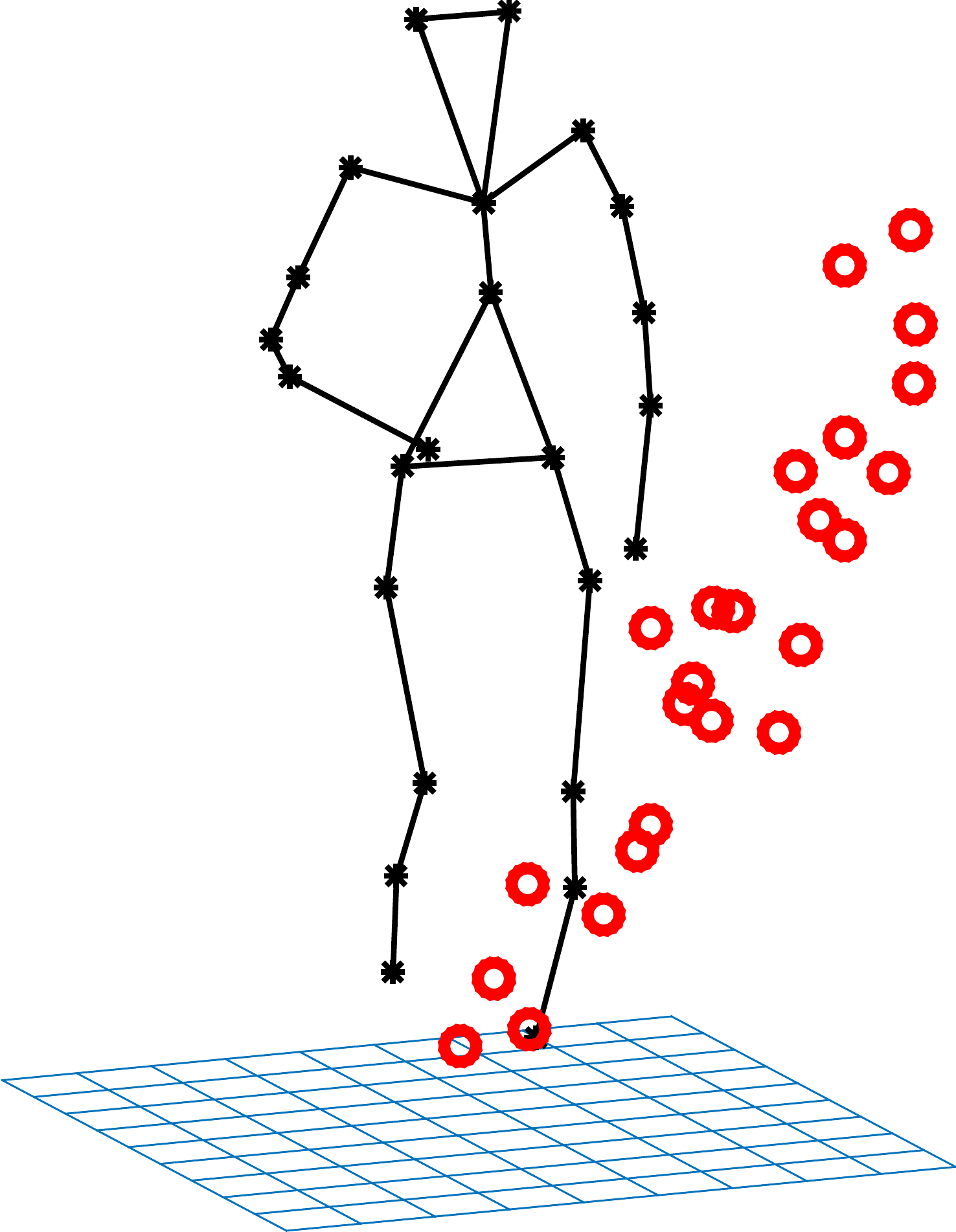}
\includegraphics[width=0.115\textwidth]{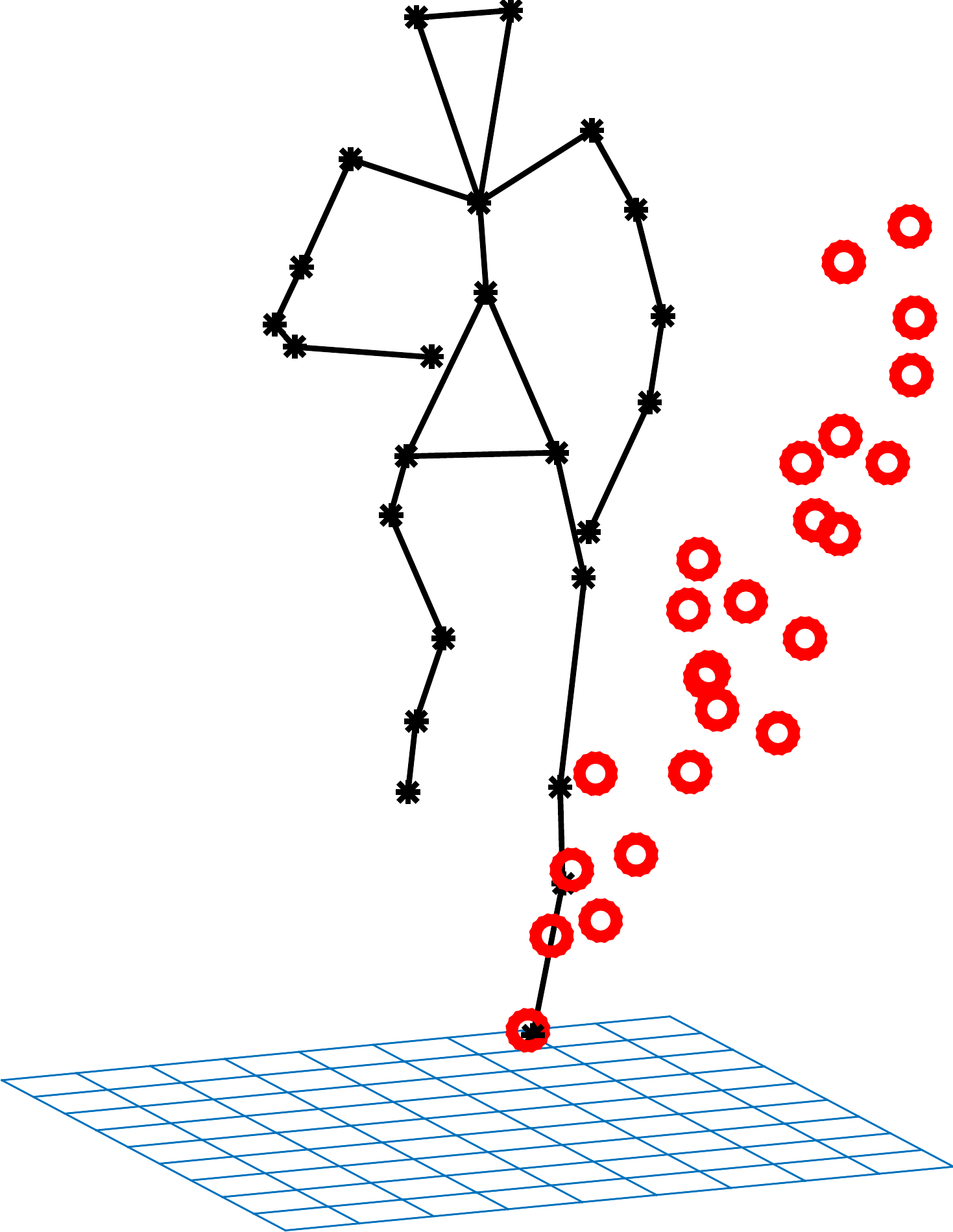}
\includegraphics[width=0.115\textwidth]{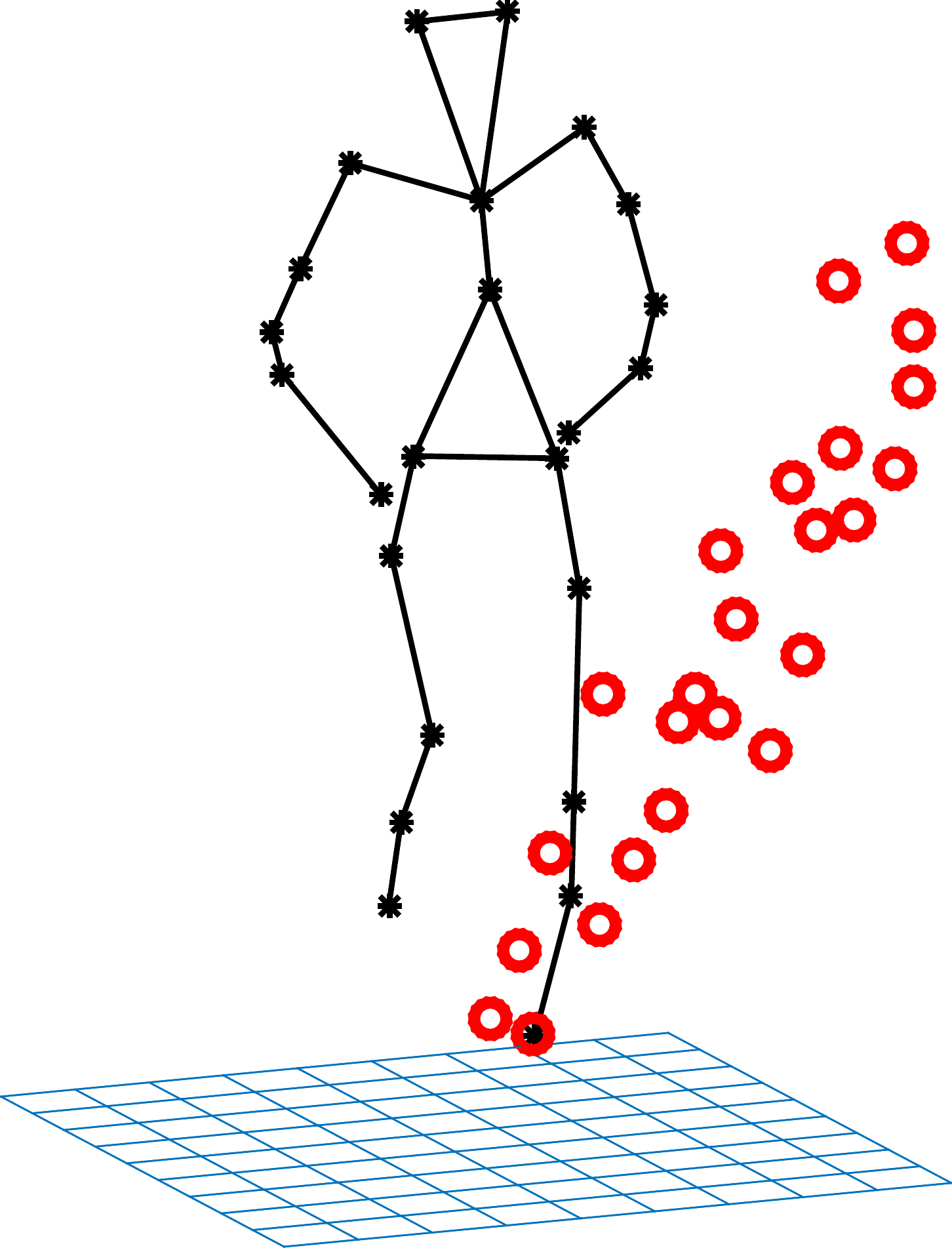}
\includegraphics[width=0.115\textwidth]{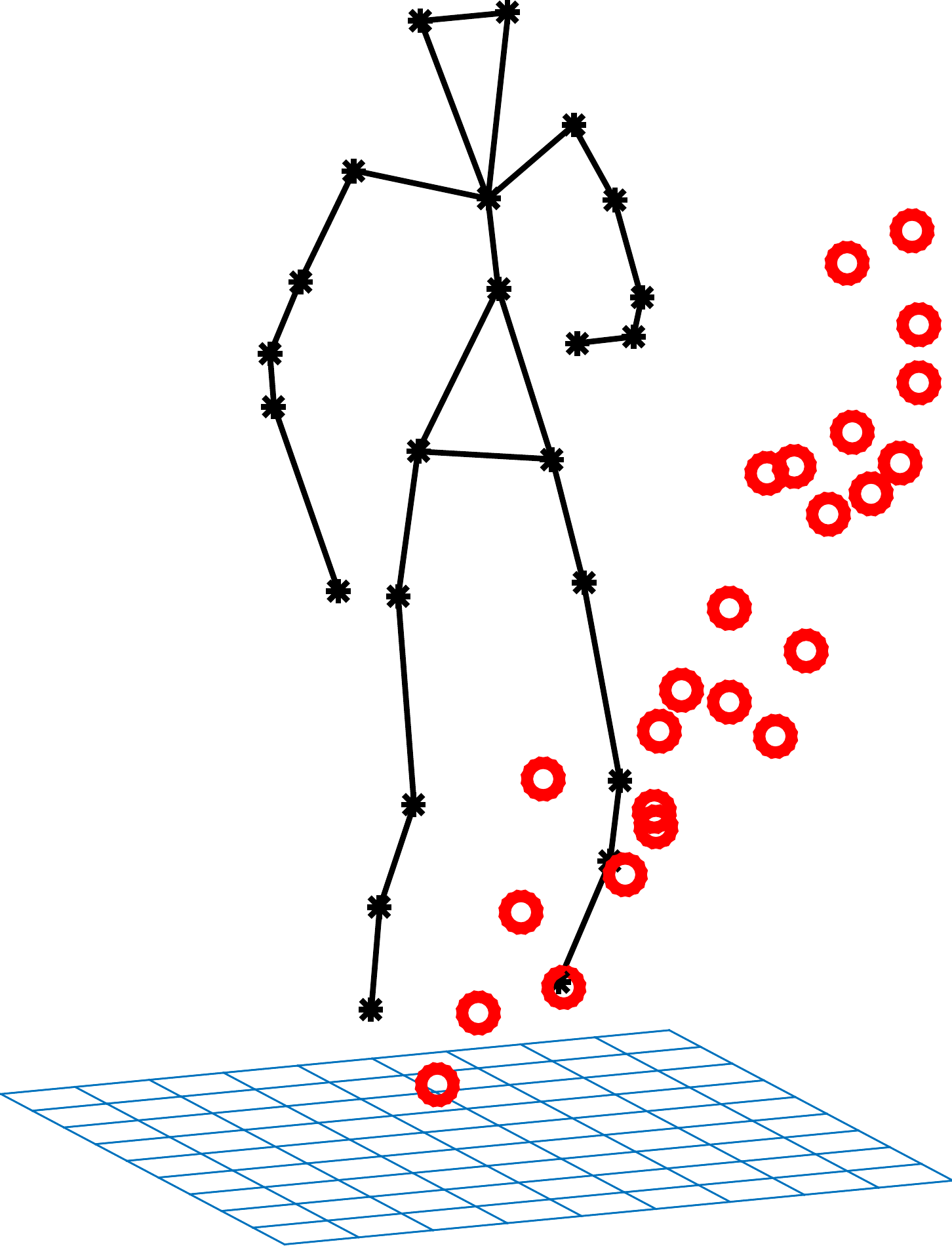}
\includegraphics[width=0.115\textwidth]{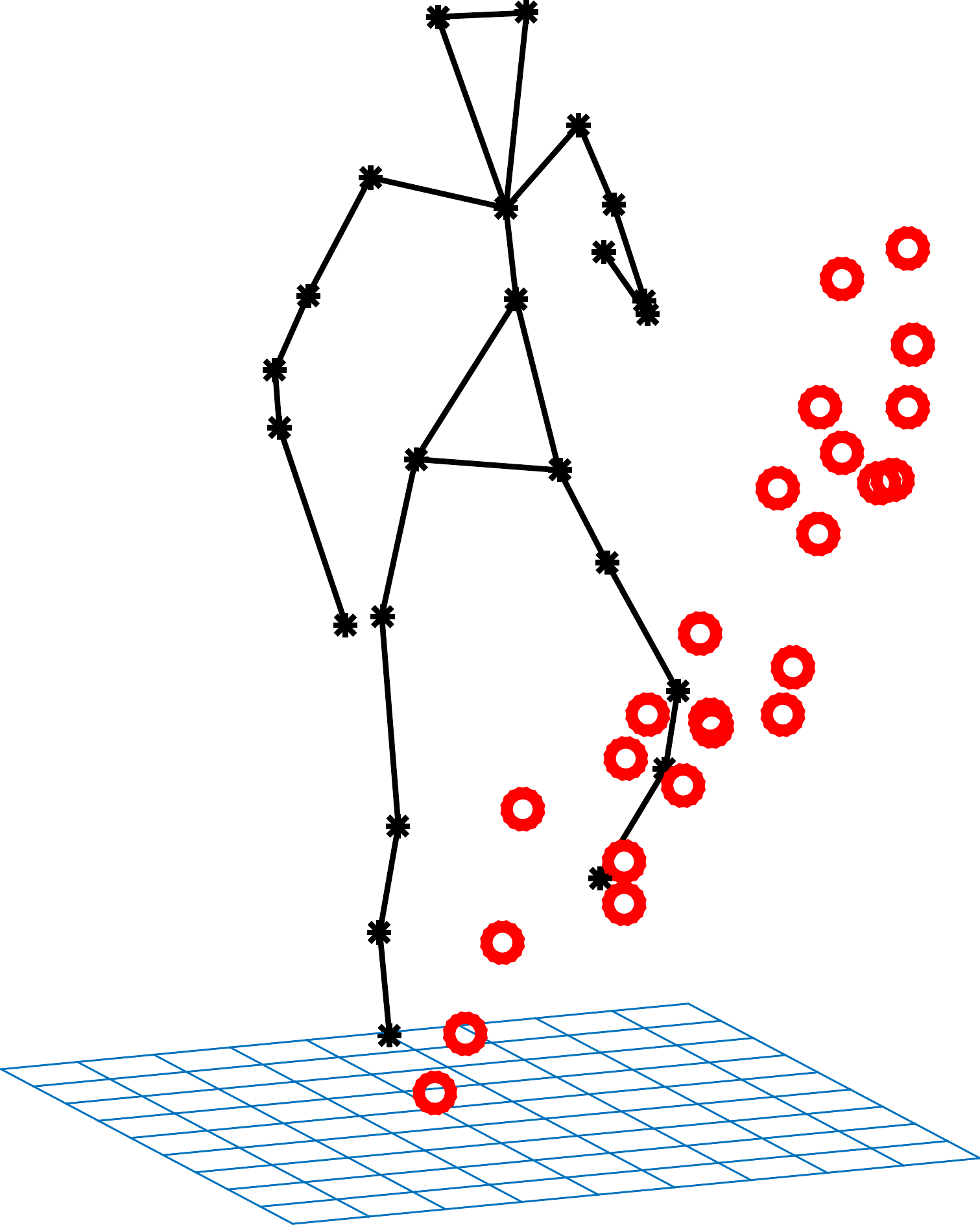}
\includegraphics[width=0.115\textwidth]{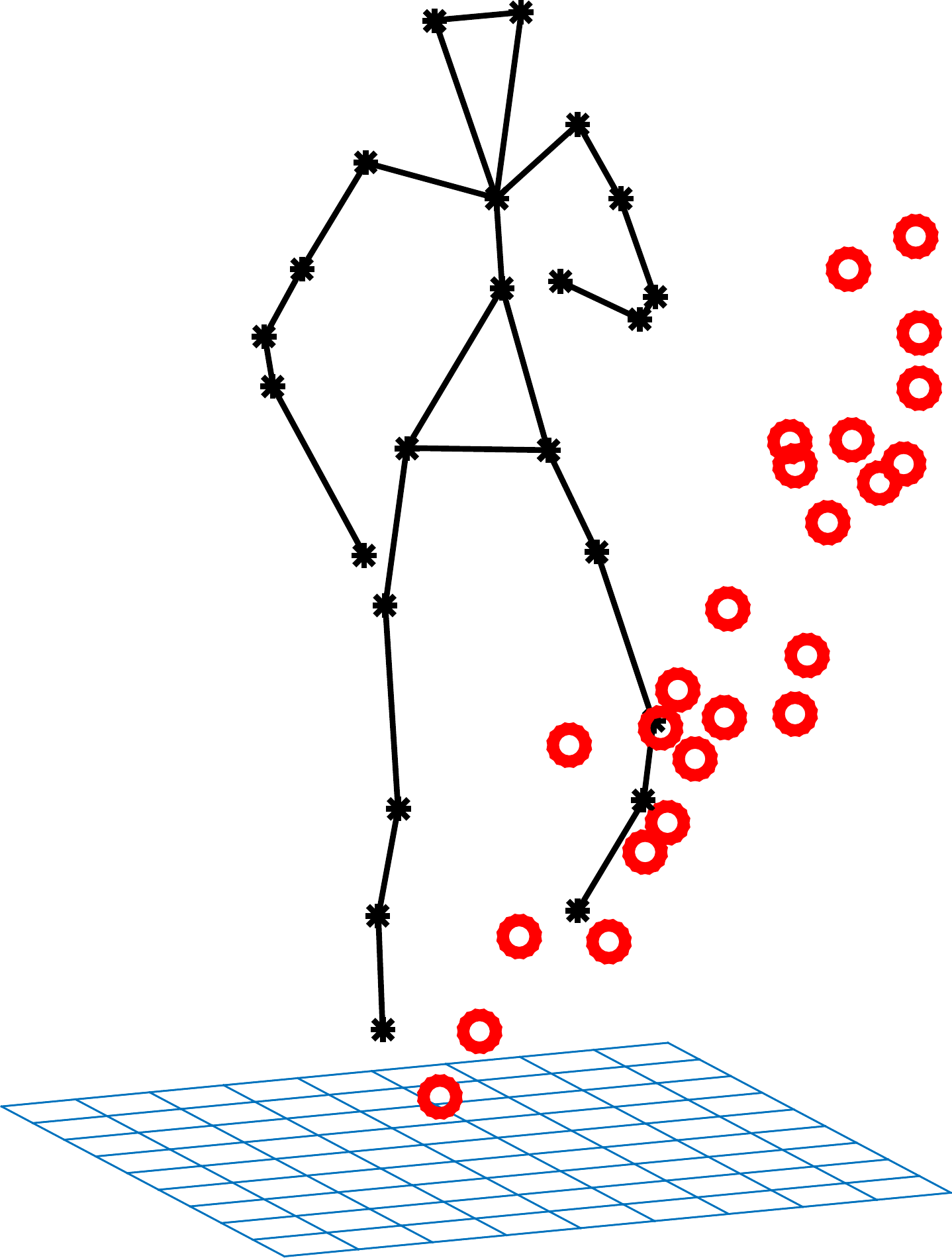}
\label{fig:prior_free_Qualitative_result}
}\\
\subfloat[General trajectory prior method \cite{valmadre2012general} produces large errors due to high system condition ($1/\sigma_{\text{min}}=2228$, $err = 76.9700$).]{
\includegraphics[width=0.115\textwidth]{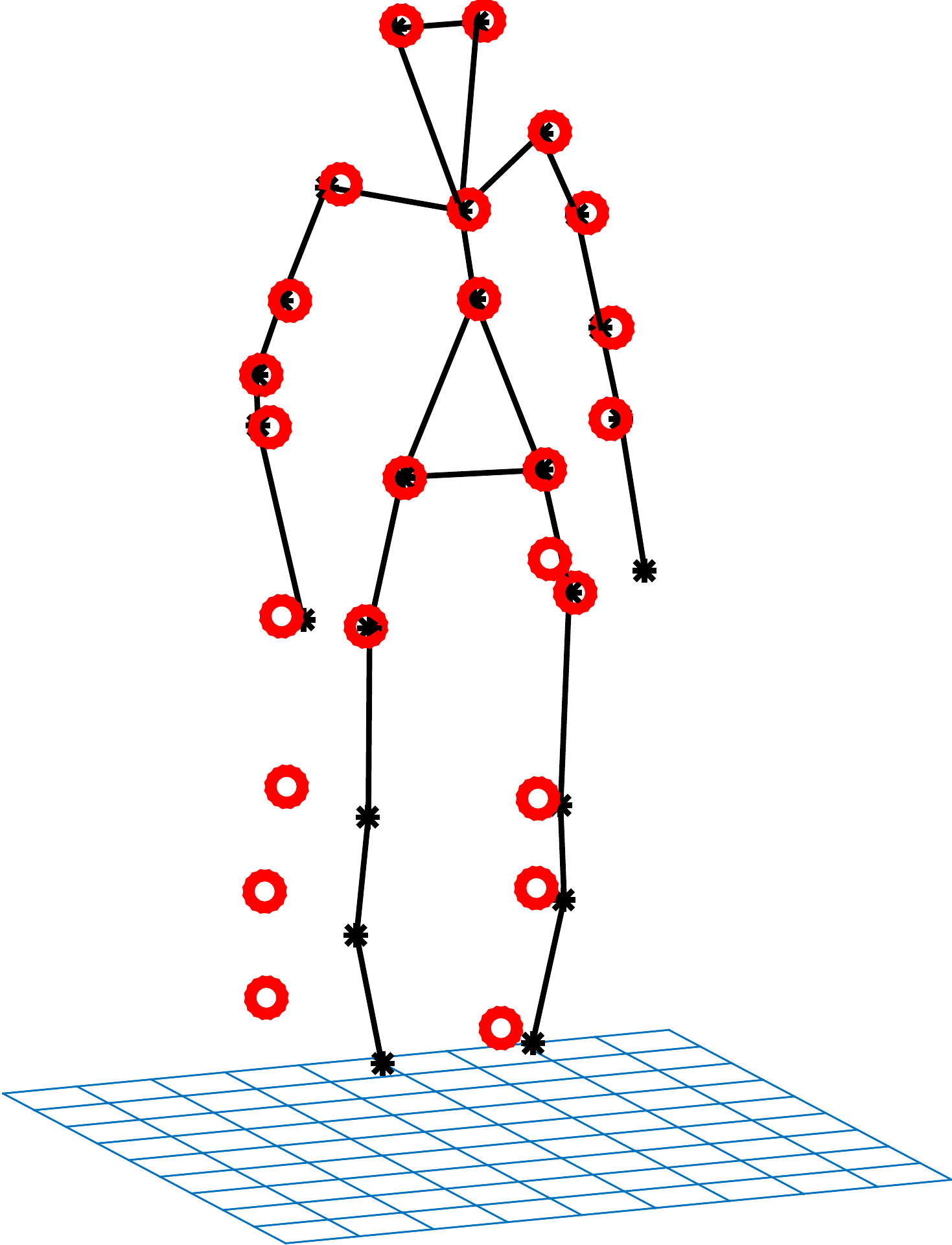}
\includegraphics[width=0.115\textwidth]{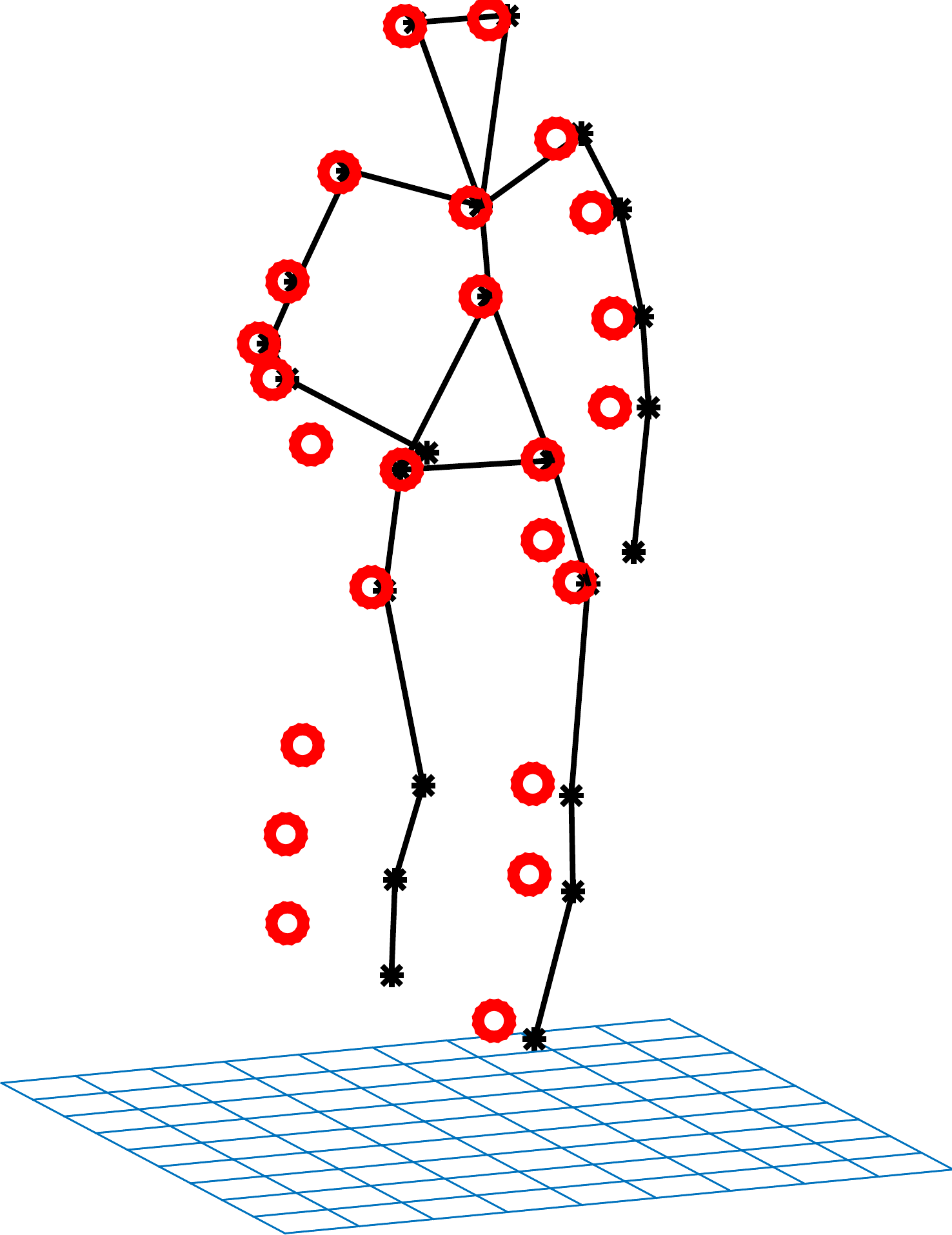}
\includegraphics[width=0.115\textwidth]{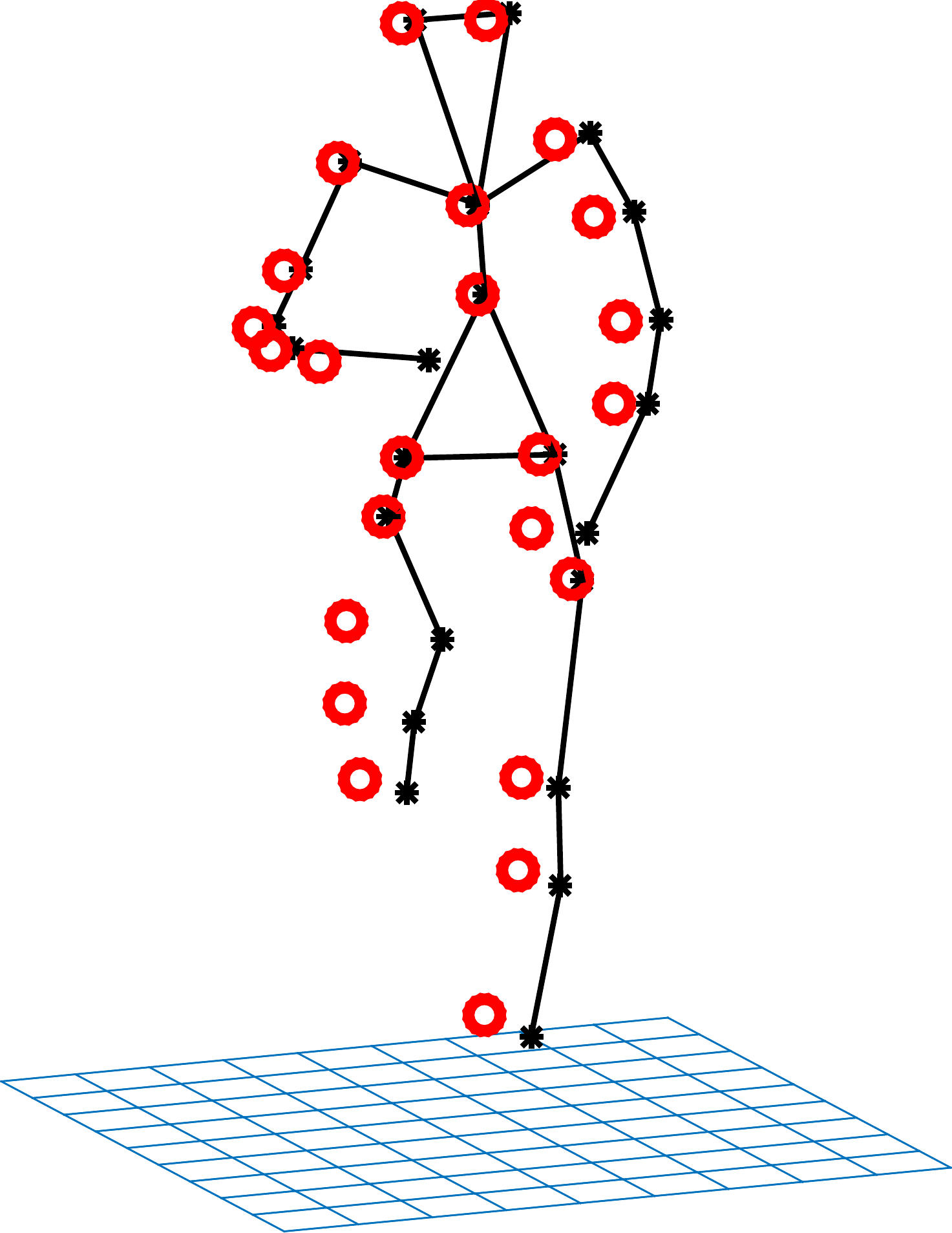}
\includegraphics[width=0.115\textwidth]{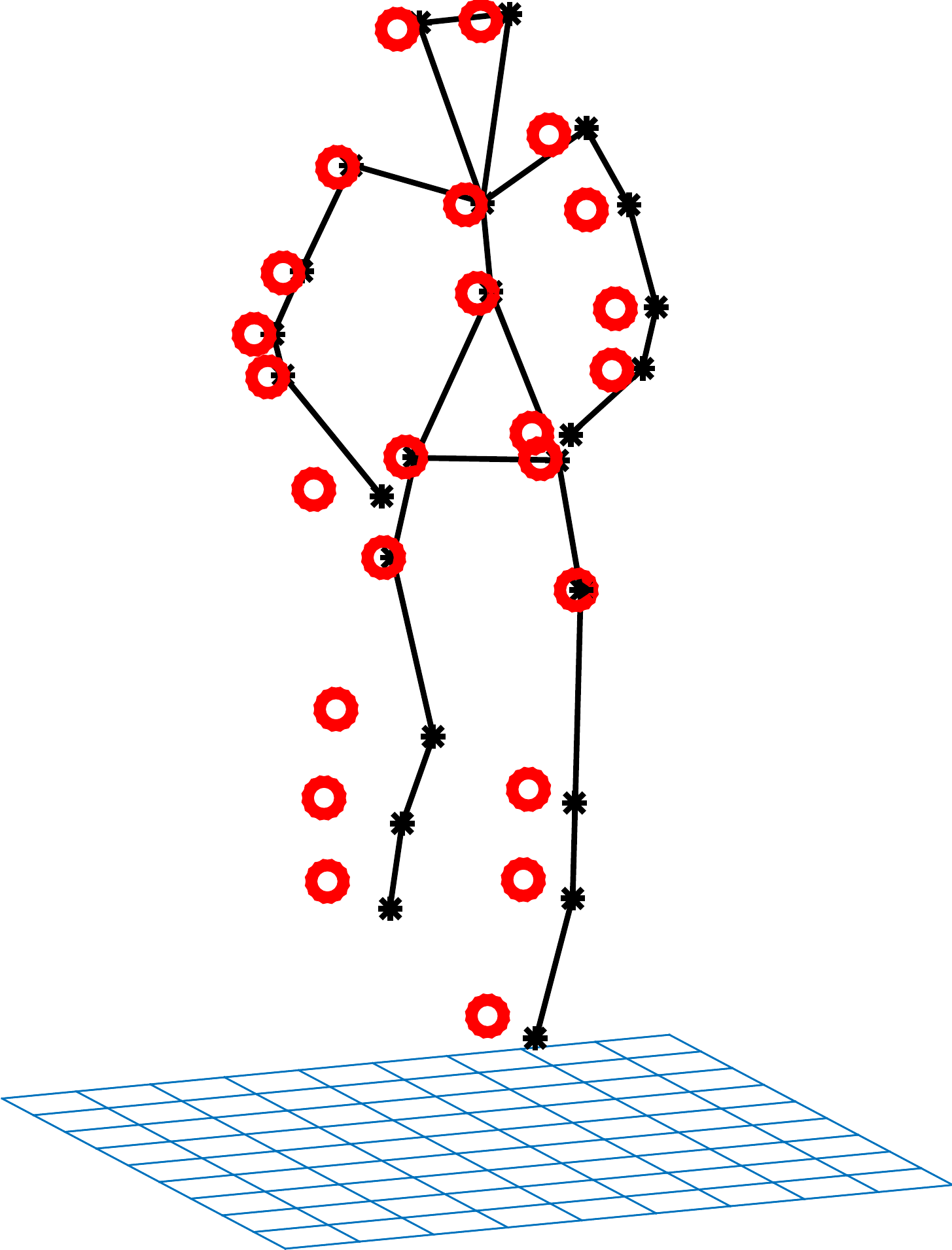}
\includegraphics[width=0.115\textwidth]{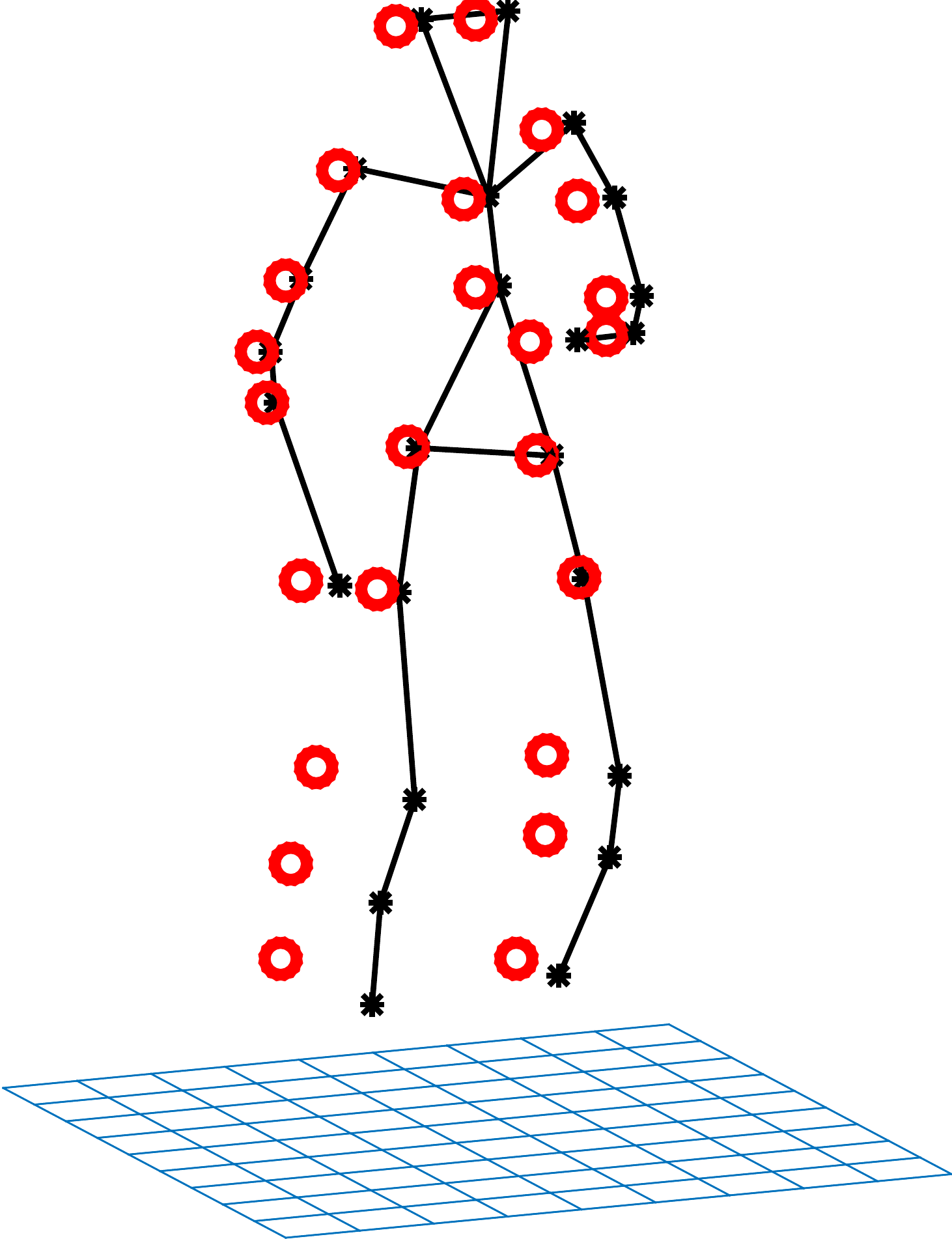}
\includegraphics[width=0.115\textwidth]{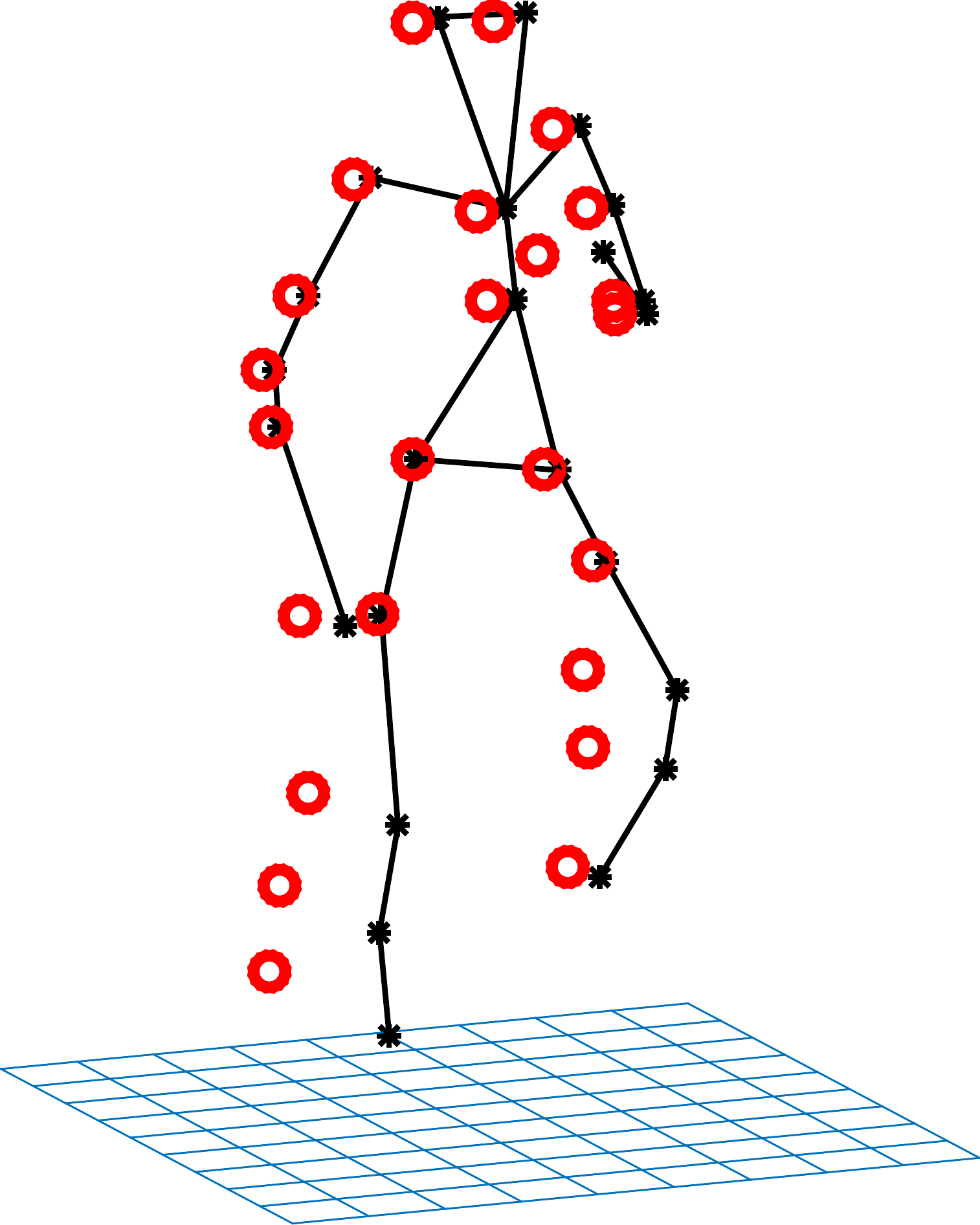}
\includegraphics[width=0.115\textwidth]{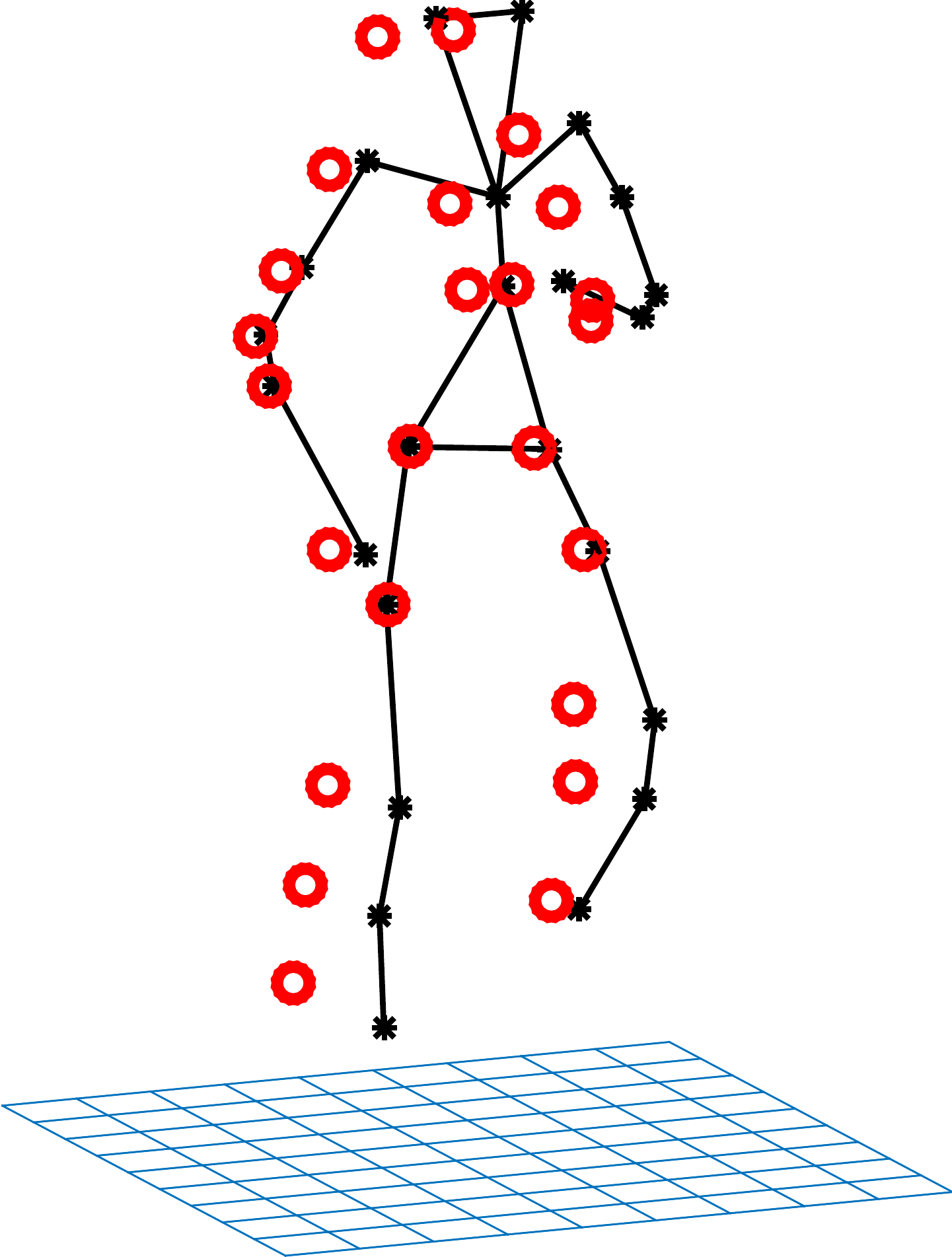}
\label{fig:trajectory_Qualitative_result}
}
\caption{Qualitative comparison of our method with \cite{dai2014simple} and \cite{valmadre2012general} on the motion capture dataset `jog on place' in \cite{cg-2007-2}. The dataset has 214 frames, with 44 points per frame (only 24 are shown for visualization purposes). The black and red points are the ground truth and the estimated results, respectively. $err$ is the average Euclidean error per point.}
\end{figure*}

This formulation seems attractive at first glance due to its convexity, in contrast to our non-convex formulation. Moreover, their method is shape-based (instead of trajectory-based), and does not require temporal information. 
To test the NRSFM method, we use synthetic data without noise and the random camera configuration shown in Fig.~\ref{fig:randCam}.
Unfortunately, the qualitative results in Fig.~\ref{fig:prior_free_Qualitative_result} show that it completely fails, as opposed to our method shown in Fig.~\ref{fig:ourmethod_Qualitative_result}. 

%The NRSFM method is closely related to our proposed solution. Their method assumes all the shapes can be represented by a linear combination of shape bases, while ours use self-representation. In the typical NRSFM, 

\textbf{Trajectory triangulation method.} We also compare with the trajectory triangulation method by Valmadre \etal \cite{valmadre2012general}, as is described in Sec.~\ref{sec:importance_of_image_sequencing}. 
Since the required sequencing information is readily available within each video stream, our test uses the simulation of one handheld camera as shown in Fig.~\ref{fig:oneCam}. The camera centers are Gaussian with 20 mm standard deviation ($\sigma_c$) around a fixed point. %The larger the standard deviation, the better the reconstructability. 
Based on the theory in Sec.~\ref{sec:system_condition}, the reconstructability increases with larger $\sigma_c$.
Considering that the framerate of the motion capture dataset is 120 Hz, the camera motion with  $\sigma_c = 20$mm is already very large compared to real handheld captures.

The method triangulates the trajectory of each dynamic point independently, and 
each trajectory has one system condition given the viewing ray directions. 
Since the motion of the person's head is relatively slower than that of his legs, the corresponding system condition is lower and the reconstructed points are more accurate, based on the theory in Sec.~\ref{sec:system_condition}. 
%The average system condition for all the points is 2228. 
Fig.~\ref{fig:trajectory_Qualitative_result} shows the large system condition ($1/\sigma_{\text{min}}=2228$) in this camera setup leads to significant reconstruction errors.

\subsection{Real datasets}
\begin{figure*}
\centering
\subfloat[Rothman dataset (250 frames)]{
\includegraphics[width=0.9\textwidth]{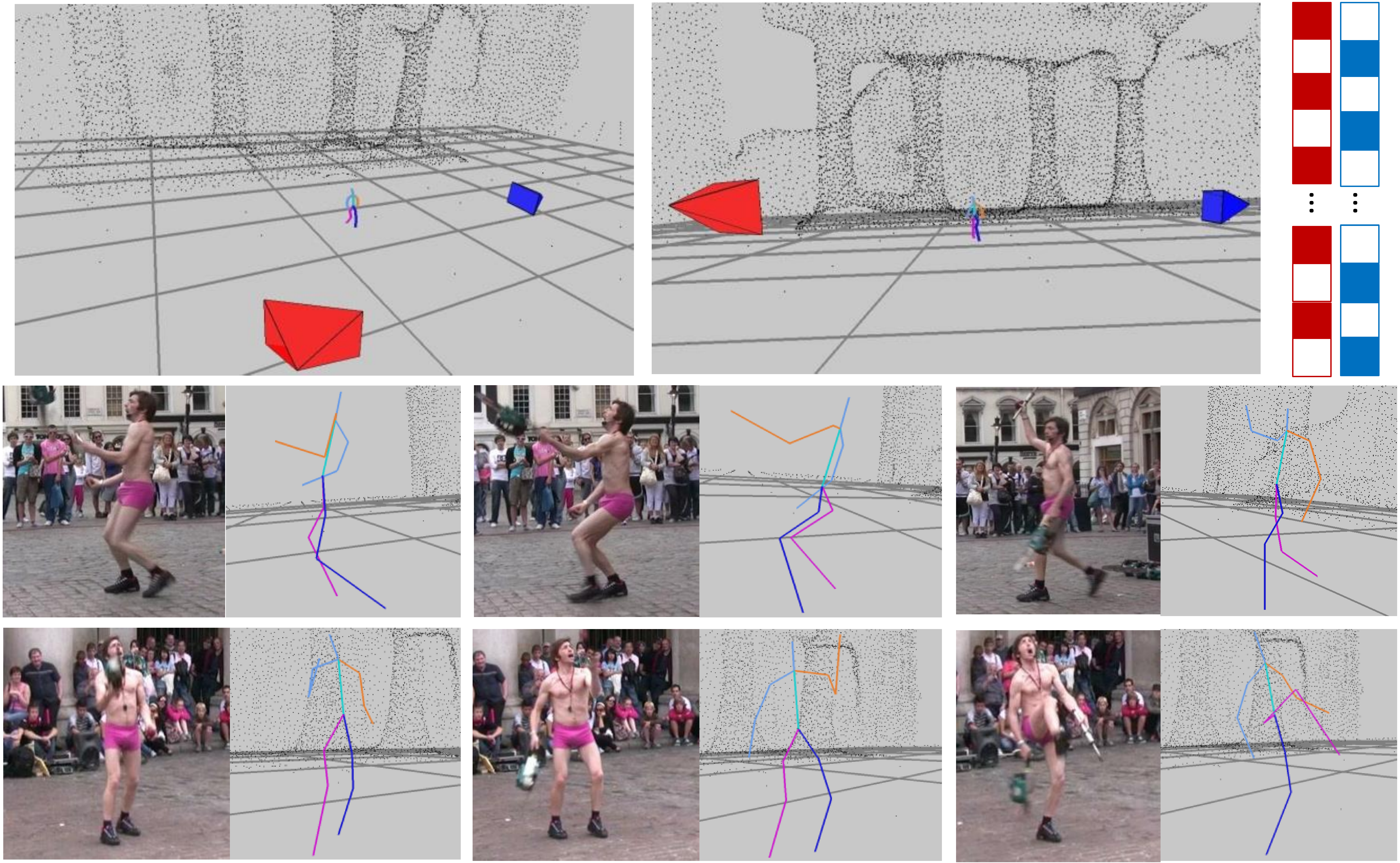} 
}

\subfloat[Juggler dataset (180 frames)]{
\includegraphics[width=0.9\textwidth]{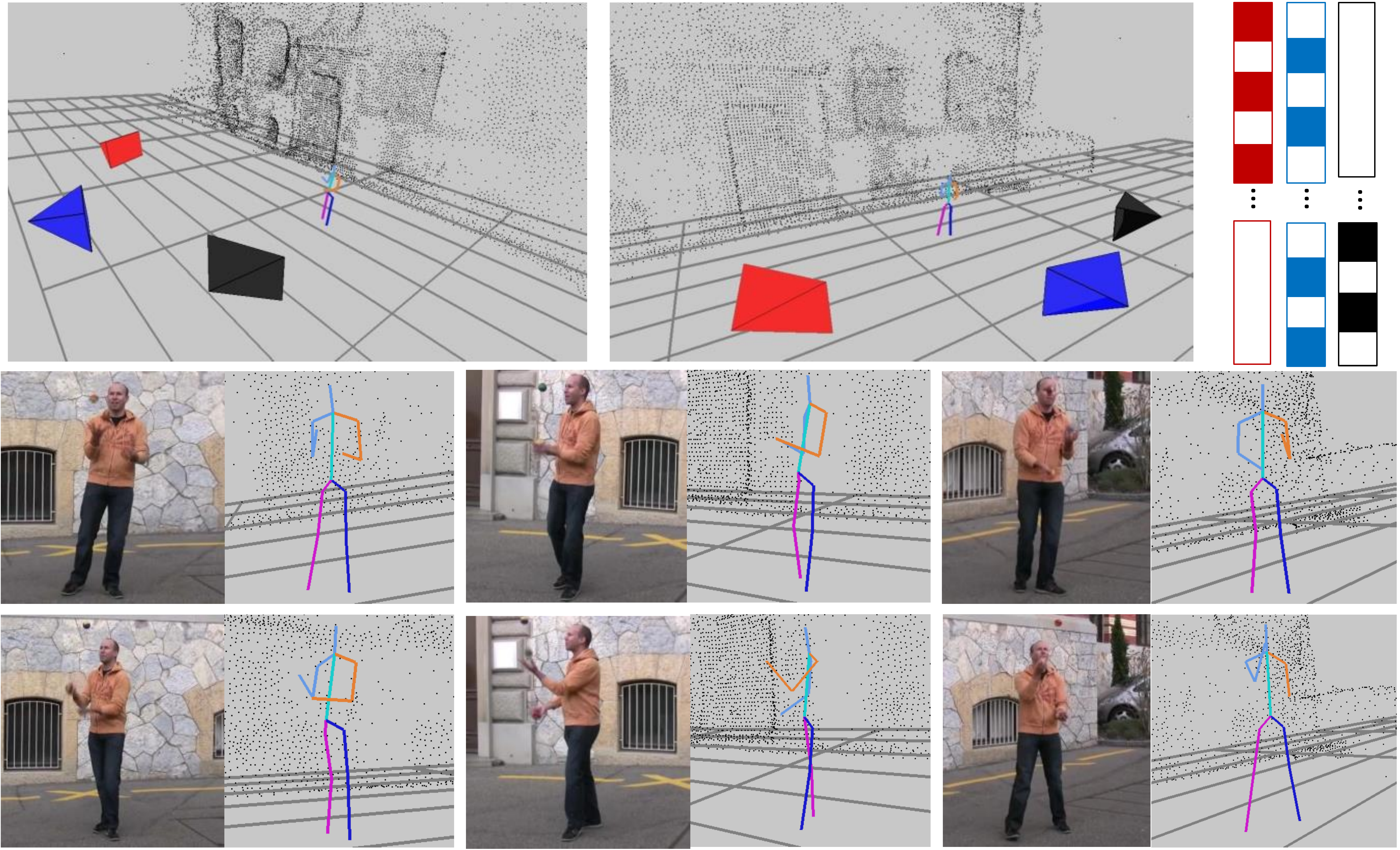}
}
\caption{The datasets presented in \cite{ballan2010unstructured}. The frame rate of each camera is 12.5 Hz. For each dataset, the top left two show the camera configuration, the top right describes the temporal distribution of each image sequence (a colored grid means the camera of the same color captures one frame at a time instance), and the bottom shows example reconstruction results. }
\label{fig:onlinedata}
\end{figure*}

For experiments on real image capture, we use the Juggler and Rothman datasets from \cite{ballan2010unstructured}. Given that the original datasets were synchronized, we sample the video frames to avoid concurrent captures (see Fig.~\ref{fig:onlinedata}). 
We do not use the datasets in \cite{Basha_ECCV2012,park20103d} because they only provide images with large temporal discrepancy, and therefore the shape residual is large (\ie Eq.~(\ref{eq:linear_comb_2}) does not hold). We also capture a new dataset of a person juggling using three iPhone6 and one iPhone5 without temporal synchronization. 

\begin{figure*}
\centering
\includegraphics[width=0.84\textwidth]{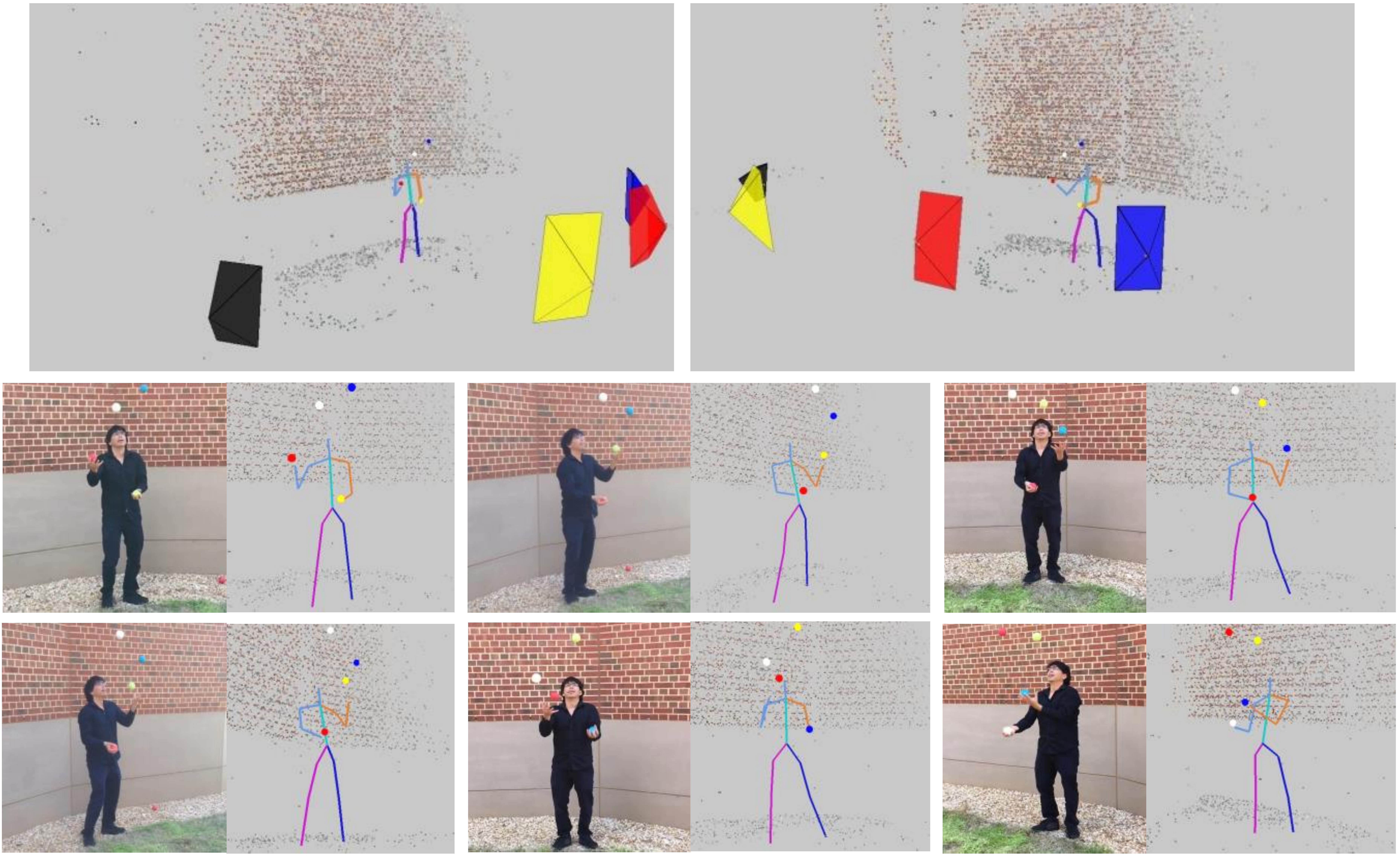}
\caption{Results of a person juggling. Note we reconstruct the four juggler balls in addition to the person. The image sequence from iPhone6 and iPhone5 have frame rates of 10 Hz and 6.25 Hz respectively}
\label{fig:juggler2}
\end{figure*}

%The camera poses and internal calibration are provided for both of these datasets.
We perform manual feature labeling on the input sequences and provide the obtained set of 2D measurements as input for our estimation process.
For visualization purposes, Figs. \ref{fig:onlinedata} and \ref{fig:juggler2} depict the estimated 3D geometry by connecting the estimated position of the detected joint elements through 3D line segments.

% !TEX root = video3D_l1.tex

\section{Conclusion and contributions} \label{sec:conclusion}

The contributions of our framework encompass:
\begin{enumerate}%[topsep=-1ex,itemsep=-1ex,partopsep=1ex,parsep=1ex]
\item {\bf Problem Definition}. We are the first to address the problem of dynamic 3D reconstruction using unsynchronized cross-video streams.
\item {\bf Methodology Formulation}. We pose the problem in terms of a self-expressive dictionary learning framework leveraging a novel data-adaptive local 3D interpolation model.  
\item {\bf Implementation Mechanisms}. We define and solve a biconvex optimization problem and develop an efficient ADMM-based solver amenable for parallel implementation.
\end{enumerate}

To our knowledge, we are the first to use the self-expression prior to solve the problem of dynamic object reconstruction. This prior has the potential to be applied in the traditional NRSFM problem, for which most of the existing methods make use of the assumption of representing shapes using a fixed number ($K$) of shape bases. Our proposed method was successfully evaluated on both real and synthetic data. This is a first step towards dynamic 3D modeling in the wild. 

%\section*{Acknowledgment}
%This research was supported by NSF grants IIS-1349074 and  CNS-1405847.

\bibliographystyle{IEEEtran}
\vspace{-25pt}
\begin{IEEEbiography}[{\includegraphics[width=1in,height=1.25in,clip,keepaspectratio]{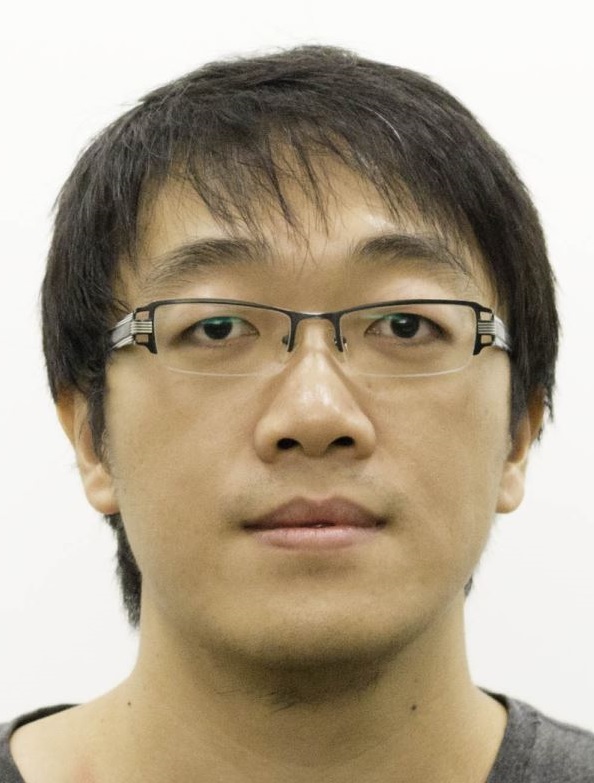}}]{Enliang Zheng}
is currently a PhD student in the computer science department of the Univeristy of North Carolina at Chapel Hill. His research interests include static and dynamic object reconstruction, camera pose estimation in structure from motion, image based virtual view synthesis, etc. Before his PhD study, he received the Bachelor and Master degrees from Shandong University and Shanghai Jiaotong University in 2006 and 2009, respectively. 
\end{IEEEbiography}
\vspace{-25pt}

\bibliographystyle{IEEEtran}
\vspace{-20pt}
\begin{IEEEbiography}[{\includegraphics[width=1in,height=1.25in,clip,keepaspectratio]{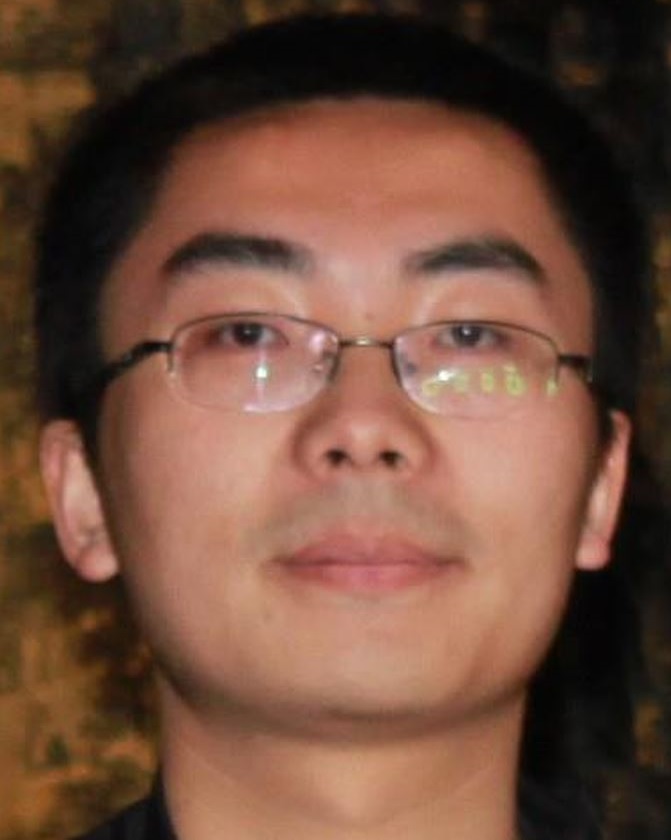}}]{Dinghuang Ji}
received the BE degree in computer science from the University of Science and Technology of China in 2009, and the MS degree in computer science from the Institute of Computing Technology in 2013. He is currently working toward his PhD degree in Department of Computer Science at UNC Chapel Hill. His research interests include 3D computer vision, image processing and graphics.
\end{IEEEbiography}
\vspace{-20pt}

\bibliographystyle{IEEEtran}
\vspace{-20pt}
\begin{IEEEbiography}[{\includegraphics[width=1in,height=1.25in,clip,keepaspectratio]{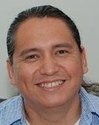}}]{Enrique Dunn}
  completed a doctorate in Electronics and Telecommunications in 2006, and a master’s degree in Computer Science in 2001, both from the Ensenada Center for Scientific Research and Higher Education, M\'exico. His degree in computer engineering is from the Autonomous University of Baja California, M\'exico in 1999. He is currently a research assistant professor in the Department of Computer Science of the University of North Carolina at Chapel Hill. His research interests include multi-view geometry, structure from motion, dense 3D modeling, large-scale crowd-sourced image analysis, and computational intelligence.
\end{IEEEbiography}
\vspace{-20pt}

\bibliographystyle{IEEEtran}
\vspace{-20pt}
\begin{IEEEbiography}[{\includegraphics[width=1in,height=1.25in,clip,keepaspectratio]{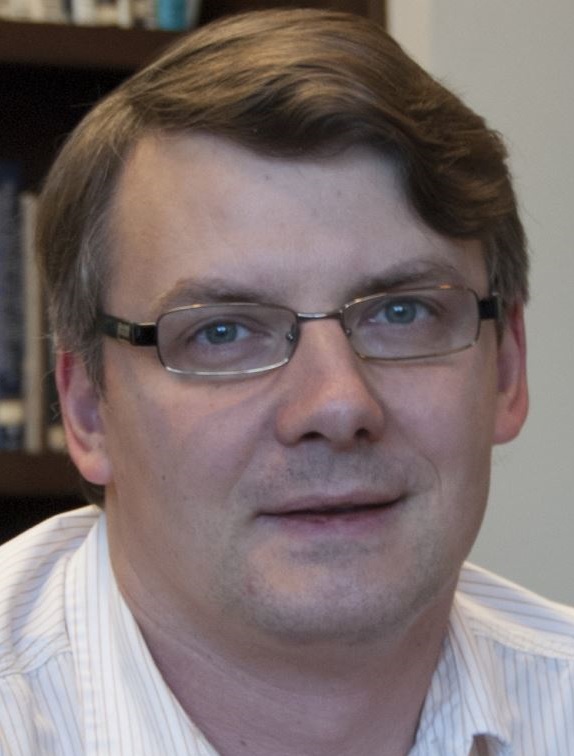}}]{Jan-Michael Frahm}
 received the PhD degree in computer vision in 2005 from the Christian Albrechts University of Kiel, Germany. 
%His dissertation, “Camera Self-Calibration with Known Camera Orientation” received the prize for that year’s best PhD dissertation in CAU’s College of Engineering. 
His diploma in computer science is from the University of L\"{u}beck. He is currently an associate professor at the University of North Carolina at Chapel Hill. His research interests include structure from motion, real-time multiview stereo, robust estimation methods, high-performance feature tracking, and the development of data-parallel algorithms for commodity graphics hardware.
\end{IEEEbiography}
\vspace{-20pt}

{\small
\bibliographystyle{ieeetran}
\bibliography{egbib}
}

%%%%%%%%%% Merge with supplemental materials %%%%%%%%%%

\clearpage
\onecolumn
\begin{center}
\textbf{\large Supplemental Materials: Self-expressive Dictionary Learning for Dynamic
3D Reconstruction}
\end{center}

\setcounter{equation}{0}
\setcounter{figure}{0}
\setcounter{table}{0}
\setcounter{page}{1}
\makeatletter
\renewcommand{\theequation}{\arabic{equation}}
\renewcommand{\thefigure}{\arabic{figure}}

\section{The reason our formulation (Eq.~(\ref{eq:l1_new_equal})) keeps the sparsity-inducing effect.}

\begin{figure}[b]
\centering
\subfloat[lasso]{
\includegraphics[width=0.4\textwidth]{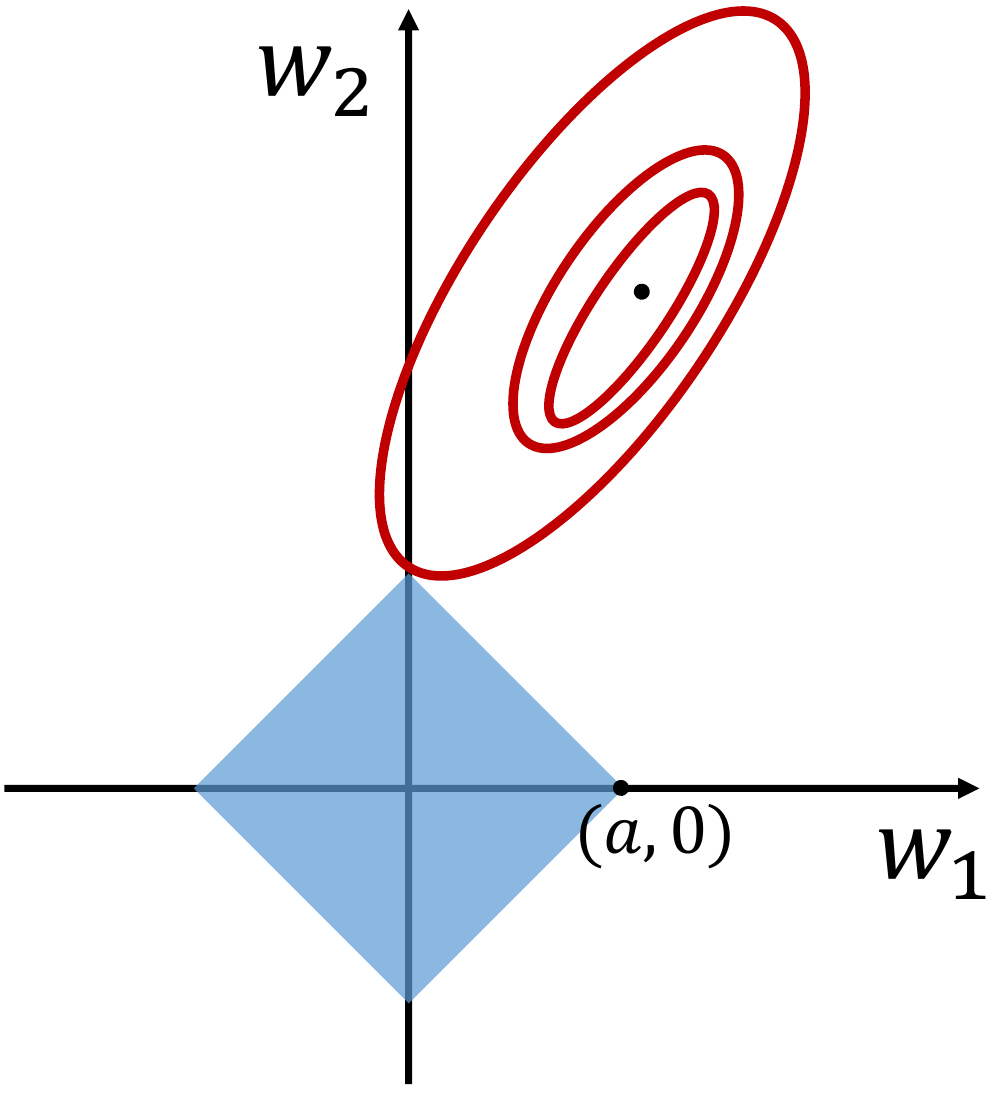}\label{fig:lasso} } 
\subfloat[our formulation]{
\includegraphics[width=0.4\textwidth]{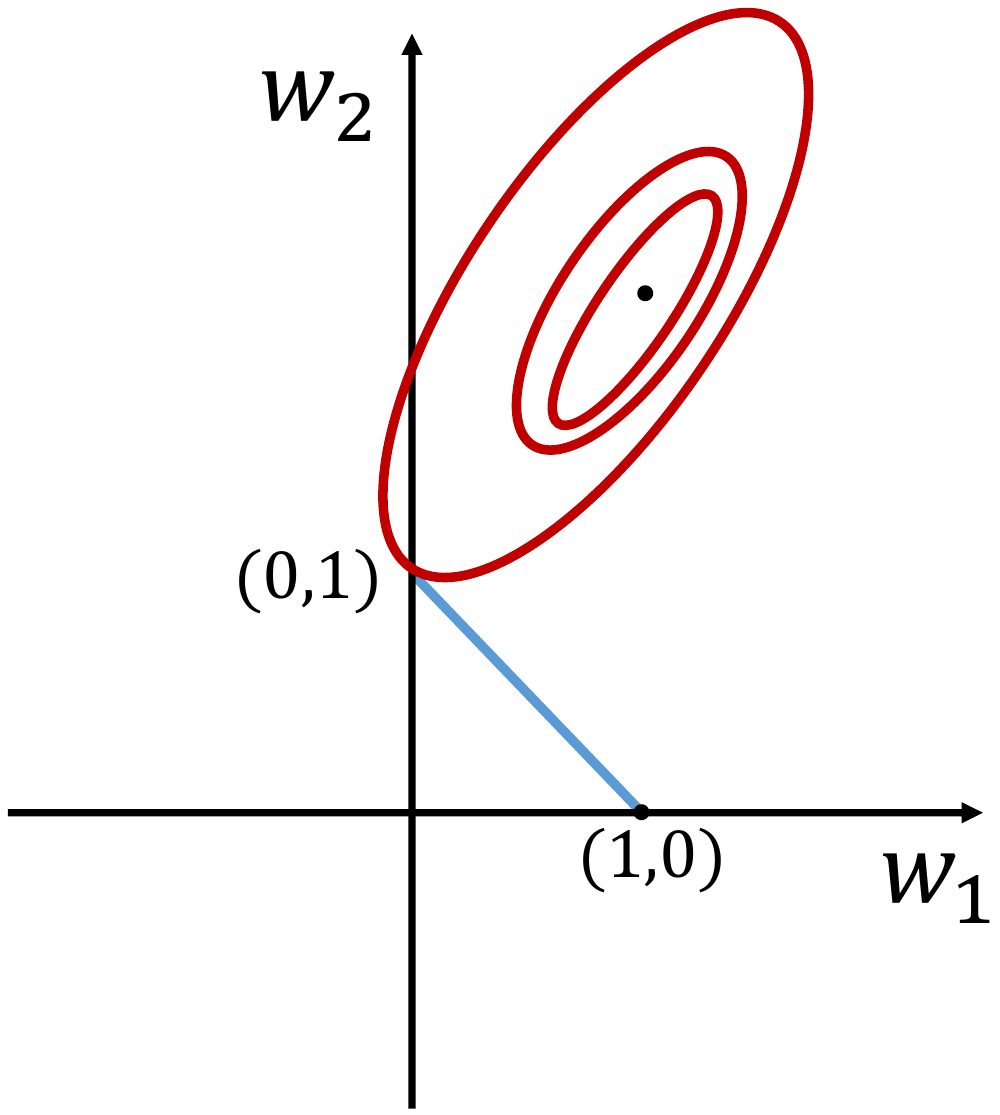} \label{fig:sup_sum_to_1}
}
\caption{Estimation picture of lasso and our formulation}
\end{figure}

First we consider the lasso problem given by
\begin{equation}
\begin{aligned}
& \underset{\mathbf{w}}{\text{minimize}} && ||\mathbf{A}\mathbf{w} - \mathbf{b}||_2^2 \\
& \text{subject to} && ||\mathbf{w}||_1 \leq a,
\end{aligned}
\label{eq:lasso}
\end{equation}
where $\mathbf{A}$ is a matrix, and $\mathbf{w}$ and $\mathbf{b}$ are vectors. 
Hastie \etal~\cite{opac-b1127878} describe the intuition behind the mechanism that Eq.~(\ref{eq:lasso}) is likely to produce sparse results.
Fig.~\ref{fig:lasso} depicts lasso when the size of vector $\mathbf{w}$ is 2. Denoting $\mathbf{w}=[w_1,w_2]$, the residual $||\mathbf{A}\mathbf{w} - \mathbf{b}||_2^2$ has elliptical contours (level set), and the constraint region for lasso is defined by the diamond $|w_1| + |w_2| \leq a$.
%The output of Eq.~\ref{lasso} is likely to produce sparse results is because 
Eq.~(\ref{eq:lasso}) finds the first point where the elliptical contours hit the diamond constraint. If the solution hits the corner, then it has one parameter $w_j$ equal to zero. When $\mathbf{w}$ is in higher dimension, there are more opportunities for the estimated $\mathbf{w}$ to be sparse.

For our formulation, we minimize the problem
\begin{equation}
\begin{aligned}
& \underset{\mathbf{w}}{\text{minimize}} && ||\mathbf{A}\mathbf{w} - \mathbf{b}||_2^2 \\
& \text{subject to} && \mathbf{w} \cdot  \mathbf{1} = 1, \\
&					&& \mathbf{w} \geq 0. 
\end{aligned}
\label{eq:our_formulation}
\end{equation}
In the case that $\mathbf{w}$ has dimension two, we have the constraint that the points have to stay on the line segment (see Fig.~\ref{fig:sup_sum_to_1}), instead of the diamond constraint for the lasso problem. The line segment connects the point (1,0) and (0,1) on the first quadrant. Similar to the lasso problem, the elliptical contours is likely to hit the end of the line segment, hence producing sparse results. 

Note the sparsity effect of Eq.~(\ref{eq:our_formulation}) is introduced by Chen \etal~\cite{chen:hal-00995911}, and they also propose an efficient solver for this problem.

%\begin{thebibliography}{11}
%\bibitem{Hastie} Hastie, Trevor ; Tibshirani, Robert ; Friedman, Jerome: The Elements of Statistical Learning. Springer New York Inc., 2001 (Springer Series in Statistics)
%\end{thebibliography}

\end{document}